\documentclass[10pt,oneside,twocolumn,journal]{IEEEtran}
\usepackage{times}
\usepackage{epsfig}
\usepackage{graphicx}
\usepackage{subfig}
\usepackage{amsmath}
\usepackage{amssymb}
\usepackage{multirow}
\usepackage{slashbox}
\usepackage{bm}
\usepackage{paralist}
\usepackage{color}
\usepackage{url}

\usepackage[square,numbers,sort&compress]{natbib}
\usepackage[font=small]{caption}

\def\widthfour{0.22\linewidth}
\def\widthfifth{0.19\linewidth}
\def\widththird{0.30\linewidth}
\def\hspacefigure{\hspace{0.02in}}

\newcommand{\CUT}[1]{}
\newcommand{\comment}[1]{}
\newcommand{\wh}[1]{\textcolor{black}{#1}}

\newcommand{\xwc}[1]{\textcolor{blue}{[Comment:#1]}}

\ifCLASSOPTIONcompsoc
\else
\fi

\ifCLASSINFOpdf
\else
\fi

\begin{document}
%
\title{Tracking-by-Counting: Using Network Flows  on Crowd Density Maps for Tracking Multiple Targets}
%
%
%
%

\author{Weihong Ren,
		Xinchao Wang, 
		Jiandong Tian,
		Yandong Tang,
		and Antoni B. Chan}

\maketitle

\begin{abstract}
State-of-the-art multi-object tracking~(MOT) methods follow the
tracking-by-detection paradigm, where object trajectories are obtained
by associating per-frame outputs of object detectors.
In crowded scenes, however, detectors often fail to obtain accurate
detections due to heavy occlusions and high crowd density.
In this paper, we propose a new MOT paradigm, tracking-by-counting, tailored for crowded scenes. Using crowd density maps, we jointly model detection, counting, and tracking of multiple targets as a network flow program, 
which simultaneously finds the global optimal detections and trajectories of multiple targets over the whole video. 
This is in contrast to prior MOT methods that either ignore the crowd density and thus are prone to errors in crowded scenes, 
or rely on a suboptimal two-step process using heuristic density-aware point-tracks for matching targets.
Our approach yields promising results on public benchmarks of various domains including people tracking,
cell tracking, and fish tracking.
\end{abstract}


%
\IEEEpeerreviewmaketitle

\section{Introduction}\label{sec:introduction}
\IEEEPARstart{M}{}ultiple-object tracking (MOT) is crucial for many computer vision tasks such
as video analytics. Despite many years of effort, MOT remains a very challenging
task due to factors like occlusions between the targets.
Recent MOT approaches have been focused on the
\emph{tracking-by-detection} paradigm, whose goal is to first detect the targets
in each frame and then associate them into full trajectories.
Such approaches have been successful in scenarios with low-density of targets.
In crowded scenes, however, they often fail to extract the
correct trajectories due to the detection failures caused by occlusions and
the high densities of targets, even with state-of-the-art
detectors trained on large-scale datasets~\cite{felzenszwalb2010object,renNIPS15fasterrcnn,Redmon2016You,tang2014detection,yang2016exploit,benenson2012pedestrian}.

\begin{figure}[t]
\centering
\includegraphics[width=0.98\linewidth]{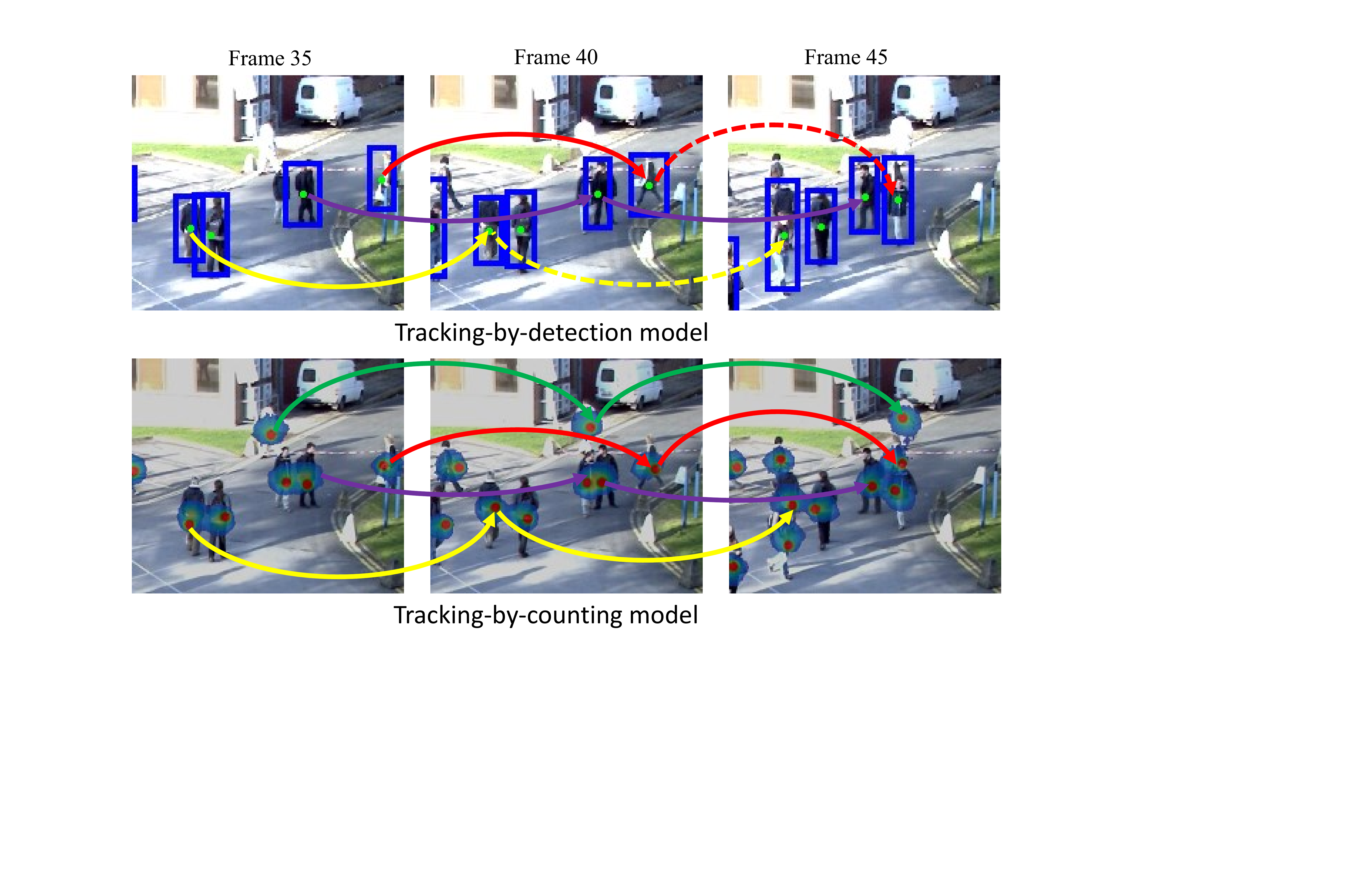}
\caption{Tracking results of tracking-by-detection and tracking-by-counting models on~\emph{PETS2009}. For clear visualization, we only show part of the trajectories in the scene. Solid lines denote the correct associations while the dotted lines denote the wrong ones. 
The object detections used by the tracking-by-detection model are obtained by Faster-RCNN, which fails to localize some occluded people. Our tracking-by-counting model, on the other hand, estimates object densities and imposes the count constraint into the joint detection and tracking framework, resulting in better tracking results especially for occluded people. The red dots on the lower row denote the recovered detections.}\label{density_tracking.pdf}\label{fig:TBDTBC}
\vspace{-0.10in}
\end{figure}

We propose in this paper a novel MOT approach, explicitly designed for handling crowded scenes. We incorporate object counting, a reliable and informative clue in crowded scenarios, into our modeling, and solve the multiple-object  \emph{detection and tracking} simultaneously over the whole video sequence.
Specifically, for each frame, we estimate an {object} density map, based on which
a 3D sliding window is applied for estimating object counts. 
We then construct a spatiotemporal graph over the whole video sequence, 
where each node denotes a candidate detection at a pixel location, 
each edge denotes a possible association, and a sum over 
a set of nodes denotes a count in the corresponding sliding window.
Using the constructed graph, we model the joint detection-tracking-counting problem
as a network flow program. The network-flow constraints and the object-count constraint 
reinforce each other and together benefit the  tracking. The {high quality} solutions of the 
network flow program, or the {complete} object trajectories,
are obtained used off-the-shelf solvers.

We show in Fig.~\ref{fig:TBDTBC} a qualitative comparison between an
advanced tracking-by-detection approach DCEM~\cite{milan2016multi}
and our proposed tracking-by-counting. 
State-of-the-art detectors like Faster-RCNN in this case
miss some partially-occluded pedestrians and thus cause tracking failures.
The proposed tracking-by-counting model, on the other hand, produces the accurate detections for all the objects using the crowd density maps, 
and finds the corresponding associations among them. Also, the tracking-by-counting model is able to localize and track objects with similar appearance 
with the background, like the person in white at the top, which is missed by Faster-RCNN.

Although MOT approaches using density maps have been explored by prior methods,
they either rely on heuristic local point tracks or human-initialized target locations in the first frame. 
Specifically, the approach of~\cite{rodriguez2011density} first detects the head locations of all the people
by encouraging the detections to be consistent with the estimated density maps,
and then counts the point tracks passing through a given paired heads in different frames. 
If this count is larger than a threshold, a match is declared.
One drawback of~\cite{rodriguez2011density} is thus the simple data association rule, which does not use an appearance or a motion model during tracking
and could therefore fail when objects with similar appearance move close to each other. The approach of~\cite{dehghan2017binary} utilizes appearance, motion and contextual cues but relies on initialization of object locations in the first frame.   Furthermore, it is not able to detect new objects coming to the scene.
In contrast to~\cite{rodriguez2011density} and~\cite{dehghan2017binary},
our approach explicitly incorporates the object-count constraints obtained using a 3D sliding window on density maps, 
and network-flow constraints that impose temporal-consistent tracks, into a joint detection-tracking-counting model over the whole video sequence.
We model this tracking-by-counting approach using a network flow program
and obtain the high quality solution using standard commercial solvers.

Our contribution is therefore a 
tracking-by-counting approach that
jointly solves multi-object detection and tracking simultaneously, 
by explicitly incorporating counting constraints from object density maps into the framework.
This is achieved by our network flow programming formulation. The proposed model is the first attempt towards bridging the gap of counting, detection and multi-object tracking.
We demonstrate the power of our approach on benchmarks of various domains including people tracking, fish tracking, and cell tracking.

{The remainder of this paper is organized as follows. In Section \ref{text:related} we review previous work on MOT and crowd counting. In Section \ref{text:tbc} we introduce our tracking-by-counting, and in Section \ref{text:experiments} conduct experiments on benchmark dattasets.  In Section \ref{text:largescale} we extend our model to handle large-scale datasets through incorporating object detection results. Finally Section \ref{text:conclusion} concludes the paper.}

\section{Related work}
\label{text:related}
In this section, we briefly review related work including
tracking-by-detection MOT approaches, deep-learning MOT
approaches and object counting approaches. Comprehensive reviews on MOT can be found in~\cite{luo2014multiple}.

\subsection{Multi-object tracking using tracking-by-detection}
Early multi-object tracking rely on filtering techniques,
e.g., Kalman filtering \cite{black2002multi,li2010multiple} and particle filtering \cite{giebel2004bayesian,okuma2004boosted,oh2009markov}.
Although these methods can be applied to real-time tracking task, they are usually prone to problems like drifts that are difficult to recover from.
Recently, tracking-by-detection has become the  standard paradigm for MOT. The main idea is to split the problem into two parts: object detection and data association. Over the past few years, object detection has seen great improvement thanks to deep learning techniques~\cite{renNIPS15fasterrcnn,Redmon2016You,girshickICCV15fastrcnn}, but data association remains a challenge for multi-object tracking.

Most of the tracking-by-detection methods regard data association as a global optimization problem and focus on designing various optimization algorithm, such as continuous energy minimization~\cite{andriyenko2011multi,milan2014continuous,yu2016solution}, Conditional Random Fields (CRFs) \cite{lafferty2001conditional,yang2012online,milan2016multi} and min-cost network flow \cite{zhang2008global,berclaz2011multiple,pirsiavash2011globally,butt2013multi,chari2015pairwise,wang2016tracking,tang2016multi,shitrit2014multi}. \cite{andriyenko2011multi} formulated multi-object tracking as minimization of a continuous energy function which can incorporate appearance information, physical information and trajectory prior together. To reduce Identity Switches (IDS) generated by the continuous model,  \cite{milan2016multi} further extended \cite{andriyenko2011multi} as a discrete-continuous model which introduces a pair-wise label cost imposing penalty if two labels co-exist that should not appear simultaneously. \cite{lafferty2001conditional} presented a pioneering work using CRFs to segment and label sequence, and explored the possibility of CRFs for multi-object tracking. Based on CRFs, \cite{yang2012online} designed a set of unary functions that model motion and appearance for discriminating all objects, as well as a set of pairwise functions that differentiate corresponding pairs of tracklets. Recent approaches have formulated multi-object tracking as a min-cost network flow optimization problem, where the optimal flow in a connected graph of all detections both selects the best matched candidates and encodes the tracks among them. The solutions to min-cost network flow can be optimized through different algorithms, e.g., shortest paths~\cite{berclaz2011multiple,pirsiavash2011globally}, integer programming~\cite{chari2015pairwise,wang2016tracking,wang2014tracking}, linear programming~\cite{jiang2007linear,shitrit2014multi,wang2017learning} and dynamic programming~\cite{dehghan2015target,wang2015learning}.
In crowded scenes, however, tracking-by-detection approaches often fail to extract the correct trajectories due to the detector failures such as missed detections. To remedy occlusions or missed detections, several works also integrate additional detection information for tracking. E.g., for~\cite{henschel2019multiple}, it uses both joint and body detections for data association, but this method also will fail in highly crowded scenes. \wh{Other early works like~\cite{kratz2011tracking, zhao2012tracking} tried to solve people tracking in structured crowd scenes that have clear and smoonth motion patterns. \cite{bera2014adapt,luo2013generic,fradi2015spatio} are tailed for crowd scene tracking, but they mainly focus on finding missed detections by using motion model and hardcraft features and thus can only handle medium crowd density.}

\subsection{Multi-object tracking using deep learning}
Deep learning techniques have been widely used in object detection~\cite{renNIPS15fasterrcnn,Redmon2016You,girshickICCV15fastrcnn}
and visual object tracking~\cite{held2016learning,huang2017learning,valmadre2017end}.
Recently, several multi-object tracking algorithms also have been proposed based on
convolutional neural networks (CNNs)~\cite{leal2016learning,wang2016joint,Chu_2017_ICCV,Son_2017_CVPR}
and recurrent neural networks (RNNs)~\cite{milan2017online,Sadeghian_2017_ICCV}.
By combining image formation and optical flow as a multi-modal input,
\cite{leal2016learning} proposed to use a siamese CNN to estimate the likelihood that two pedestrian detections belong to the same trajectory. Unlike~\cite{leal2016learning}, \cite{wang2016joint} directly used siamese CNN to encode appearance cue and motion cue to construct a generalized linear assignment model for tracklet association.
\cite{Son_2017_CVPR}  proposed a Quadruplet Convolutional Neural Networks~(Quad-CNN) which has a multi-task loss to jointly learn association metric and bounding-box regression. The target association is then performed by a minimax label propagation using the association metric and refined bounding box from the Quad-CNN. Using the merits of single object tracker, \cite{Chu_2017_ICCV} initialized each detection with a CNN-based tracker which has a spatial-temporal attention mechanism (STAM) to handle the drift caused by occlusion and interaction among targets. Inspired by the Bayesian filtering idea, \cite{milan2017online} presented an RNN that can incorporate all multi-target tracking tasks including prediction, data association, state update as well as initiation and termination of targets within a unified network structure, but maintaining a LSTM for each detection cost too much, especially for crowed scenes. \cite{Sadeghian_2017_ICCV} proposed a multi-object tracking method based on RNN which encodes appearance, motion and interactions together to compute the similarity scores between the tracked targets and the newly detected objects. Also using LSTMs, \cite{maksai2019eliminating} developed a sophisticated model to reduce Identity Switches (IDS), but it discards many occluded detections which leads to low tracking accuracy. Recently, \cite{bozek2018towards} used U-Net~\cite{ronneberger2015u} to address the problem of recognizing bees and their orientations in a densely packed honeybee comb, but it is difficult to apply the method to natural scenes. \cite{Phi2019}~adopts object detector for data association without specific training on tracking task, and its performance depends heavily on the object detector. One drawback of this method is that it causes too many IDS.

\begin{figure*}[!htbp]
\centering
\includegraphics[width=0.80\linewidth]{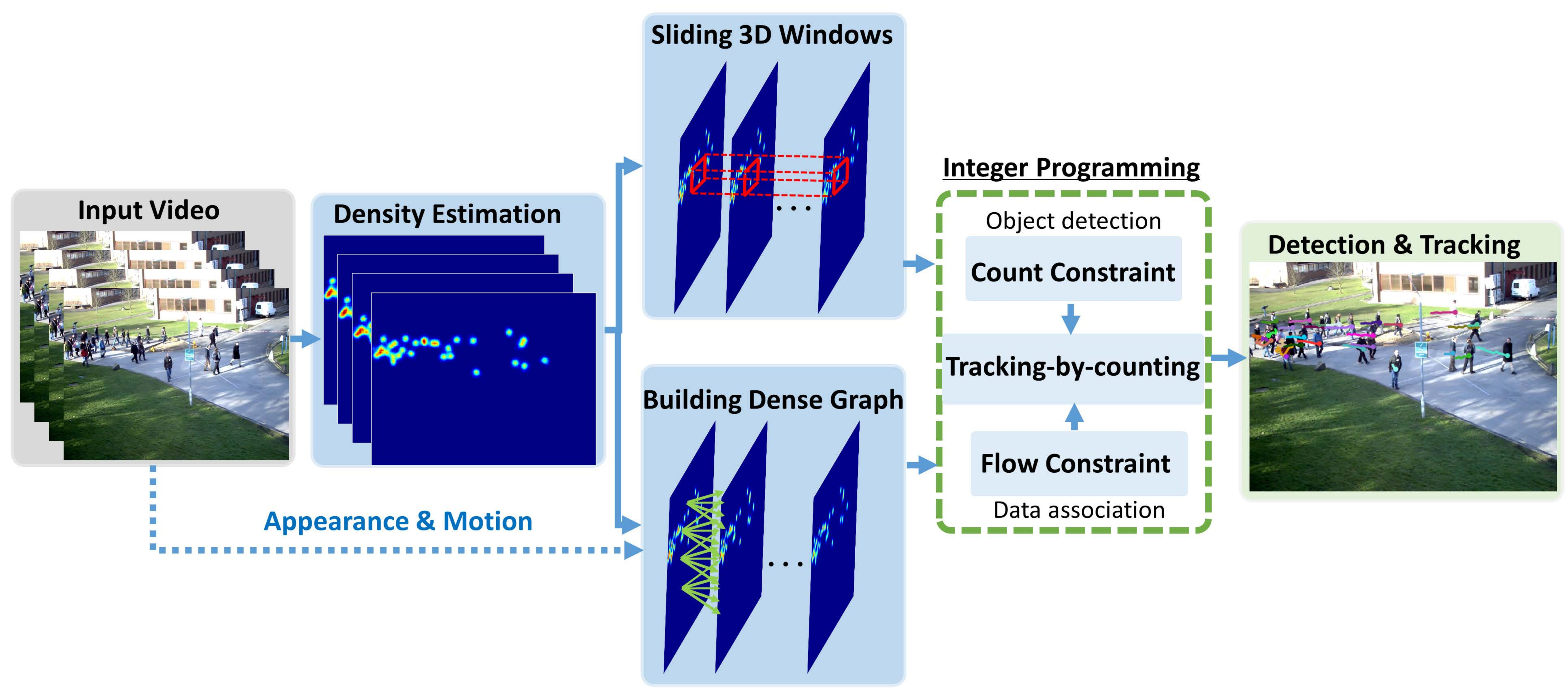}
\caption{The proposed tracking-by-counting model. We first estimate the density maps from the input video. 3D windows are generated over the density map video to obtain object count constraints for detection. Meanwhile we also build a dense graph with flow constraints for data association. Combining the two constraints, we model the tracking-by-counting problem as an Integer Program, whose solution simultaneously generates the detections and tracks.
}\label{tracking-by-counting}
\vspace{-0.2in}
\end{figure*}

\subsection{Crowd counting, detection and tracking using density maps}
Unlike object detection, which focuses on individual targets,
crowd counting methods aim to predict the number of object in an image without explicitly detecting or tracking the object. One effective method is to estimate a crowd density map~\cite{lempitsky2010learning}, where the sum over a region in the image corresponds to the number of object in that region. Previous methods \cite{lempitsky2010learning,fiaschi2012learning,idrees2013multi} usually regard density map estimation as a regression problem, and mainly focus on feature estimation and loss function design to make the estimation robust to scene changes. Recent methods use CNNs\cite{zhang2015cross,sindagi2017generating,sam2017switching,kang2017incorporating} and LSTM \cite{xiong2017spatiotemporal,zhang2017fcn} for density map estimation, and have achieved good performance for a wide range of scenes.

Density maps can also be used for object detection. \cite{ma2015small} first estimated the object density maps, and then used 2D integer programming to predict object detections based on the density maps. In~\cite{rodriguez2011density}, a ``density-aware'' detection and tracking model was proposed to combine individual person detections with crowd density maps. \cite{rodriguez2011density} solves an energy minimization problem where the candidate detections with
high scores in the detection score map are preferred,
and where the detections consistent with the crowd density map are encouraged.
However, their tracking framework does not contain appearance models for each object, and only uses simple nearest-neighbors correspondence between frames, which could fail in crowd scenes when two objects are close to each other.
Unlike the existing methods, our model jointly formulates object detection, counting and tracking  on density maps as
a network flow program, where the object  appearance, motion and spatiality are taken into account and the global flexible solution can be found.

\section{Tracking-by-counting model}
\label{text:tbc}
Our tracking-by-counting model  optimizes object detection and tracking
simultaneously in one framework, shown in Fig.~\ref{tracking-by-counting}.
It  consists of two main components: the object count constraint for detection, and the network flow constraint for data association. The workflow of our method is as follows. The density maps are first estimated from the input video. Then,  sliding windows are selected over the whole video to create count constraints for object detection on the density maps. Meanwhile, a graph is built by densely sampling pixels from the density maps for data association. Finally, our tracking-by-counting model is optimized using Integer Programming. In the remainder of this section, we first introduce object detection on crowd density maps, followed by the new proposed {\em tracking-by-counting} model.

\subsection{Object detection on density maps using count constraints}
Density maps have been used for object detection on 2D image space~\cite{ma2015small,rodriguez2011density}.
Here, we extend object detections to 3D density maps over a video space.
We first define a set of sliding windows over the density map $\bm{\mathcal{D}}\left(t \right)$ for each frame $t$ in a video sequence~\cite{ma2015small}.
The sliding windows (2D) are centered at a pixel in each frame, and move at a fixed step vertically or horizontally in a 3D video space. The spatial sliding window size is set as the average target size in a video, and the temporal size (depth) is the total frame number of the video. {In our implementation, the stride size is (3,6), and windows with low density value ($<$0.005) are dropped to reduce complexity.} Each sliding window at frame $t$ is represented as a mask vector $\textbf{w}_k^t \in \left\lbrace 0, 1\right\rbrace^{N}$, where ${N}$ is the number of pixels in the entire video and $k$ is the window index. The pixels within the ROI of the window are assigned value of~1, and 0 otherwise.

Using the density maps, the number of people in the sliding window $\textbf{w}_k^t$ is estimated as 
\begin{equation} \label{eq1}
	\hat{n}_k^t \approx {\left(\textbf{w}_k^t\right)}^T \textbf{d},
\end{equation}
where the vector $\textbf{d} \in \mathbb{R}^{N}$ is the vectorization of all the density maps $\bm{\mathcal{D}}$ in a video. On the other hand, we encode candidate object detections in the entire video by a single $N$-vector $ \textbf{x} = \left[x_1,x_2,...,x_N\right] \in \left\lbrace0,1\right\rbrace^{N}$, where $x_i$ = 1 if pixel location $i$ has an object centered there, and 0 otherwise. Thus, the object count in the same window $\textbf{w}_k^t$ can also be represented as
\begin{equation}\label{eq2}
	n_k^t ={\left(\textbf{w}_k^t\right)}^T \textbf{x}.
\end{equation}
The optimal object detections $\textbf{x}$ in the video can be obtained by minimizing the counting difference between (\ref{eq1}) and (\ref{eq2}) over all the sliding windows, as in \cite{ma2015small},
\begin{equation}\label{eq3}
{\textbf{x}}^{\ast}= \arg \min \sum\limits_{t=1}^{T}{\sum\limits_{k=1}^{{{K}_{t}}}{\left| {{\left( \textbf{w}_{k}^{t} \right)}^{T}\textbf{x}}-\hat{n}_k^t \right|}},
\end{equation}
where $T$ is the total frame number of a video, and $K_t$ is the number of sliding windows in frame $t$. Note that (\ref{eq3}) only yields object detections without any continuity constraints. In the next subsections, we show how (\ref{eq3}) can be integrated with network flow to perform tracking.

\subsubsection{Bounding box estimation}
To estimate the bounding box for each detected object, we use the density map and perspective map for scale estimation, following~\cite{ma2015small}.  The bounding box is found whose sum on the density map is close to 1, and also consistent with the perspective map of the scene. The perspective map is estimated by linearly interpolating {the detection boxes (produced by detector)} between two extremes of the scene and can reflect the scale changes of an object in a scene. {For
sequences taken with fixed cameras, we only need to estimate one perspective map for all the frames. However, for those with moving cameras (e.g., \emph{MOT17}), we estimate a perspective map for each frame.}
Referring to \cite{ma2015small}, the bounding box for a detection in frame $t$ is estimated by
\begin{align}\label{eq4}
  {\textbf{b}^{*}}=\underset{\textbf{b}_i}{\arg\min} \left| \textstyle{\sum_{p\in {\textbf{b}_{i}}}}{\bm{\mathcal{D}}\left( p,t \right)}-c \right|+\lambda_{b} \Delta \left( {\textbf{b}_{i}},{\textbf{b}_{0}} \right),
\end{align}
where $\Delta \left( {\textbf{b}_{i}},{\textbf{b}_{0}} \right)$ is the total difference between the estimated bounding box $\textbf{b}_i$ and the reference box $\textbf{b}_0$, which is set using the perspective map.
Parameter $\lambda_{b}$ controls the weight of the prior from $\textbf{b}_0$, and $c$ is the target density value, which is between $1$ and $0.8$, which respectively correspond to a looser or tighter bounding box. We have also tried using object detectors for scale estimation, but they often fail in these crowded scenes. {We solve~(\ref{eq4}) using a simple exhaustive search, within a range of plausible bounding boxes ($\pm20\%$) consistent with the perspective map.}

\subsubsection{Density map estimation}
We  adopt the approach of~\cite{zhang2015cross} to estimate density maps using a CNN with 3 convolutional layers and 3 fully connected layers. Note that we estimate a high-resolution density map for tracking by using a sliding window CNN to predict the density for each pixel in the tracking image patch.
\wh{For each dataset, we only train the density estimation network on its own training set, and didn't use other extra information. If a dataset (e.g., \emph{MOT17}) has different  types of sequences, we will fine-tune each sequence using the pretrained model on the whole training set.
For the datasets \emph{Fish} and \emph{Cell}, we use the traditional method [67] to estimate the density maps due to the small size of the training set.}

\subsection{Tracking-by-counting: joint detection and tracking with density maps}
Object detectors often fail to localize objects in crowded scenes, whereas density maps are more robust in these scenarios. By incorporating the count constraint (for object detection) with flow constraint (for tracking) on crowd density maps,
the multi-object tracking problem can be cast as a {\em joint} optimization, which
simultaneously predicts people detections and connections between them
across video frames. Formally, we incorporate the data association cost and flow constraints used in flow-tracking methods with the object count constraint in (\ref{eq3}), resulting in the {\em tracking-by-counting} model:
\begin{equation}\label{eq6}
\resizebox{0.50\textwidth}{!}{
$\begin{aligned}
  \underset{\textbf{x}}{\mathop{\min }}\,&\sum\limits_{t=1}^{T}\sum\limits_{k=1}^{K_t}{\left| \left(\textbf{w}_{k}^t\right)^{T}{\textbf{x}}-{\hat{n}_{k}^t} \right|}+\sum\limits_{ij\in E}{{{c}_{ij}}{{x}_{ij}}} +\sum\limits_{i}{{{c}_{si}}{{x}_{si}}} +\sum\limits_{i}{{{c}_{it}}{{x}_{it}}}\\
 \text{s.t.} &\sum\limits_{i:ij\in E}{{{x}_{ij}}}+x_{sj}={{x}_{j}}=\sum\limits_{i:ji\in E}{{{x}_{ji}}}+x_{jt} \\
 & \sum\limits_{i}{{{x}_{it}}} =\sum\limits_{i}{{{x}_{si}}} , \quad x_i, x_{ij} \in\{0,1\},
\end{aligned}$}
\end{equation}
where $\textbf{x} = [x_1, \cdots, x_N]$.
Note that here the binary indicator variables $\{x_i\}$ denote potential detections on all pixel locations densely sampled from the density maps, {such that $x_i = 1$ if a location $i$ is selected as a detection and appears in some track. ${x}_{ij} \in \left\lbrace0,1 \right\rbrace$ is also a binary indicator variable, where ${x}_{ij}=1$ when location $i$ and location $j$ are both selected as detections for the same track in consecutive frames. The set of possible connections between all the potential object detections, built using spatial proximity, is represented as $E$.  The connection variables ${x}_{si}$ and ${x}_{it}$ represent the start and end of tracks respectively ($s$ is the ``source" node and $t$ is the ``sink" node), and  $c_{si}$ and  $c_{it}$ are the cost of the track start and terminate respectively.} The variable $c_{ij}$ is the edge cost for associating detections at 
locations $i$ and $j$, which 
incorporates geometric location, target appearance and motion direction: 
\begin{equation}\label{eq7}
{{c}_{ij}}=-\alpha {{e}^{-\lambda {{\left\| \bm{\phi}_i-\bm{\phi}_j\right\|}_{2}}}}-\beta H\left( \bm{\phi}_i,\bm{\phi}_j\right)-\gamma \cos(V_i, V_j),
\end{equation}
where $\bm{\phi}_i$ and $\bm{\phi}_j$ are the 2D coordinates of locations $i$ and $j$, respectively, $H\left( \phi_i,\bm{\phi}_j\right)$ is the histogram intersection between the histograms of image patches extracted from locations $i$ and $j$, and $\cos(V_i, V_j)$ is the cosine similarity between the velocities  $V_i$ and $V_j$, which are estimated through {optical flow~\cite{liu2009beyond}}
The tracking-by-counting model  optimizes the object detection and data association at the same time. Furthermore, the flow constraint on the density maps will make the outputs of the detection term in (\ref{eq3}) consistent between frames, which will greatly reduce missed detections. In turn, the enhanced detections benefits the tracking performance. The proposed model makes the first attempt towards the joint multi-object detection, tracking, and counting, and can potentially bridge the gap between MOT and video-based object counting. 


{
To solve the energy function in (\ref{eq6}), we further rewrite it as a standard linear form. We introduce an auxiliary variable $z_k^t$, and let
\begin{equation}
{\left| \left(\textbf{w}_{k}^t\right)^{T}{{\textbf{x}}}-{{n}_{k}^t} \right|} \leq z_k^t,
\end{equation}
where implicitly $z_k^t \geq 0$. Thus, the minimization of ${\left| \left(\textbf{w}_{k}^t\right)^{T}{{\textbf{x}}^{d}}-{{n}_{k}^t} \right|}$ can be 
regarded as the minimization of $z_k^t$. Equation \ref{eq6} can then be rewritten  as
\begin{equation}\label{eq8}
\begin{aligned}
   \underset{\textbf{x,z}}{\mathop{\min }}\,&\sum\limits_{t=1}^{T}\sum\limits_{k=1}^{K_t}{z_k^t}+\sum\limits_{ij\in E}{{{c}_{ij}}{{x}_{ij}}} +\sum\limits_{i}{{{c}_{si}}{{x}_{si}}} +\sum\limits_{i}{{{c}_{it}}{{x}_{it}}}\\
 \text{s.t.} &\sum\limits_{i:ij\in E}{{{x}_{ij}}}+x_{sj}={{x}_{j}}=\sum\limits_{i:ji\in E}{{{x}_{ji}}}+x_{jt} \\
 & \sum\limits_{i}{{{x}_{it}}}=\sum\limits_{i}{{{x}_{si}}} \\
 &\sum\limits_{t=1}^{T}\sum\limits_{k=1}^{K_t}{    \left(\textbf{w}_{k}^t\right)^{T}{{\textbf{x}}}-{{n}_{k}^t} - z_k^t \leq 0 }\\
 &\sum\limits_{t=1}^{T}\sum\limits_{k=1}^{K_t}{    -\left(\textbf{w}_{k}^t\right)^{T}{{\textbf{x}}}+{{n}_{k}^t} - z_k^t \leq 0 }\\
&x_i, x_{ij} \in\{0,1\}, 0\le z_k^t.
\end{aligned}
\end{equation}
At the optimum of (\ref{eq8}), we have $z_k^t = \left| \left(\textbf{w}_{k}^t\right)^{T}{{\textbf{x}}^{d}}-{{n}_{k}^t}\right|$ since $z_k^t$ is minimized in the objective. The above formulation is a {Mixed Integer Linear Programming problem (MILP)} and can be directly solved through a optimization toolbox, such as CPLEX/MOSEK.} { Although (5) as a mixed integer linear program is NP-complete, there are approximation algorithms that can provide quality solutions with moderate computation. CPLEX uses branch-and-cut search, which works well in practice~\cite{chari2015pairwise,ma2015small,rodriguez2011density,
wang2014tracking,wang2016tracking}. When solving (5), we drop candidate points with low density values. In all experiments CPLEX was able to find an efficient and stable solution to (\ref{eq8}) (using tolerance gap of 0.001) without any special initialization of \textbf{x}.} {The small tolerance gap between the primal and dual solutions indicates that a global optimum was likely found.}

\section{Experiments}
\label{text:experiments}
In this section, we evaluate our tracking-by-counting model on MOT using five small-scale datasets featuring crowded scenes, \emph{UCSD}~\cite{chan2012counting}, \emph{LHI}~\cite{Yao2007Introduction}, \emph{Fish}~\cite{ma2015small}, \emph{Cell}~ \cite{mavska2014benchmark,ulman2017objective}
and \emph{PETS2009}~\cite{ferryman2009pets2009}.
In Section \ref{text:largescale} we test on a 2 large-scale datasets, \emph{MOT17}~\cite{milan2016mot16} and \emph{DukeMTMC}~\cite{ristani2016MTMC}. As for the UCF benchmark in \cite{dehghan2017binary}, it only has annotations for part of the people in the highly crowded scene. Besides, even the IP method~\cite{ma2015small} is not able to localize objects in that scene,
since density maps cannot be accurately estimated for UCF.
This is also the reason why~\cite{dehghan2017binary} needs to be initialized with
the ground truth annotations in the first frame and is not able to detect new objects coming into the scene. 

\subsection{Datasets}
\begin{figure*}[t]
\centering
\captionsetup[subfigure]{labelformat=empty}
\subfloat{\includegraphics[width=\widthfifth]{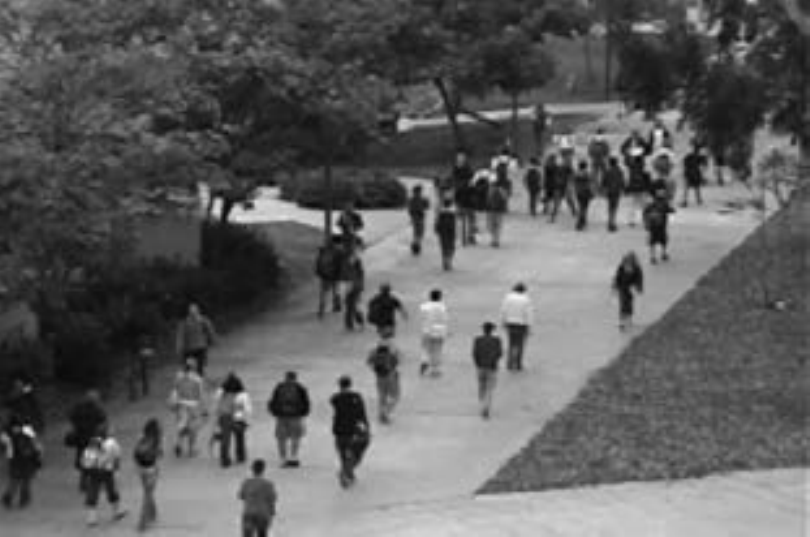}} \hspacefigure
\subfloat{\includegraphics[width=\widthfifth]{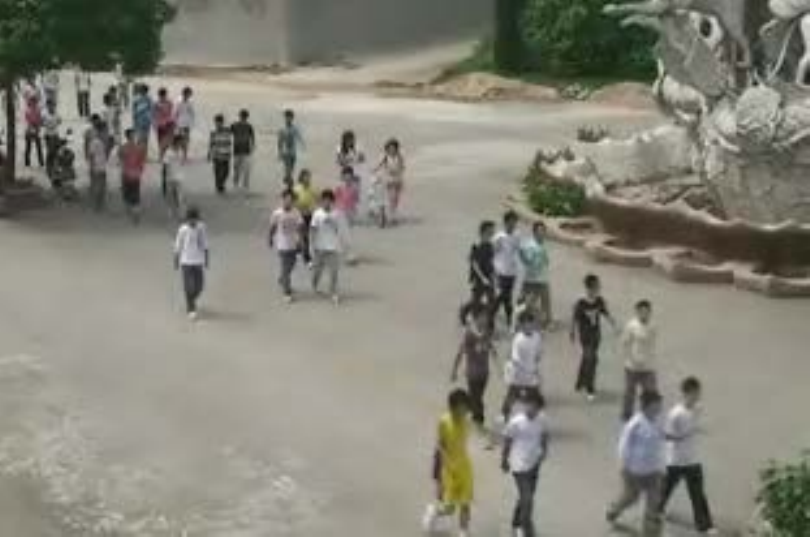}} \hspacefigure
\subfloat{\includegraphics[width=\widthfifth]{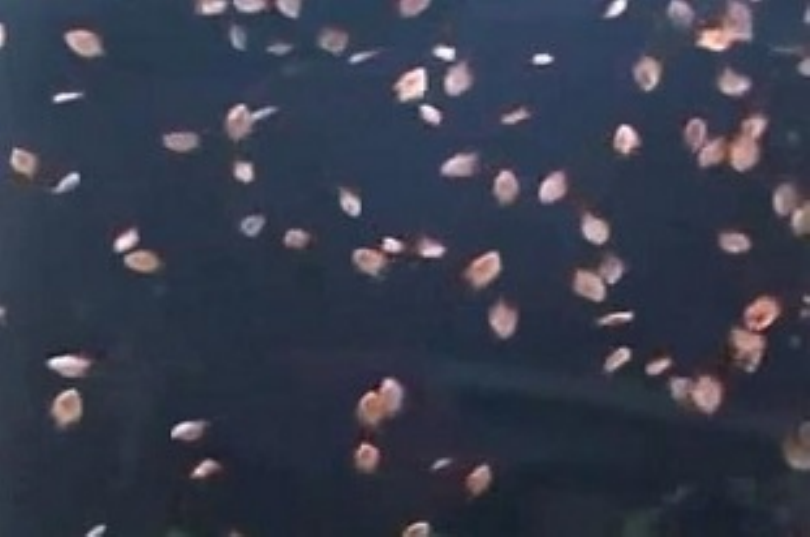}} \hspacefigure
\subfloat{\includegraphics[width=\widthfifth]{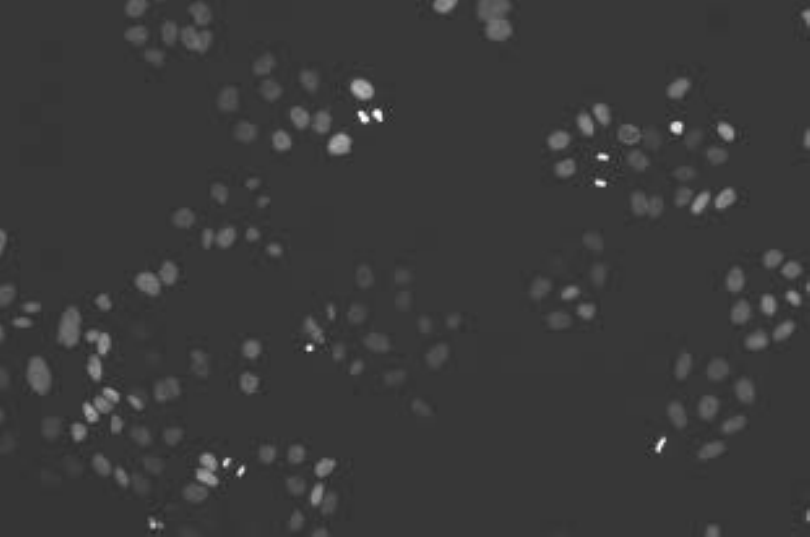}} \hspacefigure
\subfloat{\includegraphics[width=\widthfifth]{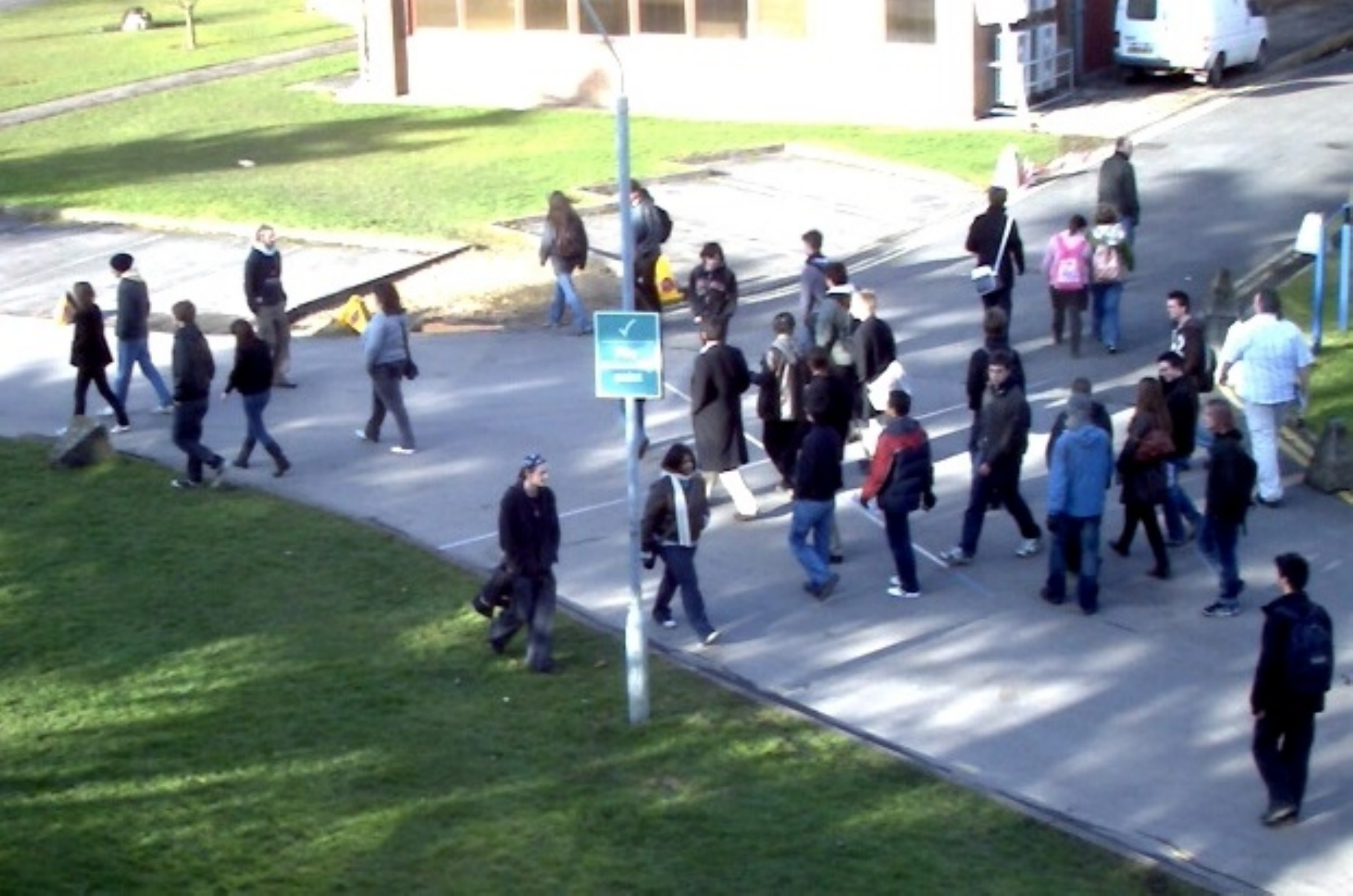}} \hspacefigure \\
\vspace{-0.1in}
\subfloat{\includegraphics[width=\widthfifth]{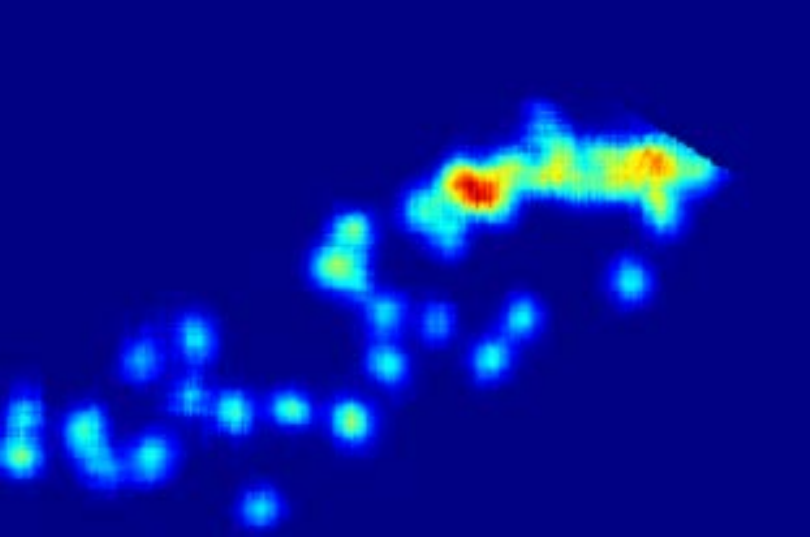}} \hspacefigure
\subfloat{\includegraphics[width=\widthfifth]{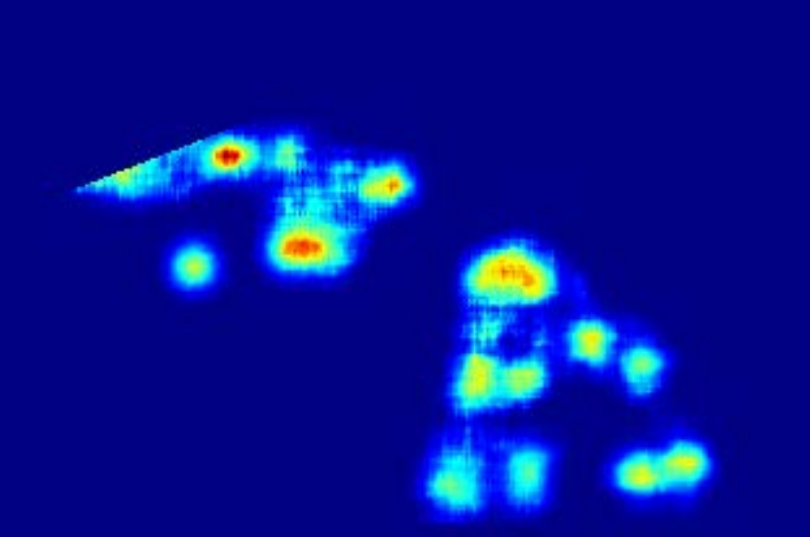}} \hspacefigure
\subfloat{\includegraphics[width=\widthfifth]{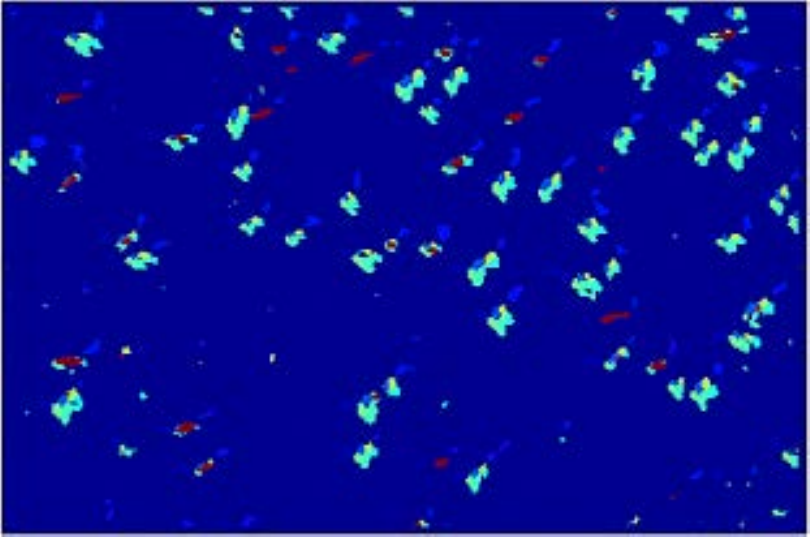}} \hspacefigure
\subfloat{\includegraphics[width=\widthfifth]{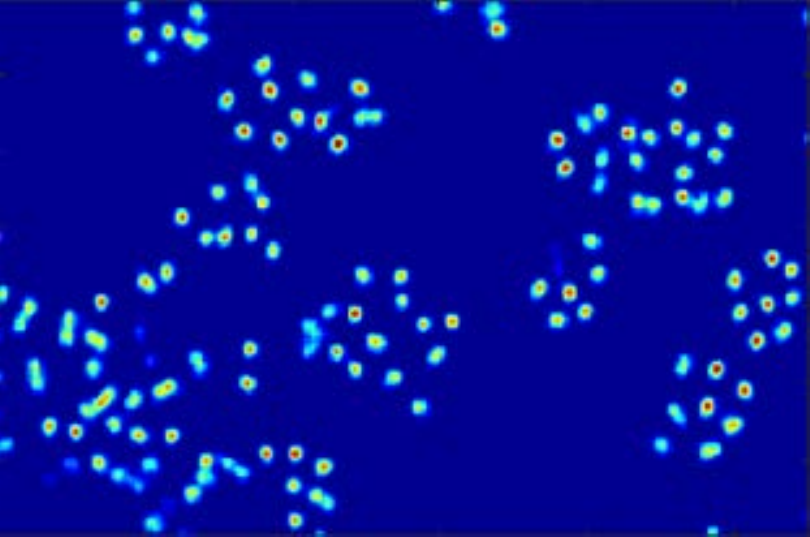}} \hspacefigure
\subfloat{\includegraphics[width=\widthfifth]{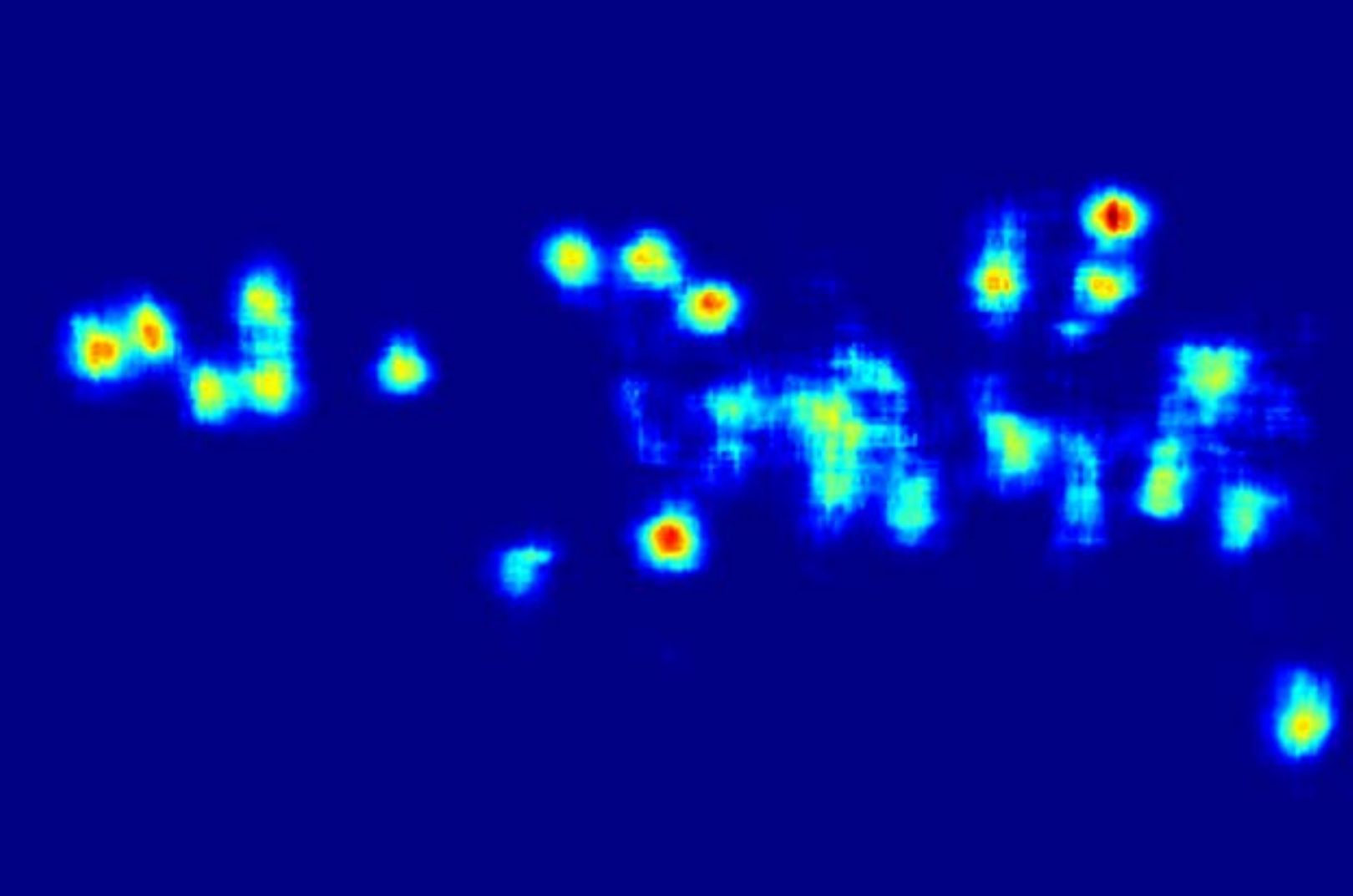}} \hspacefigure
  \caption{Example frames from left to right on the top row: \emph{UCSD}, \emph{LHI}, \emph{Fish}, \emph{Cell} and \emph{PETS2009},
  and their corresponding crowd density maps on the bottom. \emph{UCSD}, \emph{Fish} and \emph{Cell} have the higher crowd densities (average 40-80 objects per frame), and \emph{LHI} and \emph{PETS} have the relatively lower crowd ones (average 20 people per frame). Crowd levels also changes within each video (e.g., 12-34 in  PETS). } \label{example}
\end{figure*}

For the \emph{UCSD} dataset, we choose the most crowded video clip of $200$ frames (238$\times$158) featuring 63 pedestrians.
The \emph{LHI} dataset comprises a color video of $400$ frames (288$\times$352) with $43$ tracks. 
The \emph{Fish} dataset has $129$ frames (300$\times$410) with $279$ intersecting paths. 
The \emph{Cell} dataset  contains $92$ frames 
(350$\times$550), and features $265$ tracks where cells deform frequently.
For \emph{PETS2009} dataset, we choose the S2-L2 sequence of $436$ frames (576$\times$768) with $42$ tracks.
To estimate density maps, we adopt the approach of \cite{zhang2015cross}, except that we estimate a high-resolution density map for tracking by using a sliding window CNN to predict the density for each pixel. Examples of the five datasets and their corresponding density maps are shown in Fig.~\ref{example}.


\subsection{Evaluation metrics}
The trackers are evaluated using the CLEAR
MOT~\cite{bernardin2008evaluating} metrics on the 2D plane.
Multiple Object Tracking Accuracy (MOTA)
accounts for  false positives~(FP),
false negatives~(FN), and identity switches~(IDS),
while the Multiple Object Tracking Precision~(MOTP) measures the average distance
between the ground truth and the tracker output.
A track is considered as a Mostly Tracked (MT) one if
the ground truth trajectory is covered by this track
for $\geqslant$80\% of its time span,
while a  Mostly Lost (ML) track
is the one with $\leqslant$20\%.
Track fragmentation (FM) counts the number of trajectory fragmentations,
and FAF represents the number of false alarms per frame.
RCLL and PRCN are 
the recall and precision of the detections used as the input for a tracker. Finally, IDF1~\cite{ristani2016performance} means the ratio of correctly identified detections over the average number of ground-truth and predicted detections.

\subsection{Experiment setup}

We denote our tracking-by-counting model (Eq.~\ref{eq6}) as ``TBC''.
We set $c_{si}$ = $c_{it}$ = 10 in (\ref{eq6}), and $\lambda=\beta=\gamma=1$ in (\ref{eq7}).
To see the effect of the joint detection-tracking framework, we also consider a variant (denoted as ``TBC3'') that separates object detection and tracking -- TBC (Eq.~\ref{eq6}) is used as a temporally-regularized object detector for every three frames, and {the detections are input into a general flow-tracking method \cite{pirsiavash2011globally}, by replacing the first term in (\ref{eq6}) with $c_ix_i$ for data association (denote as ``FT''). Here, $c_i$ denotes the
cost of selecting detection $i$}, \wh{and we directly set it to 1, which means that we have high confidence for the detections from TBC3.} Our TBC 
model is implemented using Matlab with the CPLEX toolbox on 
a PC (i7 3.4GHz CPU, 8 GB memory).

We compare our TBC 
with four state-of-the-art multi-object trackers, DCEM~\cite{milan2016multi}, GOGA~\cite{pirsiavash2011globally}, MHT~\cite{kim2015multiple} and density-aware (DA)~\cite{rodriguez2011density}. DCEM, GOGA and MHT are typical tracking-by-detection trackers, for which the first step is to obtain a set of object detections for each frame independently.
In our experiments, we consider three object detectors: the retrained DPM 
\cite{felzenszwalb2010object}, the retrained Faster-RCNN 
\cite{renNIPS15fasterrcnn}, and the Integer Programming method based on density map (IP)~\cite{ma2015small}.
{To well retrain the DPM and Faster-RCNN, we upsample the datasets to an appropriate scale for object detections.}
DA  achieves good detection results using DPM detection score maps and density maps, but the original formulation relies on a simple association method.  For fairer comparison, we use DA as a detector and apply the flow-tracking (FT) for data association.
In the text, we denote combinations of tracker and detectors as 
``tracker-detector".

\comment{
{\subsection{Baselines}}

For the baselines DCEM \cite{milan2016multi}, GOGA \cite{pirsiavash2011globally} and MHT \cite{kim2015multiple}, they are all evaluated using three different detectors DPM \cite{felzenszwalb2010object}, Faster-rcnn \cite{renNIPS15fasterrcnn} and Integer programming (IP) \cite{ma2015small}. For density-aware method (DA) \cite{rodriguez2011density}, it uses DPM detection score maps and density maps for object detection which is good enough for tracking, so we don't apply other detections to DA for tracking \xwc{Is it still true?}.
Since DA method has a too simple association rule, we thus use the flow tracking model (\ref{eq5}) denoted as ``FT'' for data association. \xwc{This sentence is not clear, FT for what?}
We denote our tracking-by-counting model as ``DFT''. To evaluate the effect of flow constraint on object detection, we first use our model DFT $\left( \ref{eq6} \right)$ for object detection every $3$ frames, denoted as ``DFT3'', and then input all the detections to FT $\left( \ref{eq5} \right)$ for data association. \xwc{I am actually not very clear about DFT3. Does it produce full tracks?}
We denote each baseline as "tracker-detector" when  referred.
}

\begin{table*}[htbp]
\newlength{\hv}
\setlength{\hv}{0.10cm}
\renewcommand\arraystretch{1.25}
\caption{Multi-object tracking results on \emph{UCSD}.} \label{table1}
\centering
\begin{tabular}{|c@{\hspace{\hv}}c|cccccccccccccc|}
  \hline
  Tracker&Detector&RCLL(\%)$\uparrow$ &PRCN(\%)$\uparrow$&FAF$\downarrow$&GT &MT$\uparrow$&PT&ML$\downarrow$&FP$\downarrow$&FN$\downarrow$&IDS$\downarrow$&FM$\downarrow$&MOTA$\uparrow$&IDF1$\uparrow$&MOTP$\uparrow$
  \\
  \hline
 \multirow{3}{*}{DCEM} 
 &DPM
 &24.6& 66.2 &6.57&62&0&30&32&1313&6616&277&282&6.5&7.4&63.1 \\
 &Faster
 & 36.4& 77.9& 4.54& 62&  2 & 44& 16& 909& 5581& 210& 281& 23.7&31.1& 61.1 \\
 &IP
 &34.1 &74.0 &5.27&62 & 1 &45&16&1054&5782&236&317&19.4&18.5 &60.5 \\
 \hline

 \multirow{3}{*}{GOGA} 
 &DPM
 &11.2&39.0&7.71&62&0&12&50&1543&7791&159&230&-8.2&7.3&61.9\\
&Faster
 &\textbf{57.4}&72.7& 9.43& 62&9&45&8&1886&\textbf{3742}&184&485&33.8&\textbf{38.9}&63.2 \\
&IP
 &54.3&73.7&8.52&62&6&52&\textbf{4}&1703&4008&799&879&25.8&28.8&67.6\\
 \hline

 \multirow{3}{*}{MHT} &DPM
 &13.7 &36.6&10.39&62& 0 &18&44&2078&7575&93&209&-11.1&5.6&61.3 \\
&Faster
 & 51.6 & 65.3&18.29&62 &12&40 &10&3657&4247&289&530&16.6&23.4&60.3 \\
&IP 
&40.3&62.0&16.34&62&7&36&19&3268&5238&291&434&10.2&19.2&62.9\\
 \hline

 \multirow{5}{*}{FT} &DPM 
 & 13.6 & 36.8&10.21&62 &0 &17&45&2043&7586&211&340& -12.1&6.8& 62.2 \\
&Faster 
 & 46.0 &  61.2 & 12.81& 62& 7 & 43&12& 2563& 4735& 108& 593& 15.6& 34.6&61.2 \\
&IP
 &51.6 &64.7&12.35&62 & 8 &47 &7&2470&4248&207&702& 21.1 &31.2&62.1 \\
&DA 
 &20.3 &76.9&\textbf{2.69}&62& 0 &26 &36&\textbf{537}&6991&\textbf{114}&\textbf{302}&12.9&14.4&70.1  \\
 \multicolumn{2}{|c|}{TBC3 (ours)} &50.5 &80.5& 5.38&62 &\textbf{14}& 34 &14&1077&4341&166&419&{36.4}&36.1&\textbf{74.1} \\
\hline
 \multicolumn{2}{|c|}{TBC (ours)}& 49.5& \textbf{82.1}&4.74&62&11& 37&14&948&4435&140&389& \textbf{37.1}& {38.7}&{74.0}  \\
 \hline
\end{tabular}
\end{table*}

\subsection{Evaluation on \emph{UCSD} and \emph{LHI}}
\label{text:ucsdlhi}
Both  \emph{UCSD} and \emph{LHI} are low resolution datasets but contain crowded scenes,
which are challenging for multi-object tracking. To well retrain the DPM and Faster-rnn, we upsample the two datasets to an approproate scale. In Table \ref{table1}, we report the tracking results for different combinations of trackers and detectors on \emph{UCSD}.
{Using the overall MOTA scores}, our 
TBC performs better than DCEM-Faster, GOGA-Faster, MHT-Faster and FT-Faster (e.g., MOTA of 37.1 vs 23.7, 33.8, 16.6, and 15.6), 
which indicates that jointly imposing object count constraint and flow constraint on crowd density maps will greatly improve the tracking results in crowded scenes.

TBC3 uses the model in (\ref{eq6}) with flow constraint for object detection, while FT-IP only uses IP method for object detection. Though the two baselines have the same data association model FT, the detection result of TBC3 is much better than FT-IP (e.g., PRCN of 80.5 vs 64.7), which means that flow constraint on density maps can improve object detection performance by making the detections consistent between frames. In addition, TBC3 also performs better than FT-IP in terms of tracking results (MOTA) (e.g., MOTA of 36.4 vs 21.1). TBC jointly optimizes object detection and tracking over all the frames, while TBC3 separates object detection and tracking. From the tracking results,
TBC achieves higher MOTA than TBC3, \wh{which indicates that using flow constraint on all the frames is
more helpful for detection and tracking than only using flow constraint on part of
the frames. However, TBC takes more time for optimization compared with TBC3}. For the FT tracker, using the IP detector performs better than using the traditional trackers DPM and Faster-rcnn (e.g., MOTA of 21.1 vs -12.1 and 15.6), which indicates that density maps are effective for object detection in crowded scenes.

In Table~\ref{table2}, we summarize the tracking results
 on \emph{LHI}. Our TBC still performs better compared to DCEM-IP, GOGA-Faster, MHT-Faster and FT-DA, showing higher MT and MOTA, e.g., MT of 29 vs 13, 23, 26 and 5, and MOTA of 73.1 vs 25.7, 62.7, 25.7, and 39.0. 
It also demonstrates that min-cost network flow tracking methods perform better than other methods in crowded scenes. For example, GOGA-Faster has higher MOTA 
than DCEM-Faster and MHT-Faster (e.g., MOTA of 62.7 vs 20.8 and 25.7). 
{DA has a relatively better detection result than that of Faster-RCNN and DPM, which also indicates that density maps are effective for object detection in crowded scenes.} From IDF1, our tracker can also achieve good tracking performance on both \emph{UCSD} and \emph{LHI}.

\begin{table*}[htbp]
\setlength{\hv}{0.10cm}
\caption{Multi-object tracking results on \emph{LHI}.} \label{table2}
\centering
\begin{tabular}{|c@{\hspace{\hv}}c|cccccccccccccc|}
  \hline
  Tracker&Detector&RCLL(\%)$\uparrow$ &PRCN(\%)$\uparrow$&FAF$\downarrow$&GT &MT$\uparrow$&PT&ML$\downarrow$&FP$\downarrow$&FN$\downarrow$&IDS$\downarrow$&FM$\downarrow$&MOTA$\uparrow$&IDF1$\uparrow$&MOTP$\uparrow$
  \\
  \hline
\multirow{3}{*}{DCEM} &DPM
 &23.4&63.1&{3.19}&43&1&19&23&{1275}&7147&274&313&6.8&6.6&67.6 \\

&{Faster}
&45.6&66.0&5.46&43&7&23&13&2186&5077&126&198&20.8&31.1&68.4 \\

&IP
 & 58.4 & 65.8 & 7.08& 43 &13&24 & 6&2830&3878&224&238&25.7&31.6&70.9 \\
 \hline

\multirow{3}{*}{GOGA} &DPM
 &34.3&89.2&{0.97}&43&4&26&13&{388}&6129&307&251&26.8&33.4&74.1\\
 &Faster 
 &74.5&88.3&2.31&43&23 &19&1& 924 &2379 &173 &295&62.7 & 51.9&70.0\\
 &IP 
 &60.8&81.0&3.33&43&10&28&5&1331&3660&503&554&41.1&42.5&71.4\\
 \hline
\multirow{3}{*}{MHT} &DPM 
 &43.0&66.2&5.12&43&5&23&15&2046&5313&90&\textbf{69}&20.1&29.6&74.5 \\
&{Faster }
 & \textbf{79.6}& 61.1&11.84&43& 26&16&1&4734&\textbf{1903}&291&206&25.7&38.8&69.5 \\
&IP 
  &58.2&63.0&7.97&43&12& 26& 5&3187&3897&232&220&21.6&36.3&72.1\\
 \hline
\multirow{5}{*}{FT} &DPM 
 &39.4&82.2&1.99&43&5&25&13&797&5656&402&399&26.5&36.4&73.6  \\
&Faster 
 &77.1&78.4&4.95& 43&26 &16& 1&1979&2139&121&297& 54.6&56.9&75.9 \\
&IP 
 &72.6&95.3& 0.84&43 & 16&26&1&{337}&2559&167&242&67.2 &52.4&77.0\\
&DA 
 &61.6&82.0&3.15&43&5 &35&3&1258&3584&848&888&39.0&21.5&72.6    \\
\multicolumn{2}{|c|}{TBC3 (ours)}&73.3&\textbf{95.9}&\textbf{0.73}&43&18&24 & 1&\textbf{293}&2490&76&132& 69.3&63.5&\textbf{81.6} \\
 \hline
 \multicolumn{2}{|c|}{{TBC (ours)}}&79.1 &93.6 &1.26&43&\textbf{29}& 13 & 1& 504 &1945 & \textbf{63} &110& \textbf{73.1}&\textbf{64.6}& 81.3 \\

\hline
\end{tabular}
\end{table*}

\subsection{Evaluation on \emph{Fish} and \emph{Cell}}

\begin{table*}[htbp]
\setlength{\hv}{0.10cm}
\renewcommand\arraystretch{1.20}
\caption{Multi-object tracking results on \emph{Fish} and \emph{Cell}.} \label{table3}
\centering
\scalebox{0.93}{
\begin{tabular}{|c|c@{\hspace{\hv}}c|cccccccccccccc|}
  \hline
  Dataset&Tracker&Detector&RCLL(\%)$\uparrow$ &PRCN(\%)$\uparrow$&FAF$\downarrow$&GT &MT$\uparrow$&PT&ML$\downarrow$&FP$\downarrow$&FN$\downarrow$&IDS$\downarrow$&FM$\downarrow$&MOTA$\uparrow$&IDF1$\uparrow$&MOTP$\uparrow$ \\
  \hline
 \multirow{6}{*}{\emph{Fish}}&{DCEM} &IP
 & 37.5&59.5&30.51&279 &11&196&{72}&3936&4280&602&592&8.7&16.7&65.9 \\
 &{GOGA} &IP
 &35.5&\textbf{71.5}&\textbf{7.51}&279 &12&170&97&\textbf{969}&4415 &474&478&14.5&23.7&68.3\\
 &{MHT}&IP 
 &\textbf{44.2} &58.2&38.01&279&\textbf{26}&192& \textbf{61}&4903&\textbf{3819}&663&541&7.0&14.4 &66.7\\
 &{FT} &IP 
 &43.1 &62.9&13.49&279&15&183& 81&1740&3894&262&604&13.9&\textbf{31.3}& 67.0\\
 &\multicolumn{2}{|c|}{TBC3 (ours)}&32.5&67.3 &8.41&279&16&134&129&1085&4621&\textbf{141}& \textbf{423}&14.6&28.0&\textbf{68.6} \\
&\multicolumn{2}{|c|}{{TBC (ours)}}&37.9&65.3&10.67&279&19&155&105&1377&4254&164 &446&\textbf{15.4}&29.9&68.3 \\
\hline
\multirow{6}{*}{\emph{Cell}}&{DCEM} &IP
 &  61.5 &70.7&23.96&265& \textbf{91}&109 &65&2204&3329&161&237&34.1 &42.1&73.6 \\
 &{GOGA} &IP
 &77.7 &\textbf{98.0}&\textbf{1.52}&265&165&59 &41&\textbf{140}&1928&184&161& 73.9 &\textbf{69.3}&83.8\\
 &{MHT}&IP 
 &81.8&74.4&26.48&265 &181 &56 &\textbf{28}&2436&1576&389&165& 49.1&47.0&82.0\\
 &{FT} &IP 
 &80.6&97.4&2.01&265&185&51&29&185&1673&216&268&76.0 &61.8&83.8\\
 &\multicolumn{2}{|c|}{TBC3 (ours)}&81.2 &96.4 &2.88&265&185&51&{29}&265&1623&207 &236&75.7&62.7&\textbf{88.0} \\
&\multicolumn{2}{|c|}{{TBC (ours)}}&\textbf{85.4}&95.4&3.87&265&{199}&36&30&356&\textbf{1263}&\textbf{153}& \textbf{138}&\textbf{79.5}&63.7&\textbf{88.0} \\
\hline
\end{tabular}
}
\end{table*}

\begin{table*}[!htbp]
\setlength{\hv}{0.10cm}
\caption{ Comparisons  with 
tracking-by-counting models on S2L2: (top) using MOT evaluation server; (bottom) using ground-truth from Milan et al.~\cite{milan2014continuous} 
} \label{table4}
\centering
\begin{tabular}{|c@{\hspace{\hv}}|ccccccccccccc|}
  \hline
  Tracker&RCLL(\%)$\uparrow$ &PRCN(\%)$\uparrow$&FAF$\downarrow$&GT&MT$\uparrow$&PT &ML$\downarrow$&FP$\downarrow$&FN$\downarrow$&IDS$\downarrow$&FM$\downarrow$&MOTA$\uparrow$&MOTP$\uparrow$
  \\
 \hline
 MHT~\cite{kim2015multiple}&-&-&2.10&42&8&31&3&933&3667	&\textbf{142}&\textbf{201}&50.8 &70.4\\
 AMIR~\cite{Amir2017}&-	&-	&\textbf{1.40}	&42	&5&33&4&\textbf{616} &4236 &254&397&47.0&\textbf{70.5} \\
 TBC3~(ours)&71.8&83.7&3.23&42&15&26&1&1409 &2701 &558 &467&51.4&65.8 \\
 TBC~(ours)&78.5&85.3&2.99&42&\textbf{19}&23&\textbf{0}&1303&\textbf{2072}&531&562&\textbf{59.5}&67.2 \\
  \hline
Milan et al.~\cite{milan2014continuous}&65.5	&89.8	&1.43 &74&28&34&12	&622&2881&{99}&{73}&56.9&59.4 \\
Berclaz et al.~\cite{berclaz2011multiple}&26.8&92.1&\textbf{0.44}&74&7&27&40&\textbf{193}&6117&\textbf{22}&\textbf{38}&24.2&60.9 \\
Andriyenko et al.~\cite{Andriyenko2011}&53.9&93.7	&0.69	&74	&15&45&14	&301&3850&152&128&48.5&62.0\\\
 Andriyenko et al.~\cite{Andriyenko2012}&52.6&{94.7}&0.56	&74	&15&48&11	&245&3957&143&125&48.0&61.6 \\
Pirsiavash et al.~\cite{Pirsiavash2011}&49.0&\textbf{95.4}&0.46	&74	&7&50&17&199&4257&137&216&45.0&64.1 \\
 Wen et al.~\cite{wen2014multiple}&71.2	&90.3	&1.47	&74	&27&44&3&640 &2402&125&175&{62.1}&52.7\\
 Wen et al.~\cite{wen2016exploiting}&74.4	&89.8	&1.62 	&74	&30	&42&2&708&2141&136&235&\textbf{64.2}&57.3\\
TBC3 (ours)&75.6&83.1&2.94&74&32&40&2&1281&2036&327&397&56.4&\textbf{69.5} \\
TBC(ours)&\textbf{78.4}&81.9&3.32&74&\textbf{40}&33&\textbf{1}&1448&\textbf{1805}&300&352&57.5&66.6 \\

\hline
\end{tabular}
\end{table*}

\begin{table*}[htbp!]
\setlength{\hv}{0.10cm}
\renewcommand\arraystretch{1.25}
\caption{Effects of different parameters on \emph{UCSD}.} \label{table8}
\centering
\scalebox{0.85}{
\begin{tabular}{|c|c@{\hspace{\hv}}|cccccccccccccc|}
  \hline
  Parameter&Setting&RCLL(\%)$\uparrow$ &PRCN(\%)$\uparrow$&FAF$\downarrow$&GT &MT$\uparrow$&PT&ML$\downarrow$&FP$\downarrow$&FN$\downarrow$&IDS$\downarrow$&FM$\downarrow$&MOTA$\uparrow$&IDF1$\downarrow$&MOTP$\uparrow$ \\

\hline
\multirow{4}{*}{Edge Cost}
 &Location
 &\textbf{52.7}&77.4&6.75&62&\textbf{13}&37&12&1351&\textbf{4150}&268&518& 34.3&37.6 &73.5\\
 
&Appearance
&52.3&77.2&6.80& 62&12&38&12 &1360&4184&388&580&32.4&36.1 &73.8\\

&Motion
&51.7&76.1&7.14&62&10&39&13&1428&4242&975& 663&24.3&29.4 & 73.6\\
&All
&49.5& \textbf{82.1}&\textbf{4.74}&62&{11}& 37&{14}&\textbf{948}&{4435}&\textbf{140}&\textbf{389}& \textbf{37.1}& \textbf{38.7}&\textbf{74.0}   \\
\hline

\multirow{5}{*}{Track Cost}
&$c_{si}$=0
&44.7&\textbf{85.4}&\textbf{3.35}&62&6 &38&18&\textbf{671}&4852&\textbf{127}&\textbf{267}&35.6&35.6&\textbf{74.5} \\

&$c_{si}$=5
&48.7 &81.8 &4.77& 62&10&34&18& 954&4504&145 &373&36.2& 38.3&74.1 \\

&$c_{si}$=10
&49.5& {82.1}&4.74&62&\textbf{11}& 37&\textbf{14}&948&\textbf{4435}&140&389& \textbf{37.1}& \textbf{38.7}&{74.0}  \\

&$c_{si}$=15
&48.7 & 82.2 & 4.64&62&9 &35 &18&929&4501&141&379&36.3& 38.5 &74.1 \\

&$c_{si}$=20
&48.5&82.5&4.51& 62 & 9& 35 &18& 902&4520&133&373&35.3& \textbf{38.7}&74.1 \\
\hline

\multirow{5}{*}{Window size}
& w=$\frac{1}{3}w_0$
&47.1&79.7&5.28&62&5&37&20&1055&4646&156&393&33.3&34.1&73.8 \\

& w=$\frac{1}{2}w_0$
&48.7&81.5& 4.84& 62&10&34&18&968&4503&141&388&36.1&37.9&74.0\\

& w = $w_0$
&49.5& {82.1}&4.74&62&\textbf{11}& 37&\textbf{14}&948&\textbf{4435}&140&389& \textbf{37.1}& \textbf{38.7}&{74.0}\\

& w = $2w_0$
&45.2&81.8&4.41&62& 8& 34&20&881&4812&154&378 &33.4&33.9&\textbf{74.2}\\

& w = $3w_0$
&42.9&\textbf{82.6}&\textbf{3.98}&62&3&37&22&\textbf{796}&5011&\textbf{129}&\textbf{339}&32.4&32.2&74.1 \\

\hline
\end{tabular}}
\end{table*}

The \emph{Fish} and \emph{Cell} datasets
both feature crowded scenes with small objects that deform or change appearance,
which lead to the failures of DPM and Faster-RCNN.  We thus only use the IP detector to generate detection results for evaluating all the competing methods.
Quantitative results on \emph{Fish} and \emph{Cell} are summarized in Table \ref{table3}. 
Both DCEM and MHT assume that object trajectories are smooth and continuous, and adopt a motion model to recover the potential missed detections that follow the trajectory assumption. Although this recovery strategy works well for people tracking, it violates the fact that {objects} like fish and cell usually change their moving directions randomly and suddenly. 
This is the reason why DCEM and MHT have high FPs which in turn results in low MOTAs. In terms of MOTA, the flow-based method GOGA performs better than DCEM and MHT (e.g., for \emph{cell}, MOTA of 73.9 vs 34.1 and 49.1), 
but worse than the TBC3 model. Our TBC model achieve the best results among all the methods by simultaneously optimizing detection and tracking.

\subsection{Evaluation on \emph{PETS2009}}

The \emph{PETS2009} S2-L2 sequence 
features many intersecting trajectories, making it challenging for multi-object tracking. Table \ref{table4} reports the tracking results
on~\emph{PETS2009}. Here, we directly use the reported results from MHT and AMIR on S2L2 (see Tab. \ref{table4} (top)),
 and our result is also evaluated through the MOT server. TBC achieves higher MOTA than MHT and AMIR (59.5 vs 50.8 and 47.0) 
 on S2L2 (Note that AMIR ranks third on MOT15). We also compare with the results from a recent paper \cite{wen2016exploiting} on an ROI of S2L2 (using ground truth from Milan et al.~\cite{milan2014continuous}). Even without using the provided detection results, our model can achieve comparable results with these baselines, but is able to find more tracks (higher MT and Recall). Note that the number of GT trajectories are different due to using different annotations.

\begin{figure*}[!hbtp]
\centering
\captionsetup[subfigure]{labelformat=empty}
\subfloat{\includegraphics[width=\widthfour]{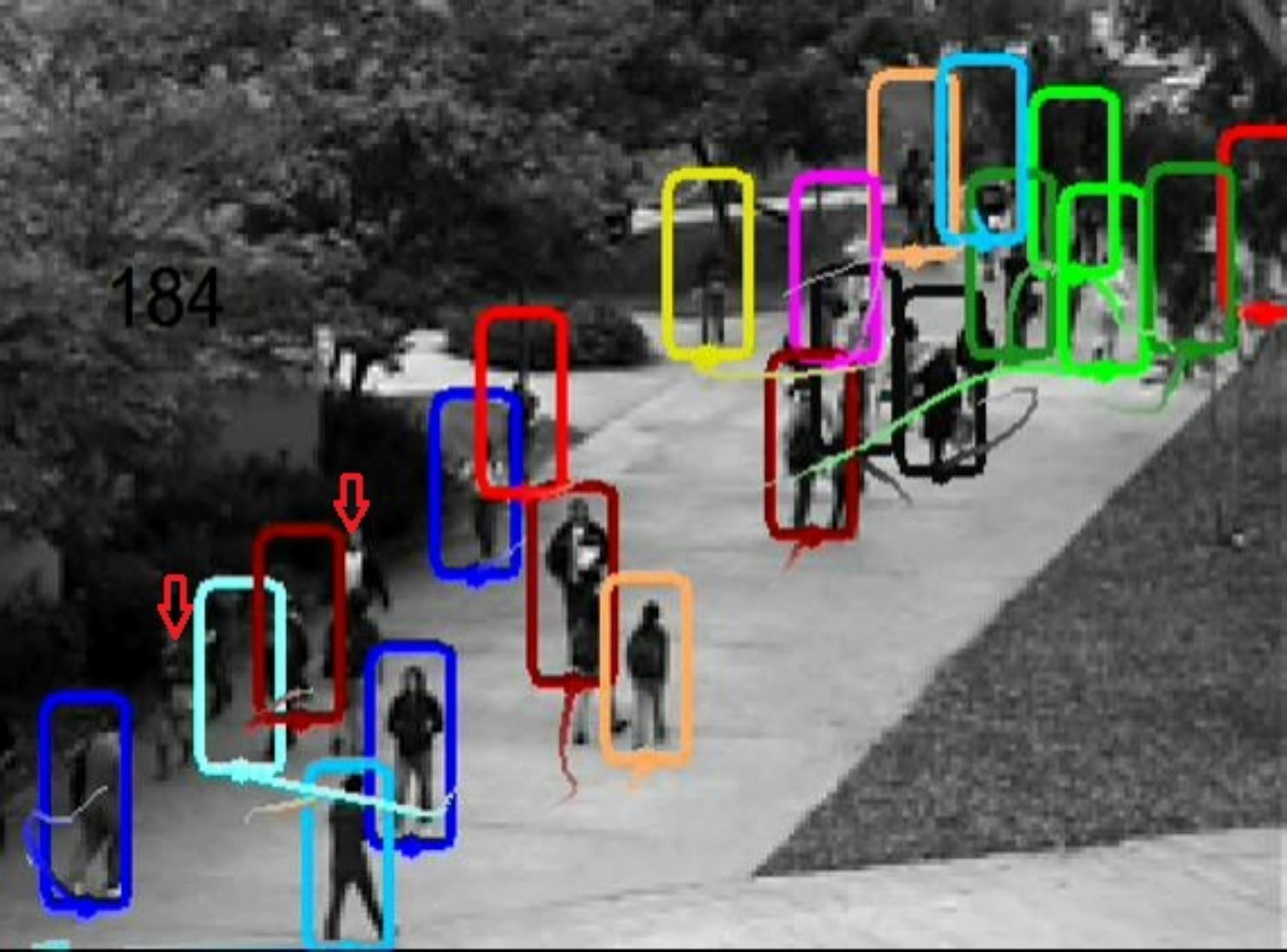}} \hspacefigure
\subfloat{\includegraphics[width=\widthfour]{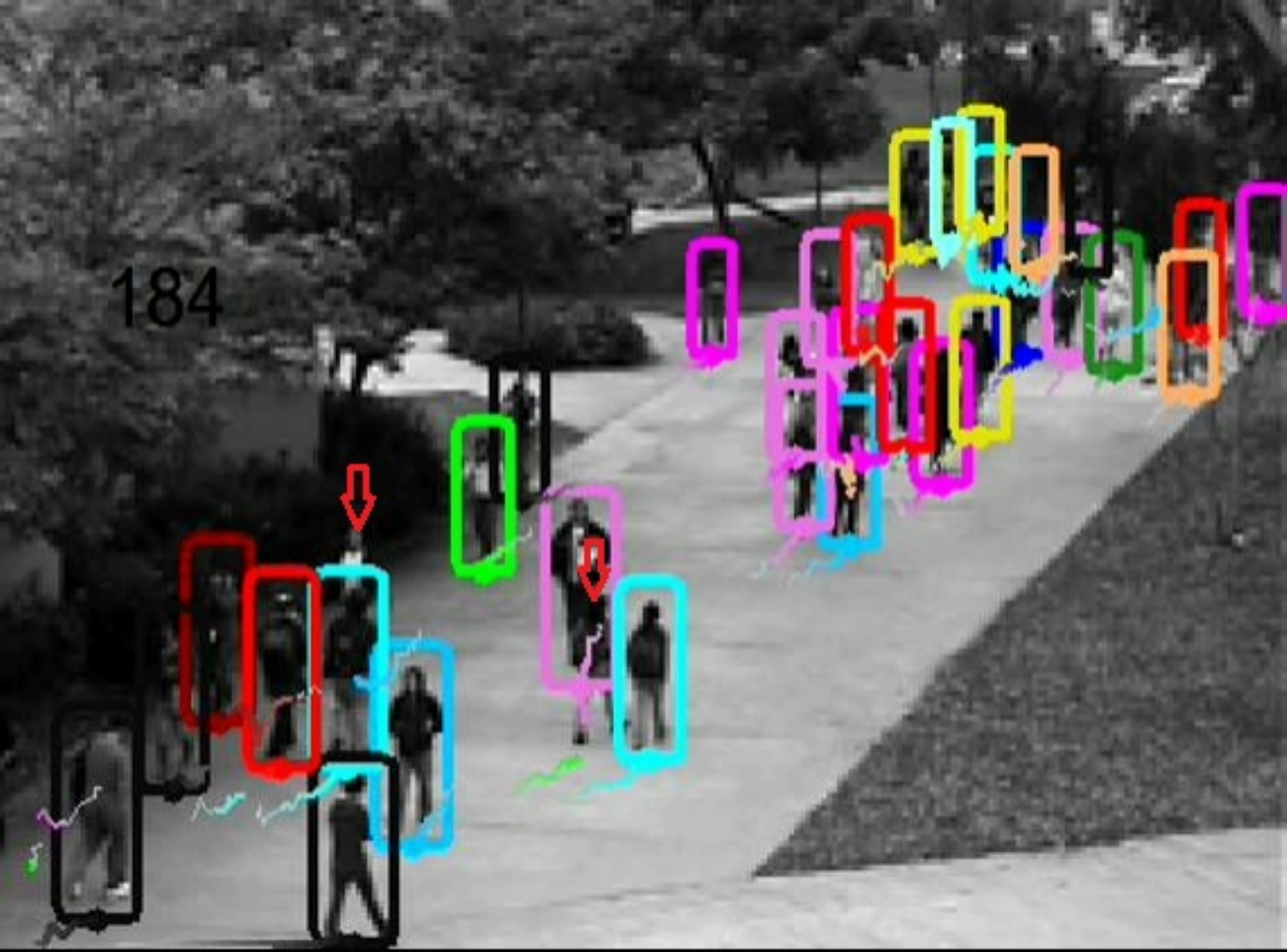}} \hspacefigure
\subfloat{\includegraphics[width=\widthfour]{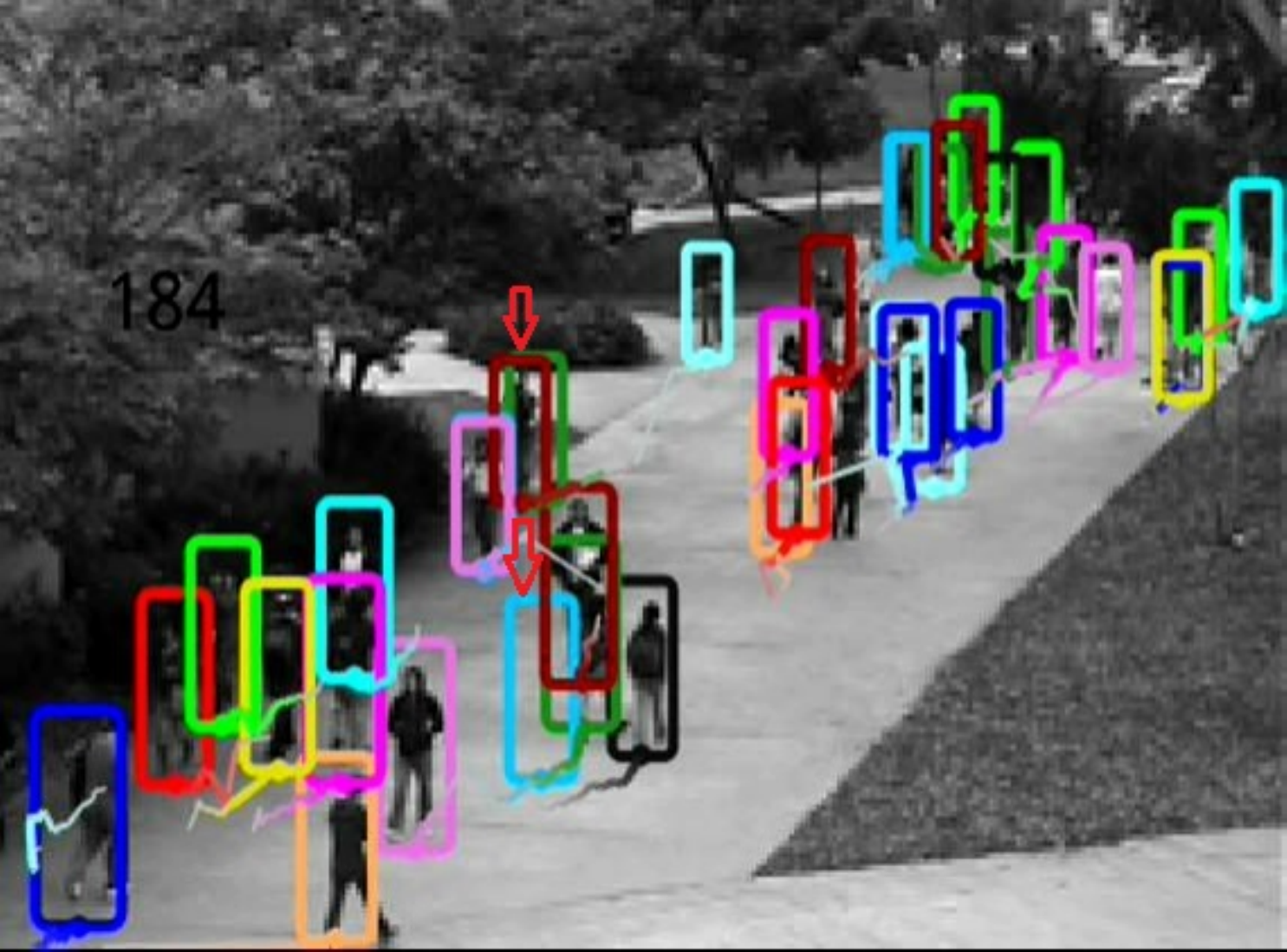}} \hspacefigure
\subfloat{\includegraphics[width=\widthfour]{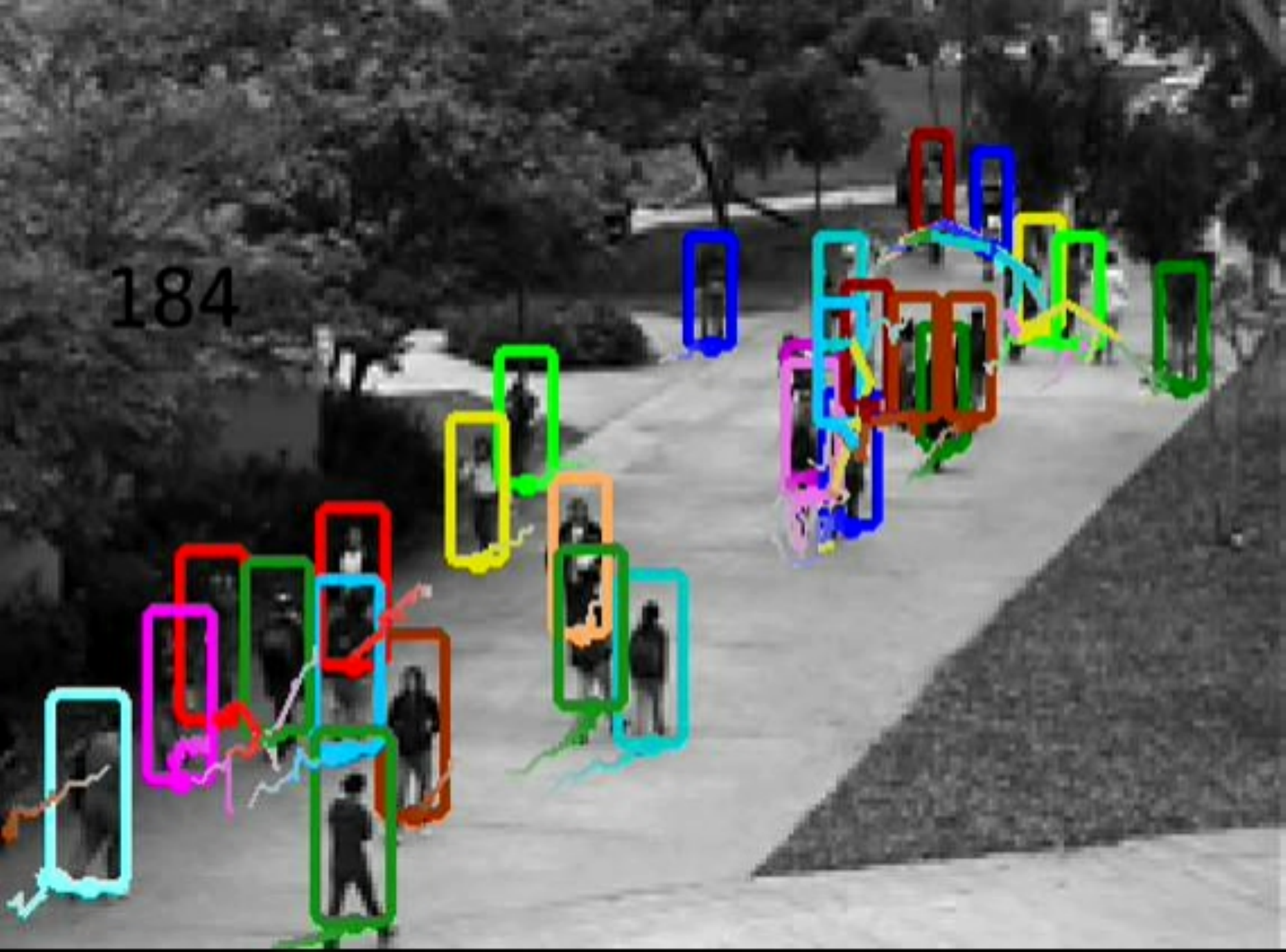}} \hspacefigure \\
\vspace{-0.10in}

\subfloat[DCEM-faster]{\includegraphics[width=\widthfour]{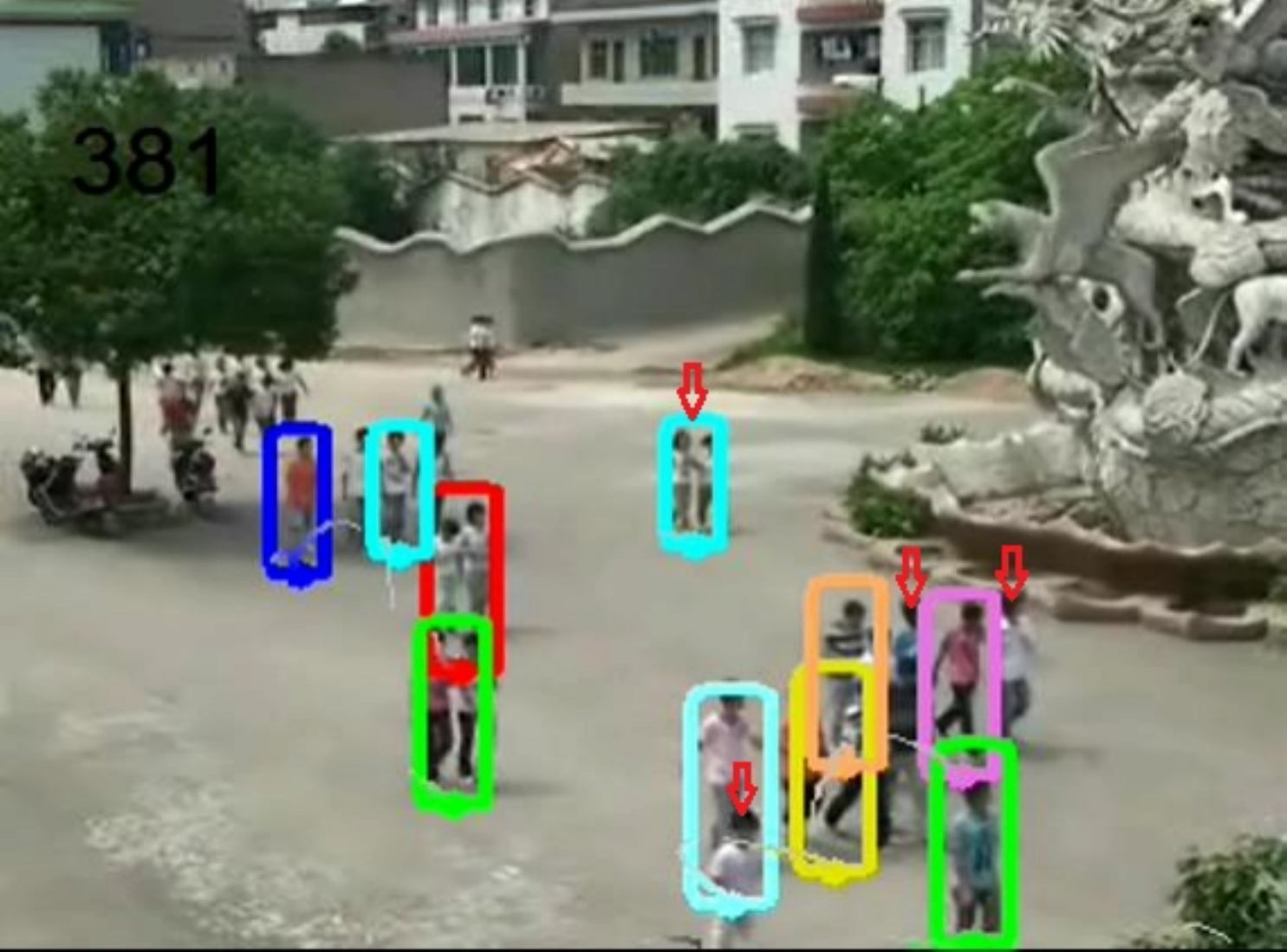}} \hspacefigure
\subfloat[GOGA-faster]{\includegraphics[width=\widthfour]{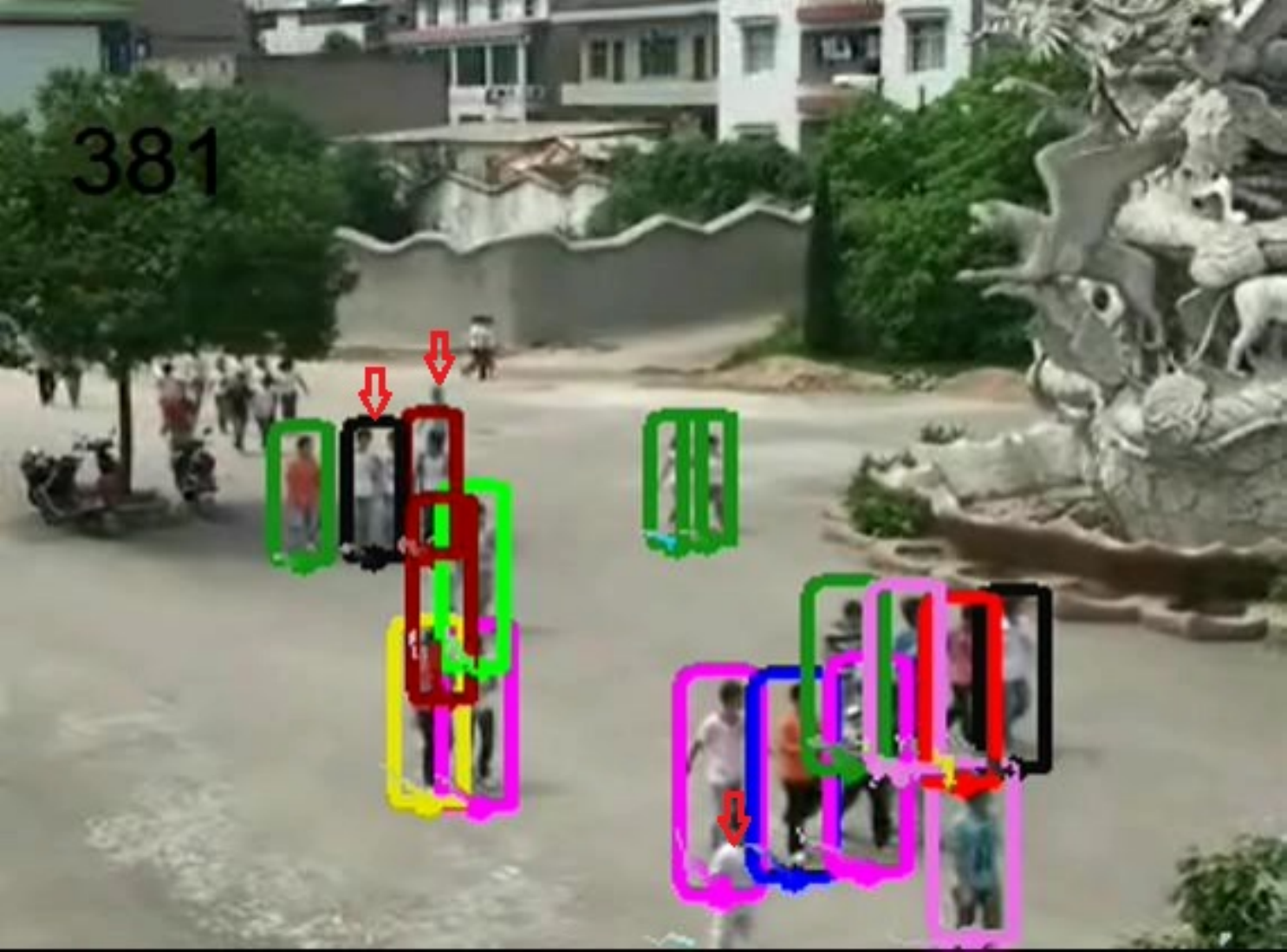}} \hspacefigure
\subfloat[MHT-faster]{\includegraphics[width=\widthfour]{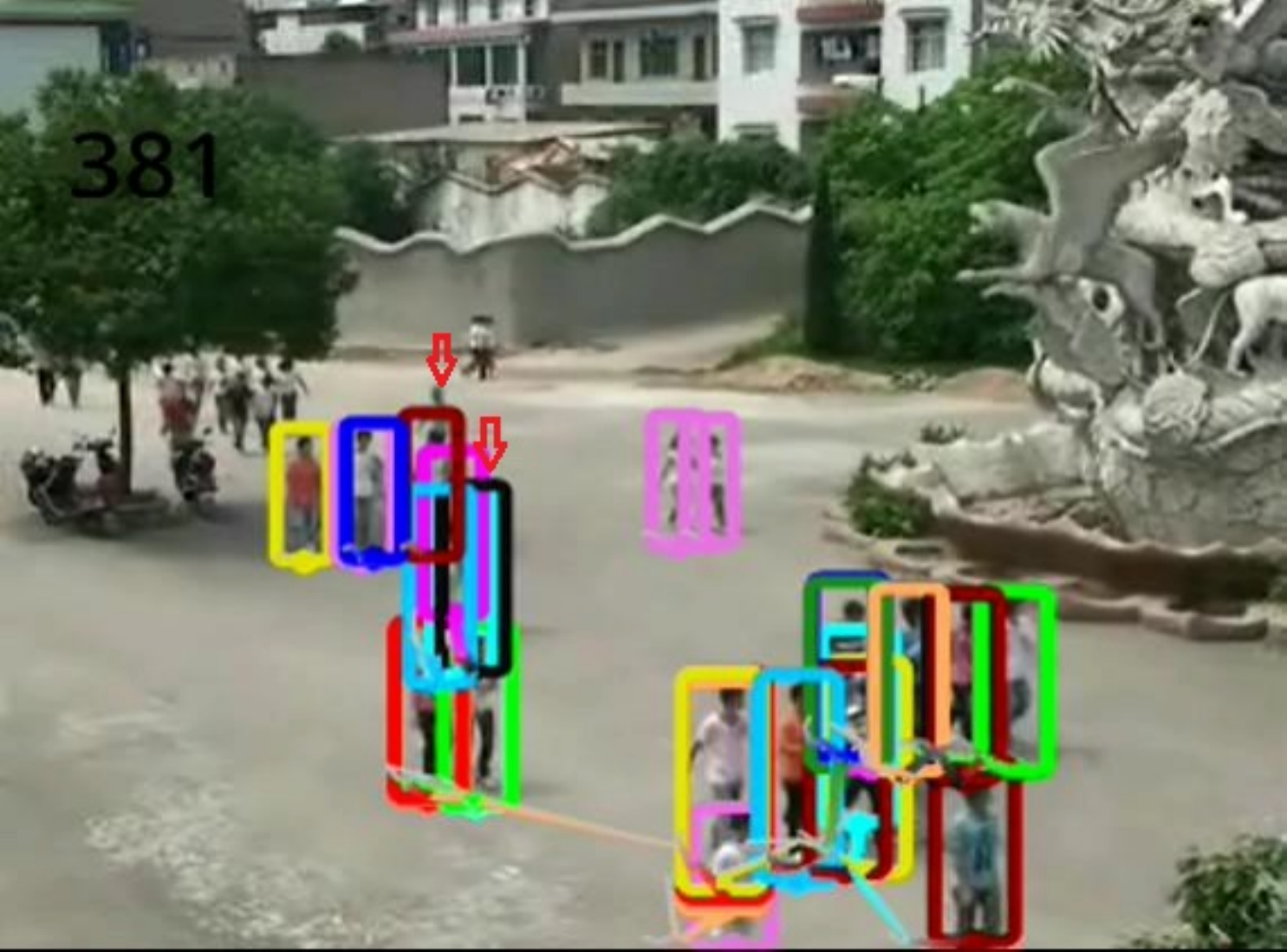}} \hspacefigure
\subfloat[TBC (ours)]{\includegraphics[width=\widthfour]{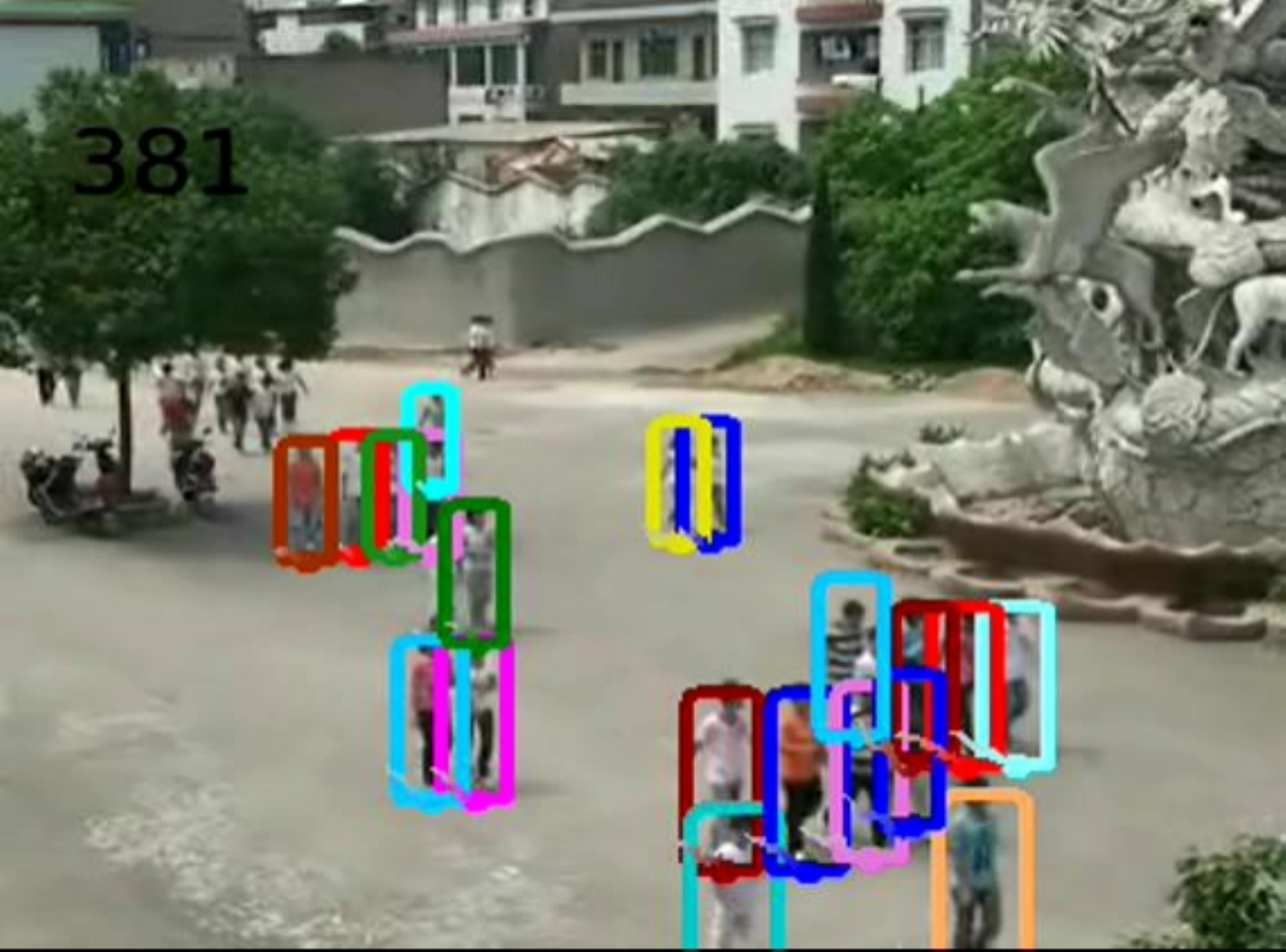}} \hspacefigure \\
\vspace{-0.10in}

\subfloat{\includegraphics[width=\widthfour]{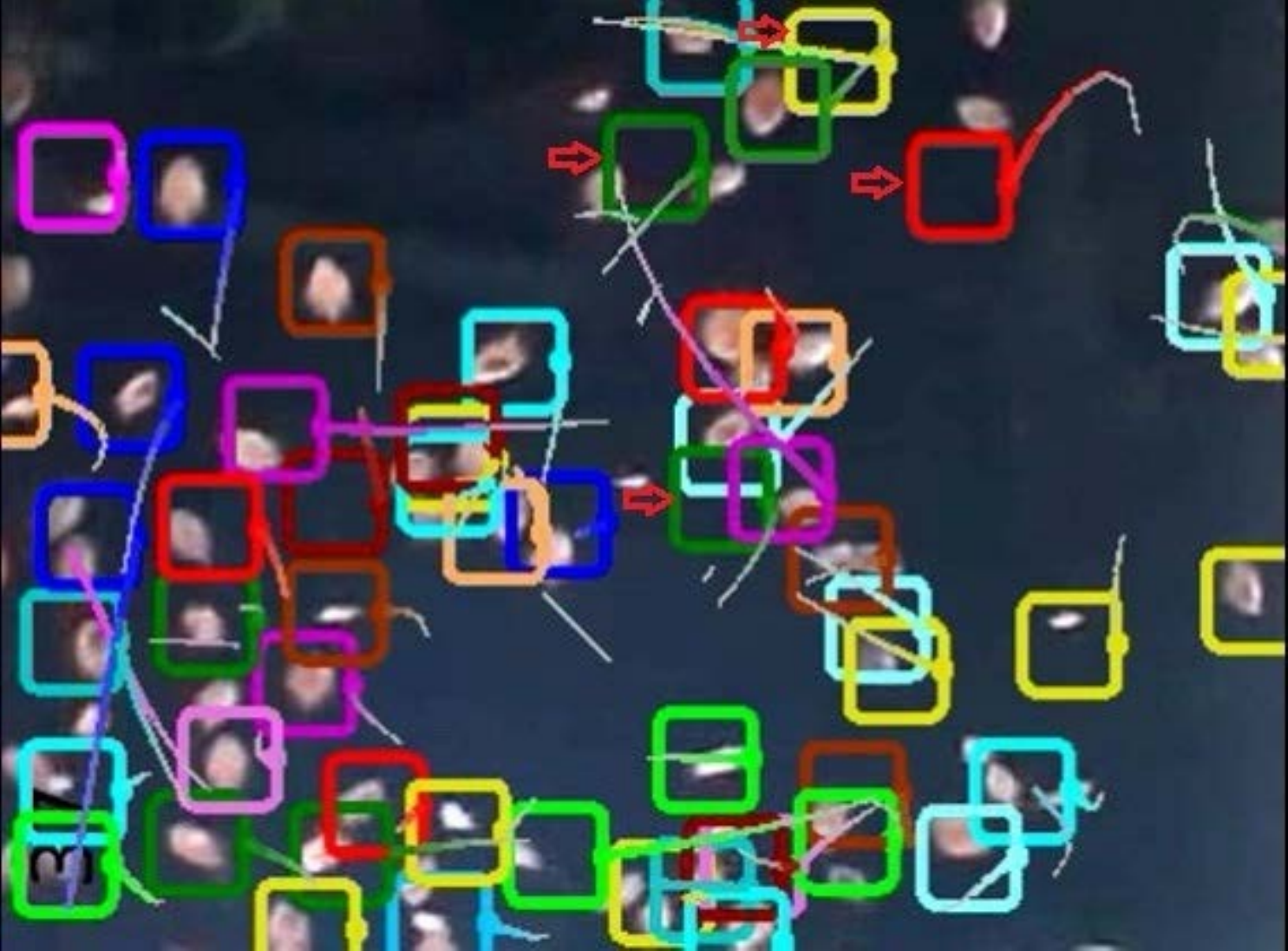}} \hspacefigure
\subfloat{\includegraphics[width=\widthfour]{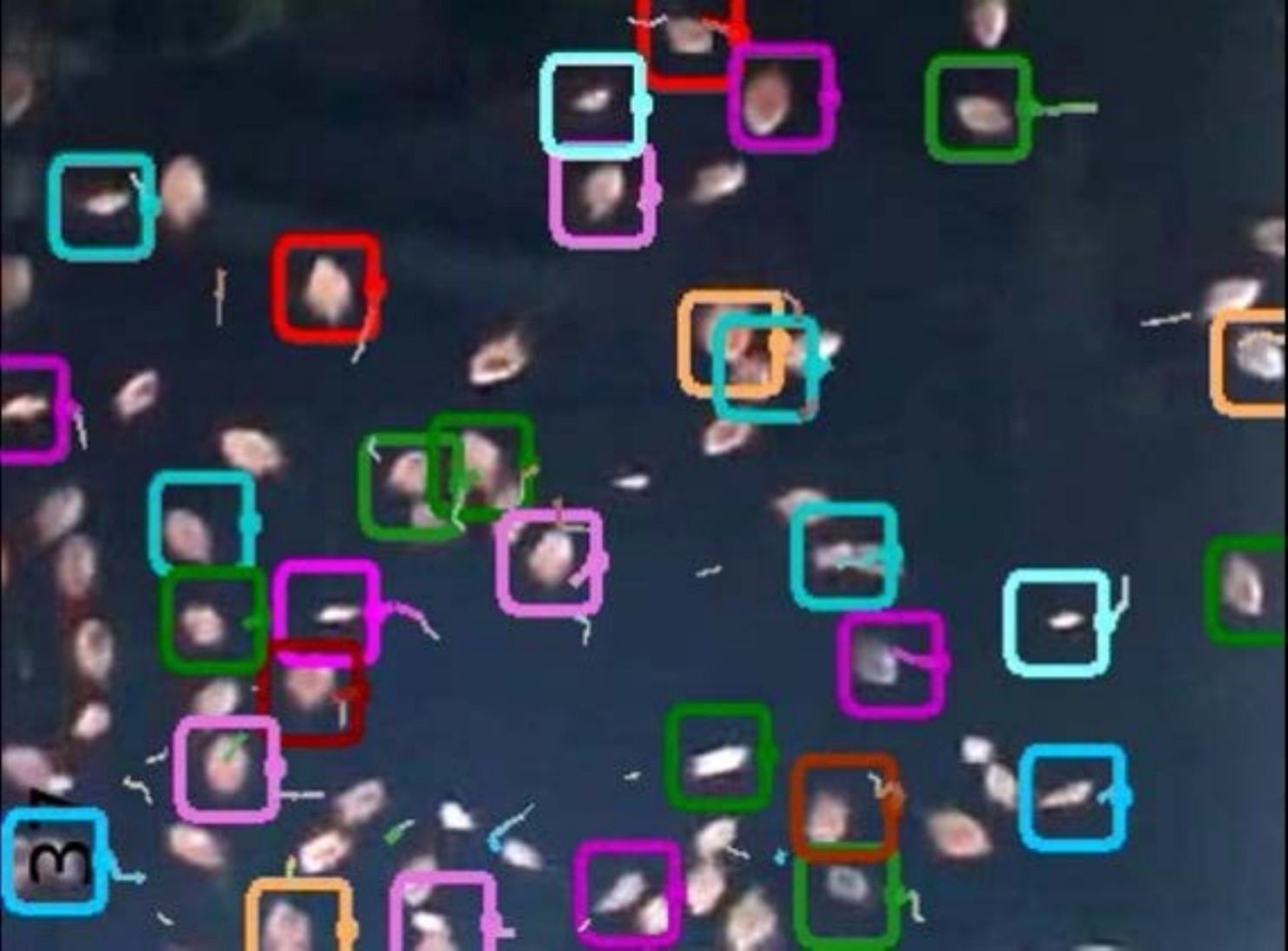}} \hspacefigure
\subfloat{\includegraphics[width=\widthfour]{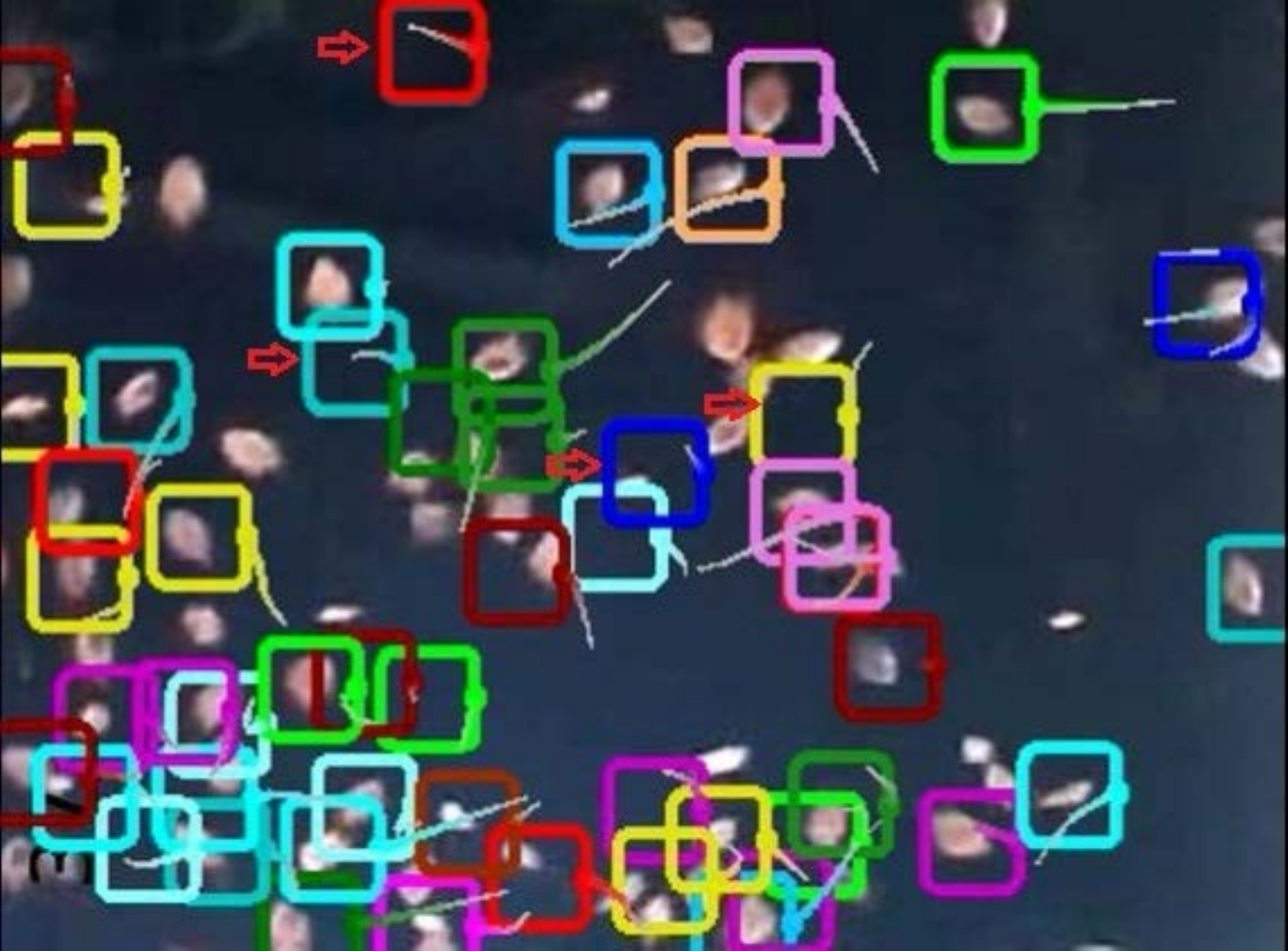}} \hspacefigure
\subfloat{\includegraphics[width=\widthfour]{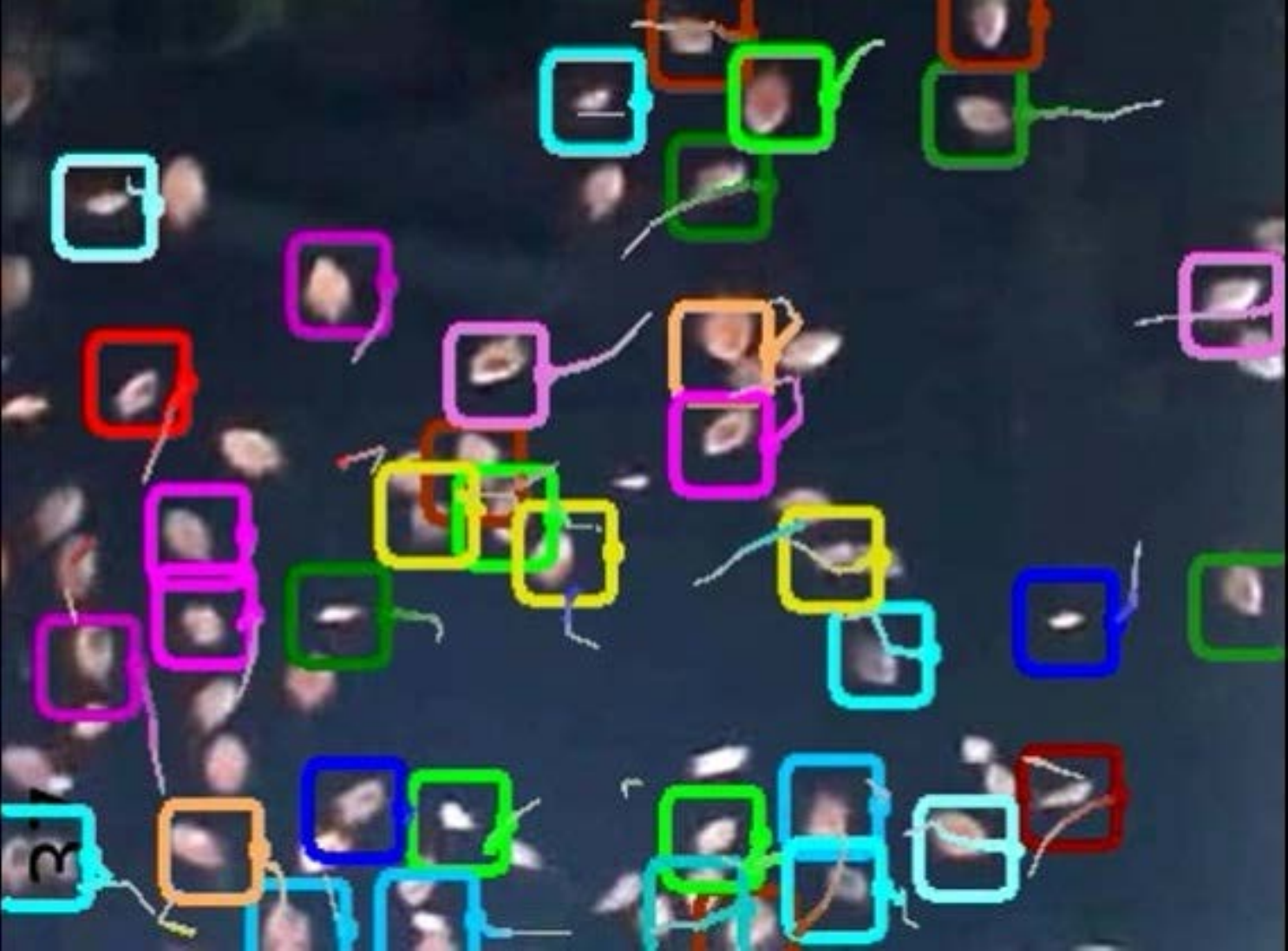}} \hspacefigure \\
\vspace{-0.10in}

\subfloat[DCEM-IP]{\includegraphics[width=\widthfour]{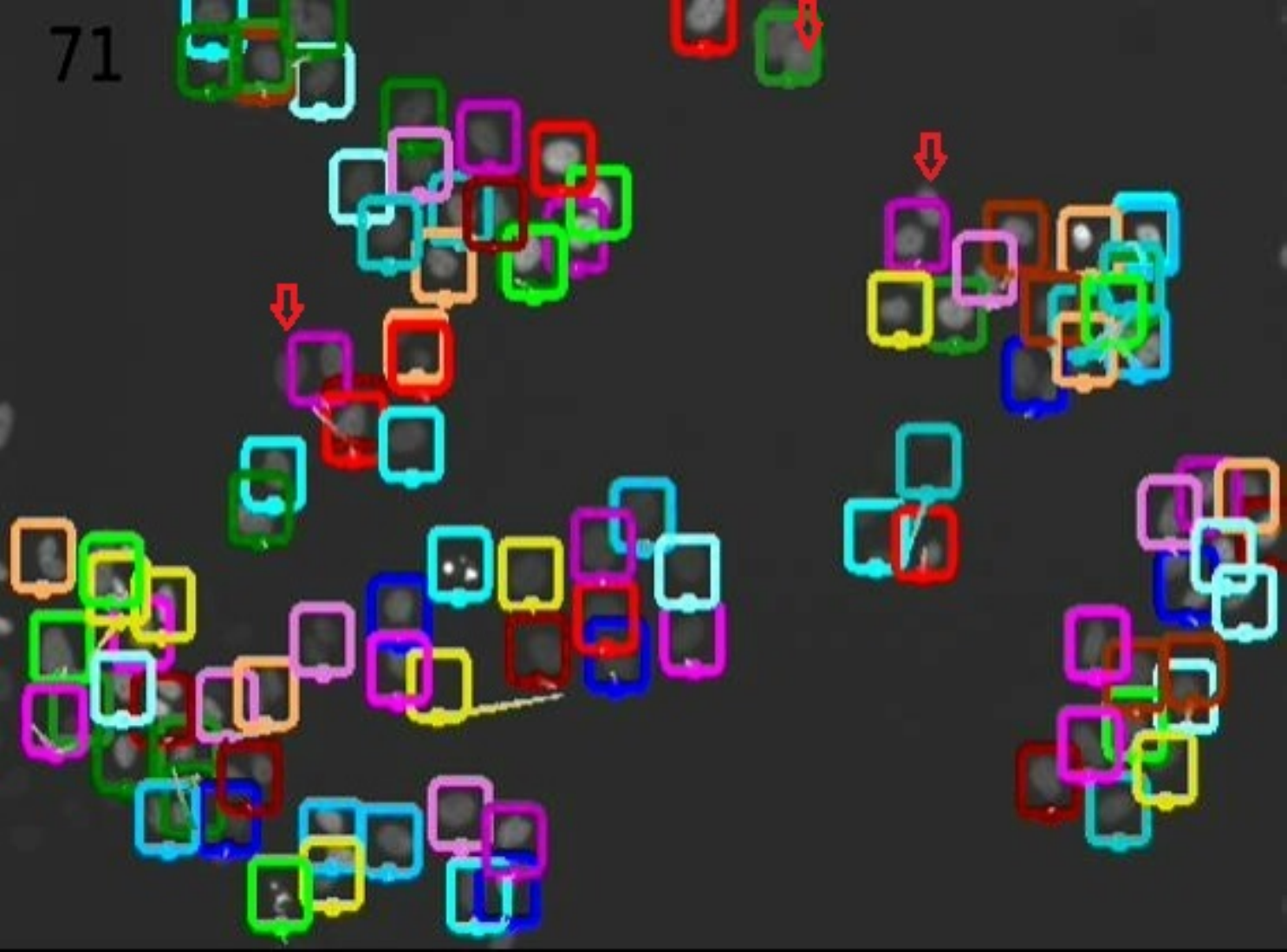}} \hspacefigure
\subfloat[GOGA-IP]{\includegraphics[width=\widthfour]{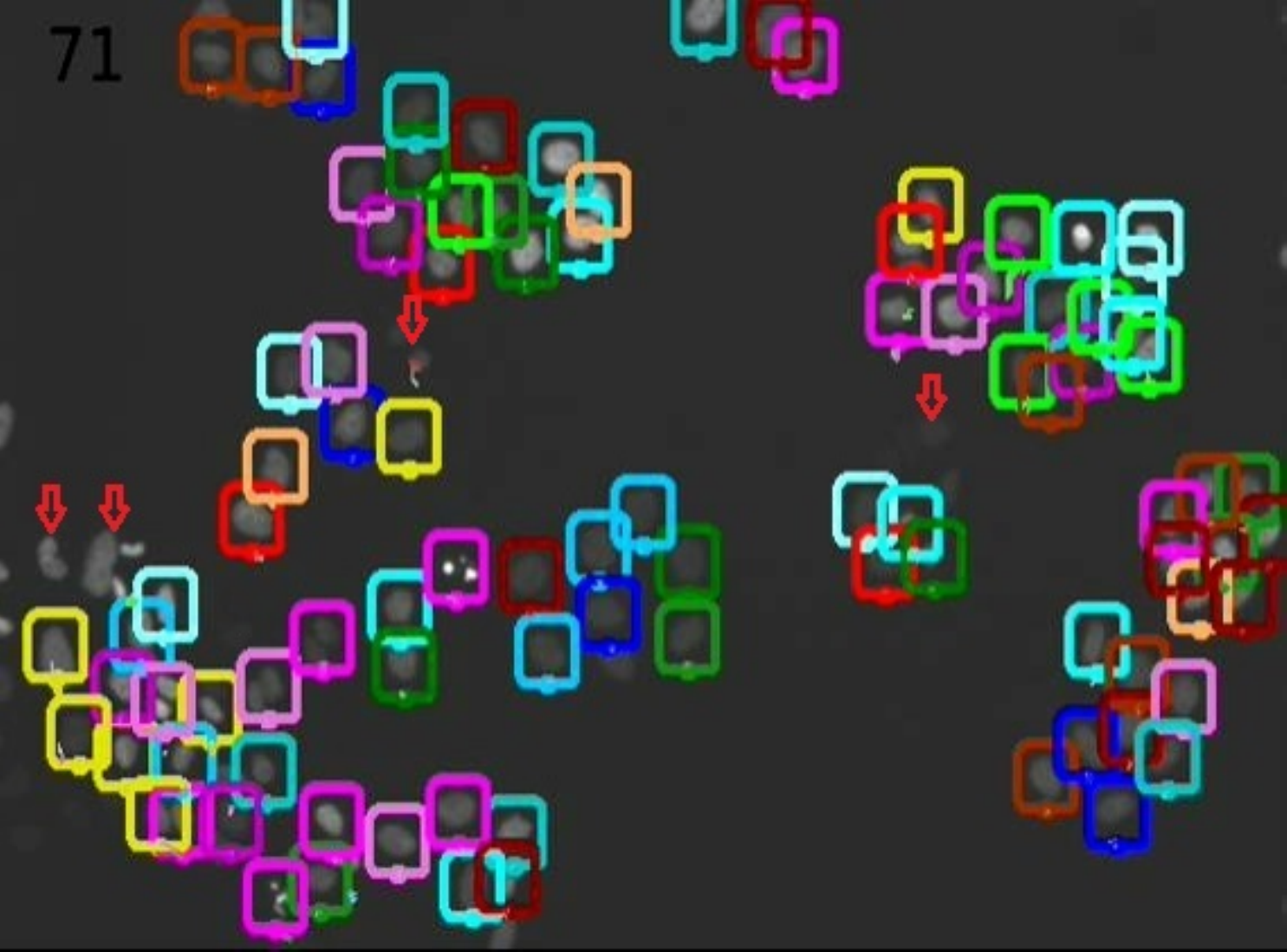}} \hspacefigure
\subfloat[MHT-IP]{\includegraphics[width=\widthfour]{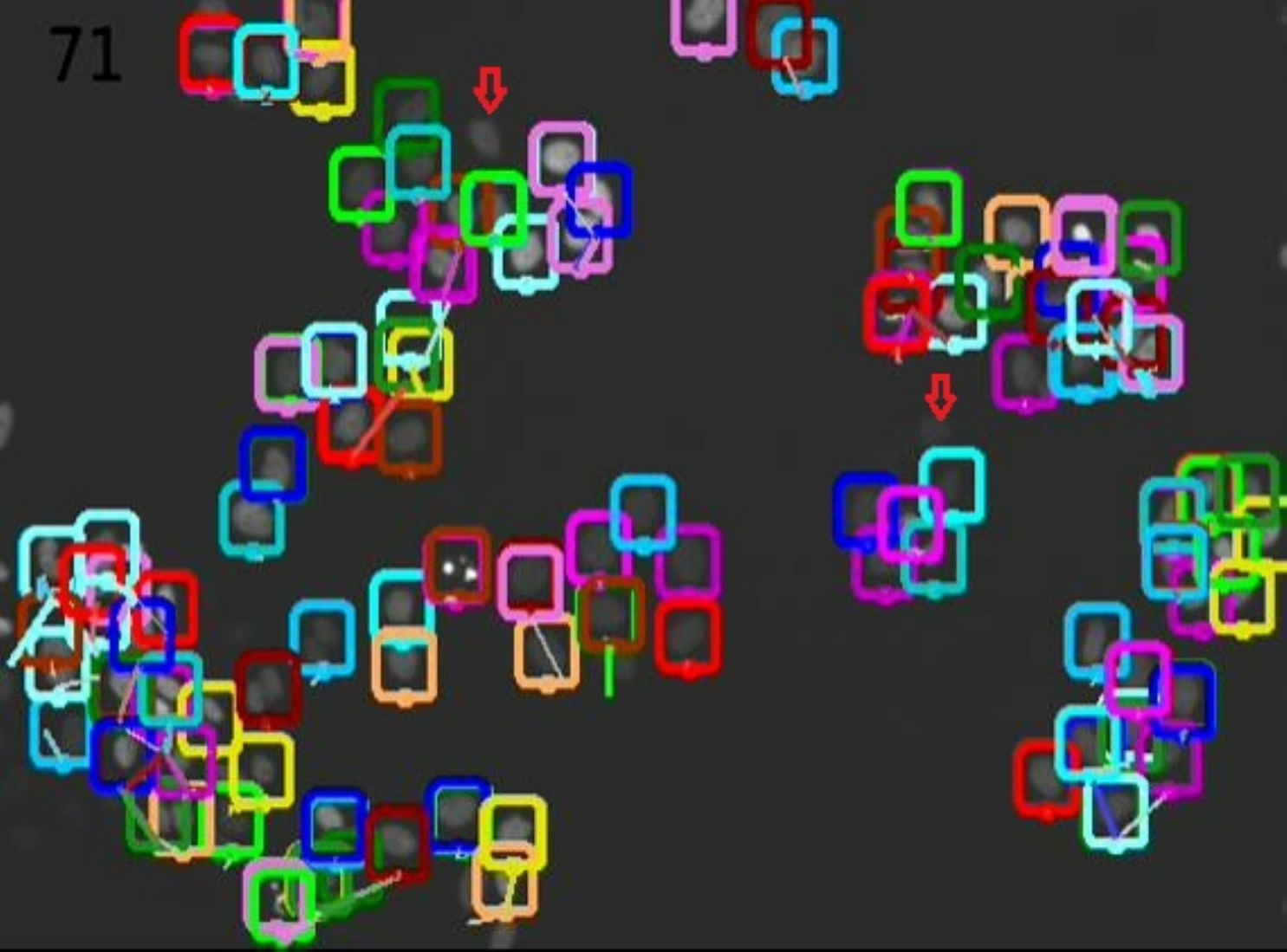}} \hspacefigure
\subfloat[TBC (ours)]{\includegraphics[width=\widthfour]{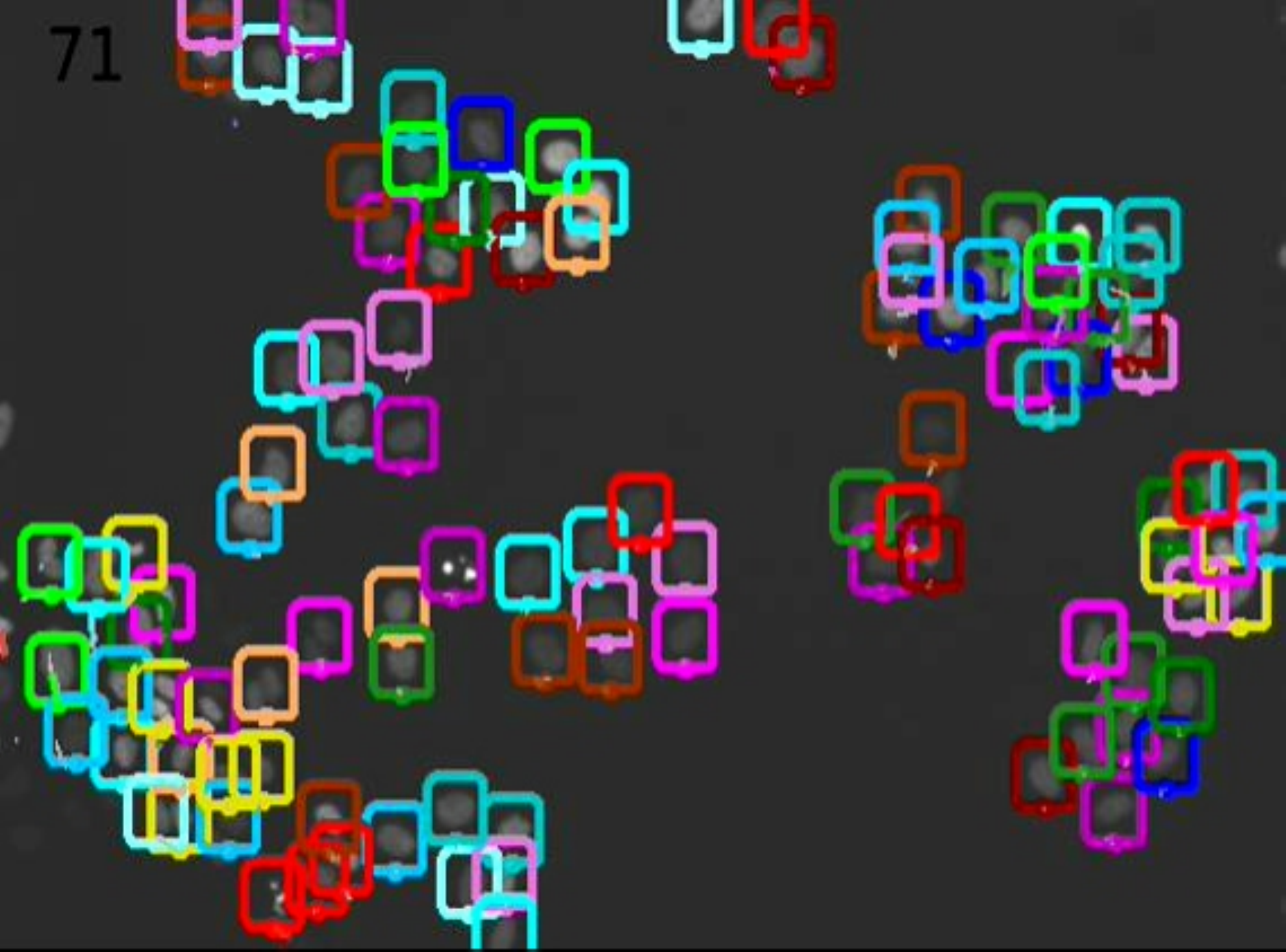}} \hspacefigure \\

\caption{Qualitative results of our tracking-by-counting model. From top to bottom, the results are on \emph{UCSD}, \emph{LHI}, \emph{Fish} and \emph{Cell}, respectively. }\label{experiment}
\end{figure*}

{The top two rows of Fig.~\ref{experiment} show the tracking results of three {competing} trackers DCEM, GOGA, MHT and our TBC using detections generated by Faster-rcnn. Though the retrained Faster-rcnn can recognize most of the people in the scene,  {it still} misses some objects (marked with red arrows) when they are heavily occluded, which suggests that crowd density maps are very useful for object detection in
crowded scenes. In the last two rows of Fig.~\ref{experiment}, we show the tracking results of \emph{Fish} and \emph{Cell}, both of which have many intersecting paths. For \emph{Fish}, DCEM and MHT generate many FPs. {GOGA  achieves} good tracking results but also  discards many {true positives}. Using global optimization of object detection and data association, our TBC model  achieves less IDS and FM. Consistent with \emph{Fish}, the TBC model also works the best among all the trackers. {The video results can be found in our supplementary material.}}
{Finally, the running time of TBC is affected by the image resolution and density maps used for building a graph. 
The average per-frame running times for \emph{UCSD}, \emph{LHI}, \emph{Fish}, \emph{Cell} and \emph{PETS2009} are 2.4, 4.1, 12.9, 16.3 and 6.4 sec, respectively.} \wh{For the reported running time, it is only for solving the energy function in Eq. (5). Note that the time is for TBC which optimizes detection and tracking over all the whole sequence. For TBC3 (using 3 frames), the average time (for all the scenes) can be reduced to within 0.03s. The tracking speed is directly correlated with the number of sliding windows, which are then determined by the number of the targets and the number of optimized frames. Taking \emph{UCSD} as an example, a crowd scene (average 38 people) with 200 frames has average running time 2.4s, while a sparse scene (average 17 people) with the same frame number has average running time 0.6s. Also, for \emph{LHI} (average 22 people) with 400 frames, the running time can reach 4.1s.}

\subsection{{Effects of parameter settings on tracking performance}}
In this subsection, we analyze the effects of parameter settings on tracking performance. The edge cost $c_{ij}$ in~(\ref{eq7}) consists of three types of input information, and we first analyze the effect of each one by removing the other two. In addition,  the track start cost $c_{si}$ and terminate cost $c_{it}$ also affect the tracking performance, and we test them by using different settings. Finally,  we analyze the performance of TBC model by using different sliding window sizes in~(\ref{eq6}).

All the experiments are carried on \emph{UCSD} using TBC model, and the results are reported in Tab.~\ref{table8}. The edge cost $c_{ij}$ incorporates geometric location, target appearance and motion information for association, and each of them  contributes to the tracking performance. Large track cost $c_{si}$ or $c_{it}$ reduces IDS and thus can improve IDF1, but this will degrade the tracking accuracy. E.g., using $c_{si}=20$ achieves less IDS compared with using $c_{si}=10$ (133 vs.~140), but has lower MOTA (35.3 vs.~37.1). 

The base sliding window is set as the average target size (i.e., $w_0$).
With the base window, the sliding windows used for detections will be adjusted by perspective map. A window with
either too large spatial size or too small spatial size will degrade the detection results, and further affects the tracking performance. For a small spatial window, the sum within the window is far less than 1, resulting in missed detection. For a large window, the window will contain too many objects that are hard to be accurately localized. As observed, when we shrink and expand the base window
to 1/3 and 3 times of its original size, the MOTA drops
from 37.1 to 33.3 and 32.4 respectively.
{On the other hand, the MOTP is not affected much by the base window size.}

\begin{figure}[hbtp!]
\centering
\includegraphics[width=0.95\linewidth]{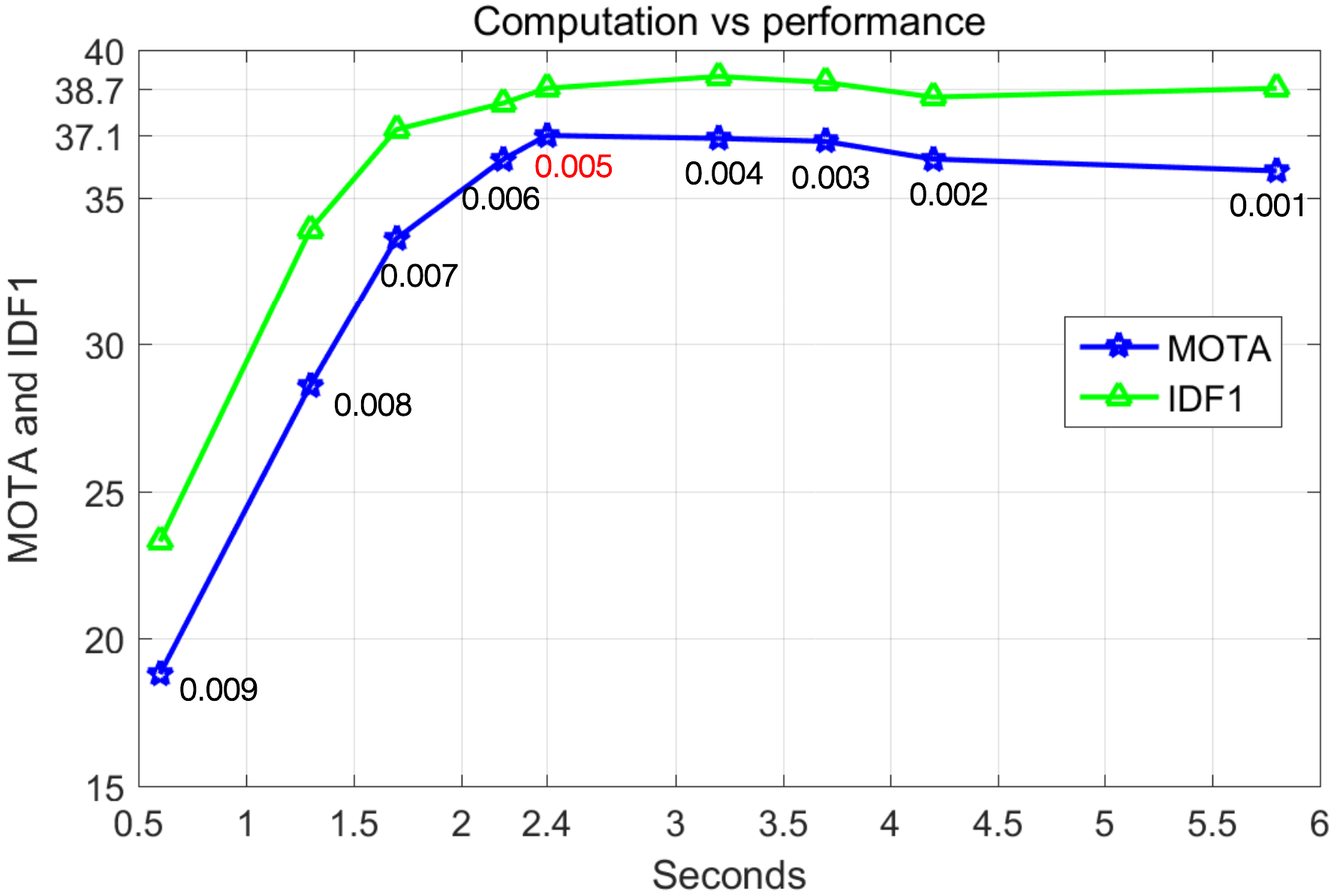}
\caption{Plot of computation vs. performance for different MILP problem sizes (i.e., different threshold settings).}\label{fig:mota}
\vspace{-0.2in}
\end{figure}

To reduce computational complexity, we dropped the candidate points with low density value ($<$0.005) when solving~(\ref{eq6}). Here, we draw a plot of computation vs. performance for different MILP problem sizes (i.e., different threshold settings). From 0.001 to 0.009, we choose 9 thresholds with step size 0.001 to separately evaluate TBC model on \emph{UCSD}. As shown in Fig.~\ref{fig:mota}, when the threshold is set to 0.005 (marked red in the plot), a good balance between computation and tracking performance is achieved (MOTA=37.1, IDF1=38.7 and time = 2.4s). Besides, the threshold 0.005 also works well for other datasets.

\section{Evaluation on large scale datasets}
\label{text:largescale}
Our previous experiments are conducted on crowd scenes with relatively low resolution, where the traditional object detector often fails to localize objects. Using crowd density maps, our \emph{tracking-by-counting} model has advantage of detecting and tracking small objects over other methods in these scenes. For large scale datasets {with high-resolution}, the traditional object detector can achieve good detection results. To further evaluate our model on large scale scenes, we extend our \emph{tracking-by-counting} model, by incorporating detection results produced by object detectors:
\begin{equation}\label{eq9}
\begin{aligned}
  \underset{\textbf{x}}{\mathop{\min }}\,  &\textbf{x}^{T}\textbf{H}\textbf{x} + \textbf{m}^T\textbf{x} + \sum\limits_{t=1}^{T}\sum\limits_{k=1}^{K_t}{\left| \left(\textbf{w}_{k}^t\right)^{T}{\textbf{x}}-{\hat{n}_{k}^t} \right|}+ \sum\limits_{ij\in E}{{{c}_{ij}}{{x}_{ij}}} +\\
 &\sum\limits_{i}{{{c}_{si}}{{x}_{si}}} +\sum\limits_{i}{{{c}_{it}}{{x}_{it}}}\\
 \text{s.t.} &\sum\limits_{i:ij\in E}{{{x}_{ij}}}+x_{sj}={{x}_{j}}=\sum\limits_{i:ji\in E}{{{x}_{ji}}}+x_{jt} \\
 & \sum\limits_{i}{{{x}_{it}}} =\sum\limits_{i}{{{x}_{si}}} , \\
 & x_i, x_{ij} \in\{0,1\},
\end{aligned}
\end{equation}
{where the first term ensures that only non-overlapping locations are selected,} by setting $\textbf{H}_{ij} = \infty$ if locations $i$ and $j$ have significant overlap ratio (\wh{it is set to 0.65}), and $0$ otherwise. The second term is the total detection score, where the vectorized score map \textbf{m} is obtained by placing Gaussian radial basis functions 
at each detected location, with height equal to its negative detection score. By adding the first two terms, our model can combine object detections with estimated density maps together to jointly optimize detections and trajectories. 

Similar to (\ref{eq8}), we also use auxiliary variable  $z_k^t$ to replace ${\left| \left(\textbf{w}_{k}^t\right)^{T}{{\textbf{x}}^{d}}-{{n}_{k}^t} \right|}$, and rewrite (\ref{eq9}) as a quadratically mixed integer least square problem which also can be solved with optimization toolboxes:
\begin{equation}\label{eq10}
\begin{aligned}
   \underset{\textbf{x,z}}{\mathop{\min }}\,&\textbf{x}^{T}\textbf{H}\textbf{x} + \textbf{m}^T\textbf{x} + \sum\limits_{t=1}^{T}\sum\limits_{k=1}^{K_t}{z_k^t}+\sum\limits_{ij\in E}{{{c}_{ij}}{{x}_{ij}}} + \\
&\sum\limits_{i}{{{c}_{si}}{{x}_{si}}} +\sum\limits_{i}{{{c}_{it}}{{x}_{it}}}\\
 \text{s.t.} &\sum\limits_{i:ij\in E}{{{x}_{ij}}}+x_{sj}={{x}_{j}}=\sum\limits_{i:ji\in E}{{{x}_{ji}}}+x_{jt} \\
 & \sum\limits_{i}{{{x}_{it}}}=\sum\limits_{i}{{{x}_{si}}} \\
 &\sum\limits_{t=1}^{T}\sum\limits_{k=1}^{K_t}{    \left(\textbf{w}_{k}^t\right)^{T}{{\textbf{x}}}-{{n}_{k}^t} - z_k^t \leq 0 }\\
 &\sum\limits_{t=1}^{T}\sum\limits_{k=1}^{K_t}{    -\left(\textbf{w}_{k}^t\right)^{T}{{\textbf{x}}}+{{n}_{k}^t} - z_k^t \leq 0 }\\
&x_i, x_{ij} \in\{0,1\}, 0\le z_k^t.
\end{aligned}
\end{equation}

\subsection{Experiments}
\begin{table*}[htbp!]
\setlength{\hv}{0.10cm}
\renewcommand\arraystretch{1.25}
\caption{Multi-object tracking results on \emph{MOT17}.} \label{table5}
\centering
\begin{tabular}{|c|ccccccccccc|}
  \hline
  Tracker&MOTA$\uparrow$&IDF1$\uparrow$ &MOTP$\uparrow$&FAF$\downarrow$&MT$\uparrow$&ML$\downarrow$&FP$\downarrow$&FN$\downarrow$&IDS$\downarrow$&FM$\downarrow$&Time$\uparrow$
  \\
  \hline
JCC~\cite{keuper2018motion}
 &{51.2}&54.5&75.9&1.5&{20.9}\%&37.0\%&25,937&247,822&{1,802}& 2,984&1.8 Hz\\

BLSTM~\cite{kim2018multi}&
 47.5&51.9&\textbf{77.5}&1.5&18.2\% &41.7\% &25,981&268,042&2,069&3,124&1.9 Hz \\

DMAN~\cite{zhu2018online}&
48.2&\textbf{55.7}&75.7&1.5&19.3 \% &38.3\% &26,218&263,608&2,194&5,378&0.3 Hz \\

MHT~\cite{kim2015multiple}&
50.7&47.2&\textbf{77.5}&\textbf{1.3}&20.8\% &{36.9\%}& \textbf{22,875}&252,889&2,314&{2,865}& 0.9 Hz \\
 
SAS\_MOT17~\cite{maksai2019eliminating}& 44.2
 &57.2& 76.4&1.7&16.1\% &44.3\% &29,473&283,611&\textbf{1,529}&\textbf{2,644}&4.8 Hz \\

MTDF17~\cite{fu2019multi}& 49.6
&45.2& 75.5&2.1&18.9\%&\textbf{33.1\%}&37,124	&241,768&5,567&9,260&1.2 Hz\\

Tracktor17~\cite{Phi2019}&53.5&52.3&78.0&0.7&19.5\%&36.6\%&12,201&248,047	&2,072&4,611&1.5 Hz \\	

TBC3 (ours)&  \textbf{53.9}&50.0& 76.8&{1.4}&20.2\% &\textbf{36.7}\% &24,584&\textbf{232,670}&2,945&4,612&6.7 Hz \\
 \hline
 
TBC3-Det& 42.8&41.9& 76.9&1.9&15.4\% &42.1\% &33,670&28,6397&2,815&5,196&10.9 Hz \\

TBC3-Count& 38.3&38.2& 73.1&5.3&17.6\%&38.9\%&9,5178&247,542&5,439&13,101&8.4 Hz \\
 \hline
\end{tabular}
\end{table*}

\begin{table*}[!hbtp]
\setlength{\hv}{0.15cm}
\renewcommand\arraystretch{1.25}
\caption{Multi-object tracking results on \emph{DukeMTMC} dataset.} \label{table6}
\centering
\begin{tabular}{|c|c|cccccccccccc|}
  \hline
{Dataset}& Tracker&IDF1$\uparrow$&IDP$\uparrow$&IDR$\uparrow$&MOTA$\uparrow$&MOTP$\uparrow$&FAF$\downarrow$&MT$\uparrow$&ML$\downarrow$&FP$\downarrow$&FN$\downarrow$&IDS$\downarrow$&FM$\downarrow$ \\
  \hline
\multirow{4}{*}{Test-easy}
&DeepCC~\cite{ristani2018features}
&\textbf{89.2}&\textbf{91.7}&{86.7}&{87.5}&77.1&\textbf{0.05}&1,103&29&\textbf{37,280}&{94,399}&\textbf{202}&\textbf{753} \\
&TAREIDMTMC~\cite{na2018online}
&75.7&77.7&73.8&84.1&73.8&0.08&1,080&\textbf{7}&56,063&110,038&2,068&10,098 \\
&PT\_BIPCC~\cite{and2017globally}
&71.2&84.8&61.4&59.3&\textbf{78.7}&0.09	&{666}&234&68,634&361,589&290&783 \\
&{TBC3 (ours)}
&\textbf{89.2}&90.4&\textbf{88.0}&\textbf{87.8}&78.1& 0.07&\textbf{1,131}& 22& 50,771&\textbf{78,263}&222&833 \\
\hline
\multirow{4}{*}{Test-hard}
&DeepCC~\cite{ristani2018features}
&{79.0}&{87.4}&{72.0}&{70.0}&75.0&0.15&524&66&43,989&170,104&\textbf{236}&\textbf{777} \\
&TAREIDMTMC~\cite{na2018online}
&68.2&74.4&63.0&68.0&72.6&0.20&515&\textbf{38}&57,995&167,625&2,485&9,219 \\
&PT\_BIPCC~\cite{and2017globally}
&65.0&81.8&54.0&54.4&\textbf{77.1}&{0.14}&{335}&104&{40,978}&283,704&661&1,054 \\
&{TBC3 (ours)}
&\textbf{82.4}&\textbf{90.0}&\textbf{76.0}&\textbf{75.2}&76.4& \textbf{0.12}&\textbf{569}&52&\textbf{32,813}&\textbf{143,883}&298&898
 \\
\hline
\end{tabular}
\end{table*}

\begin{table*}[htbp!]
\setlength{\hv}{0.10cm}
\renewcommand\arraystretch{1.25}
\caption{Tracking results on \emph{UCSD}, \emph{LHI} and \emph{S2L2} when adding detections.} \label{table7}
\centering
\scalebox{0.85}{
\begin{tabular}{|c|c@{\hspace{\hv}}c|cccccccccccccc|}
  \hline
  Dataset&Tracker&Detector&RCLL(\%)$\uparrow$ &PRCN(\%)$\uparrow$&FAF$\downarrow$&GT &MT$\uparrow$&PT&ML$\downarrow$&FP$\downarrow$&FN$\downarrow$&IDS$\downarrow$&FM$\downarrow$&MOTA$\uparrow$&IDF1$\downarrow$&MOTP$\uparrow$ \\
  \hline
 \multirow{3}{*}{\emph{UCSD}}

 &{GOGA} &Faster
 &{57.4}&72.7& 9.43& 62&9&45&8&1886&{3742}&184&485&33.8&38.9&63.2 \\
 
&\multicolumn{2}{|c|}{{TBC (ours)}}& 49.5& \textbf{82.1}&\textbf{4.74}&62&11& 37&14&\textbf{948}&4435&\textbf{140}&\textbf{389}& \textbf{37.1}& {38.7}&\textbf{74.0}  \\

&\multicolumn{2}{|c|}{\textbf{TBC+Det}}&\textbf{58.7}&72.6&9.72& 62&\textbf{16}&39&\textbf{7}&1944&\textbf{3627} &{171}&508&35.6&\textbf{39.2}&69.6  \\

\hline
\multirow{3}{*}{\emph{LHI}}
 &{GOGA} &Faster
 &74.5&88.3&2.31&43&23 &19&1& 924 &2379 &173 &295&62.7 &{51.9}&70.0\\
 
 &\multicolumn{2}{|c|}{{TBC (ours)}}&79.1 &\textbf{93.6} &\textbf{1.26}&43&{29}& 13 & 1& \textbf{504} &1945 & \textbf{63} &\textbf{110}& {73.1}&64.6& \textbf{81.3} \\

&\multicolumn{2}{|c|}{{\textbf{TBC+Det}}}&\textbf{81.1}&93.0&1.42&43&\textbf{30}&12&1&569&\textbf{1759}&86 &132&\textbf{74.1}&\textbf{65.3}&81.2 \\

\hline
\multirow{4}{*}{\emph{S2L2}}
&\multicolumn{2}{|c|}{Wen et al.~\cite{wen2014multiple}}&71.2	&90.3	&1.47	&74	&27&44&3&640 &2402&125&175&{62.1}&-&52.7\\
&\multicolumn{2}{|c|}{Wen et al.~\cite{wen2016exploiting}}&74.4	&89.8	&1.62 	&74	&30	&42&2&708&2141&136&235&\textbf{64.2}&-&57.3\\
&\multicolumn{2}{|c|}{TBC(ours)}&{78.4}&81.9&3.32&74&{40}&33&\textbf{1}&1448&{1805}&300&352&57.5&-&66.6 \\
&\multicolumn{2}{|c|}{\textbf{TBC+Det}}& \textbf{79.5} &82.6&3.22&74&\textbf{41}& 32&\textbf{1}&1404&\textbf{1710} &257&281&59.6&-&\textbf{69.7} \\
\hline
\end{tabular}}
\vspace{-0.1in}
\end{table*}

The average results of \emph{MOT17} are reported in Tab.~\ref{table5}, and detailed results can be found on the MOT challenge website (our tracker is ``MOT$\_$TBC''). Here, we only report the results of TBC3, since some sequences are too long, which will consume excessive memory to build a graph over the whole sequence. In terms of MOTA,  TBC3 performs better than other \emph{tracking-by-detection} methods. TBC3 has lower FN, showing that our model can recover the missed objects using estimated density maps. To validate the efficiency of our model in (\ref{eq9}), we also create two variants, by removing the first two detection terms (``TBC3-Count'') and by removing the third counting term (``TBC3-Det''). As observed, both the detection term and the counting term contribute to improving the tracking performance. For the running time, TBC3 model runs the fastest among all the trackers.

Also, we evaluate our model on \emph{DukeMTMC}, and the results are summarized in Tab.~\ref{table6} (our tracker is ``MTMC$\_$TBC'' in the MOT challenge). For the easy test set, our TBC3 tracker achieves the same IDF1 value (89.2) with DeepCC, but has higher MOTA and MT. For the hard test set, TBC3 has higher IDF1 and MOTA, and lower FP and FN, which indicates that density maps can improve both detection and tracking performance especially in crowd scenes. The running time of TBC3 on \emph{DukeMTMC} is similar on \emph{MOT17}. 

\begin{figure*}[!hbtp]
\centering
\captionsetup[subfigure]{labelformat=empty}

{Detections} \\
\vspace{-0.10in}
\subfloat{\includegraphics[width=\widththird]{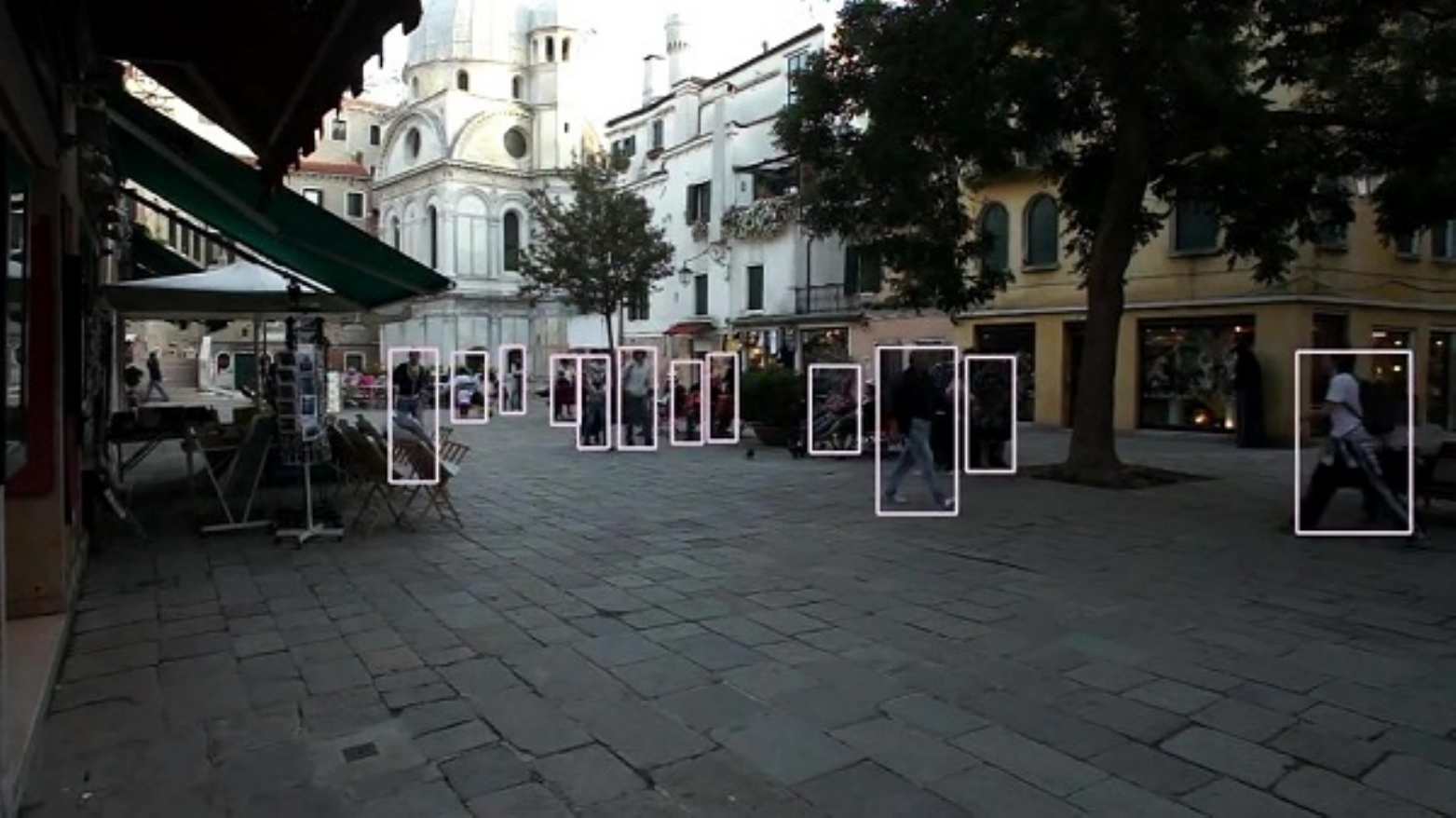}} \hspacefigure
\subfloat{\includegraphics[width=\widththird]{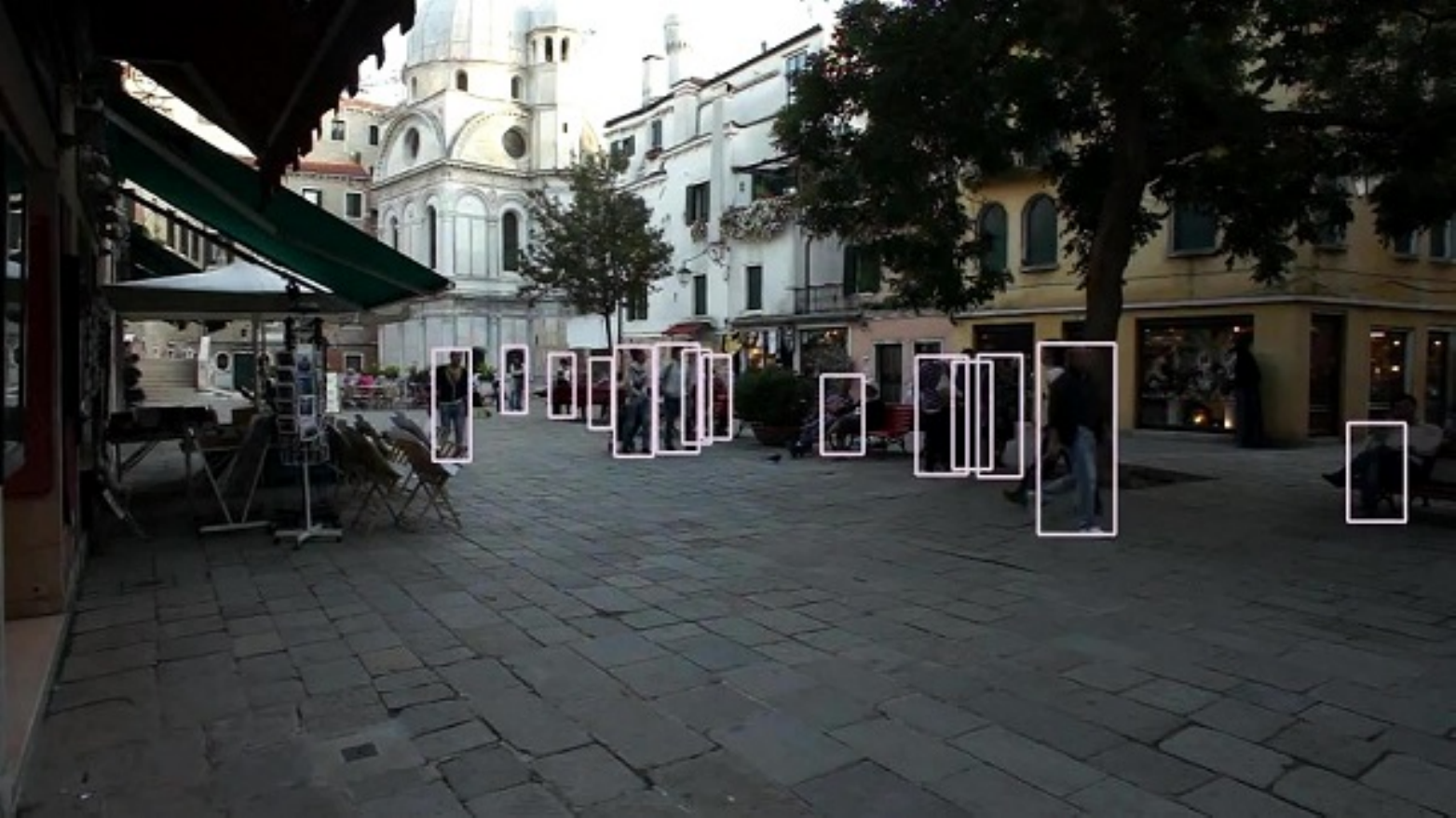}} \hspacefigure
\subfloat{\includegraphics[width=\widththird]{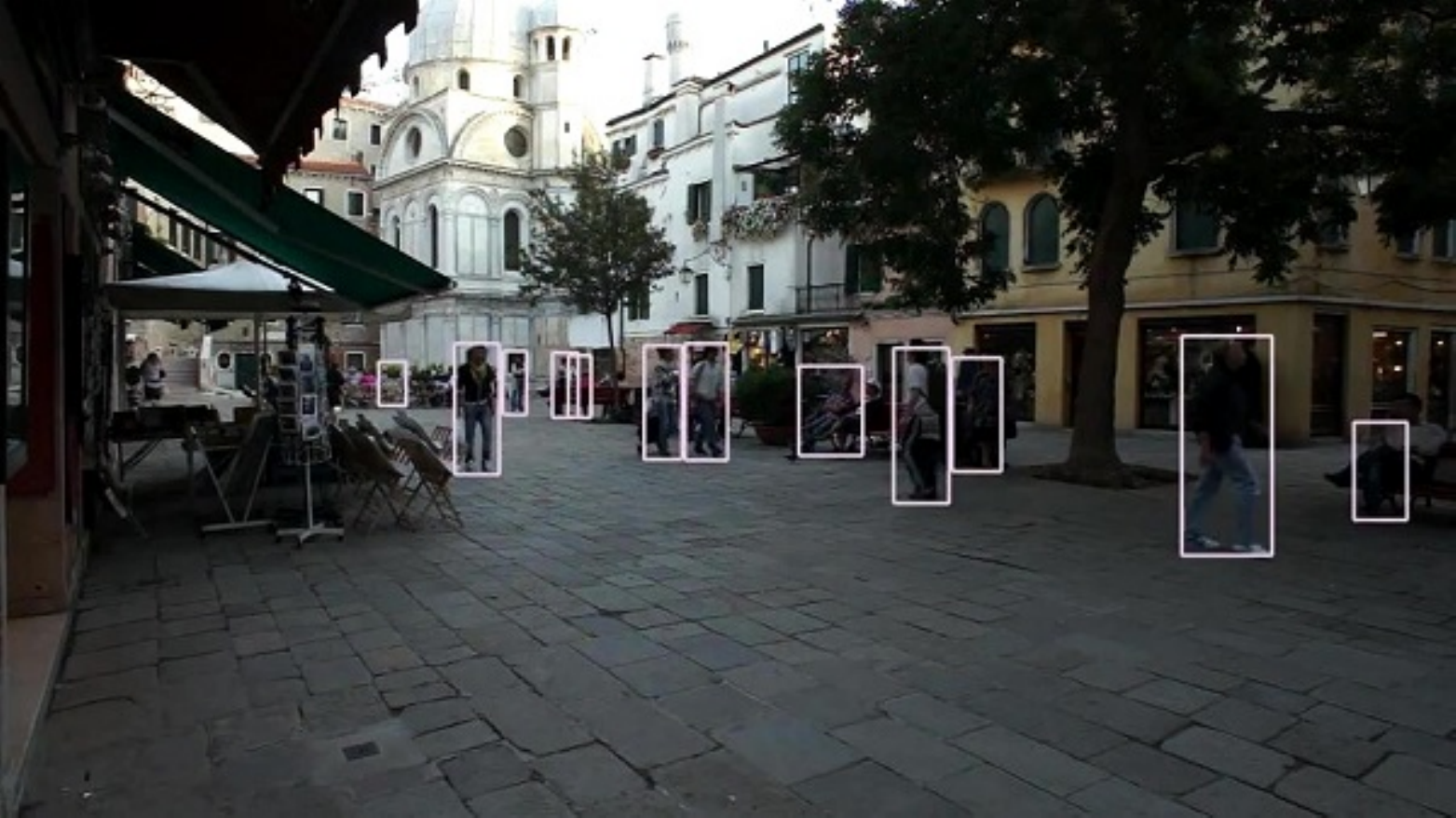}} \hspacefigure \\

JCC~\cite{keuper2018motion} \\
\vspace{-0.10in}
\subfloat{\includegraphics[width=\widththird]{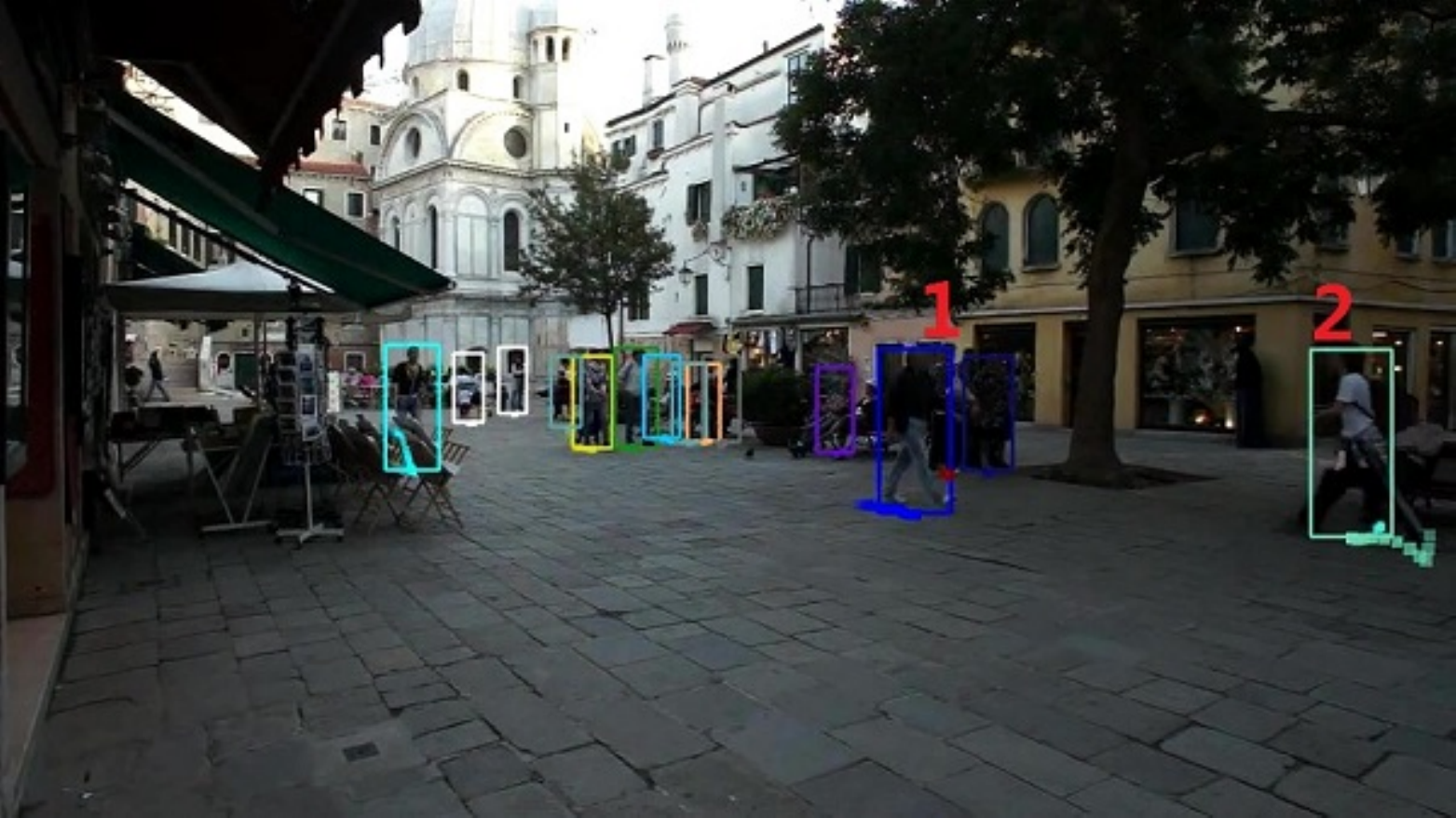}} \hspacefigure
\subfloat{\includegraphics[width=\widththird]{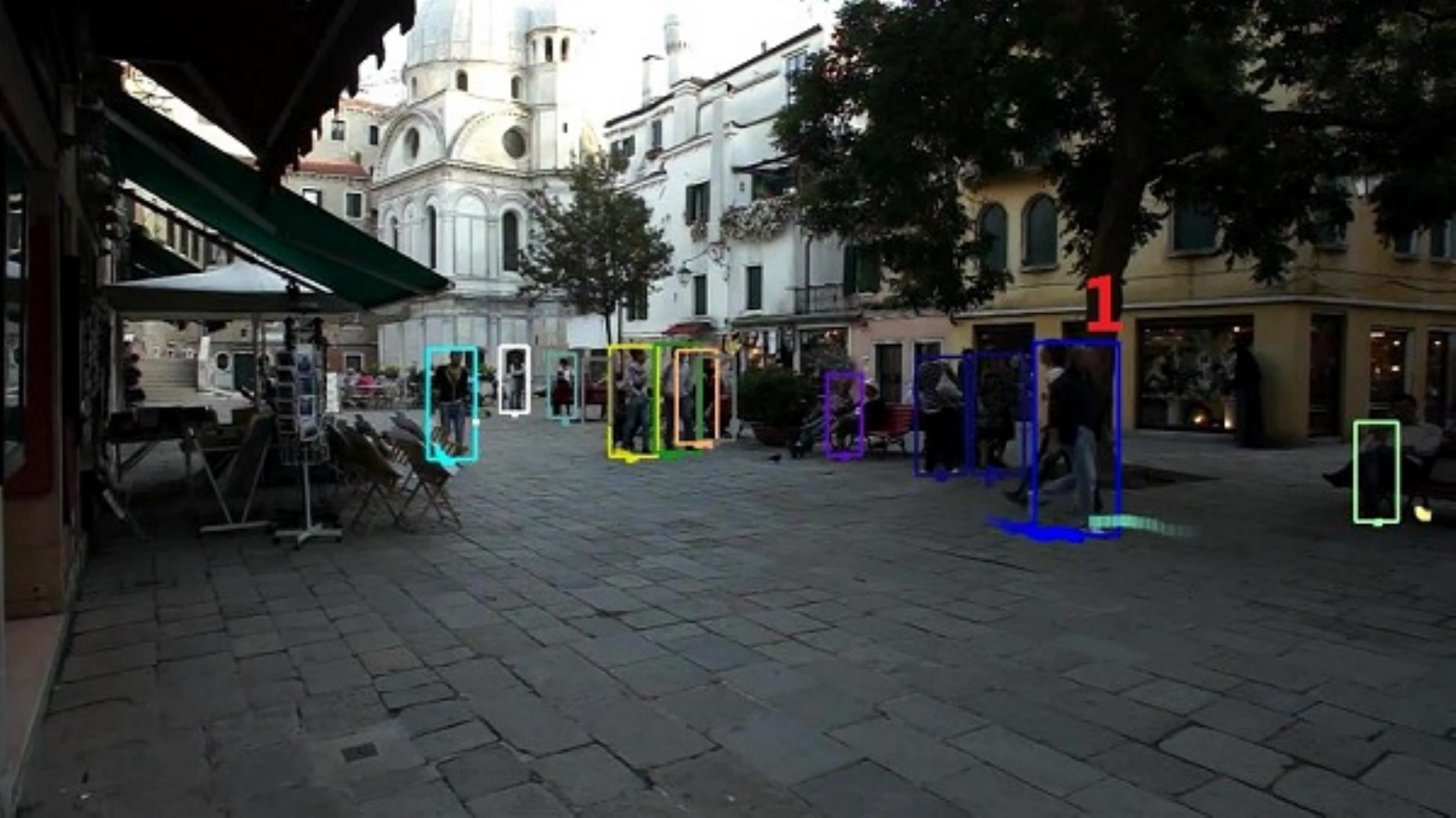}} \hspacefigure
\subfloat{\includegraphics[width=\widththird]{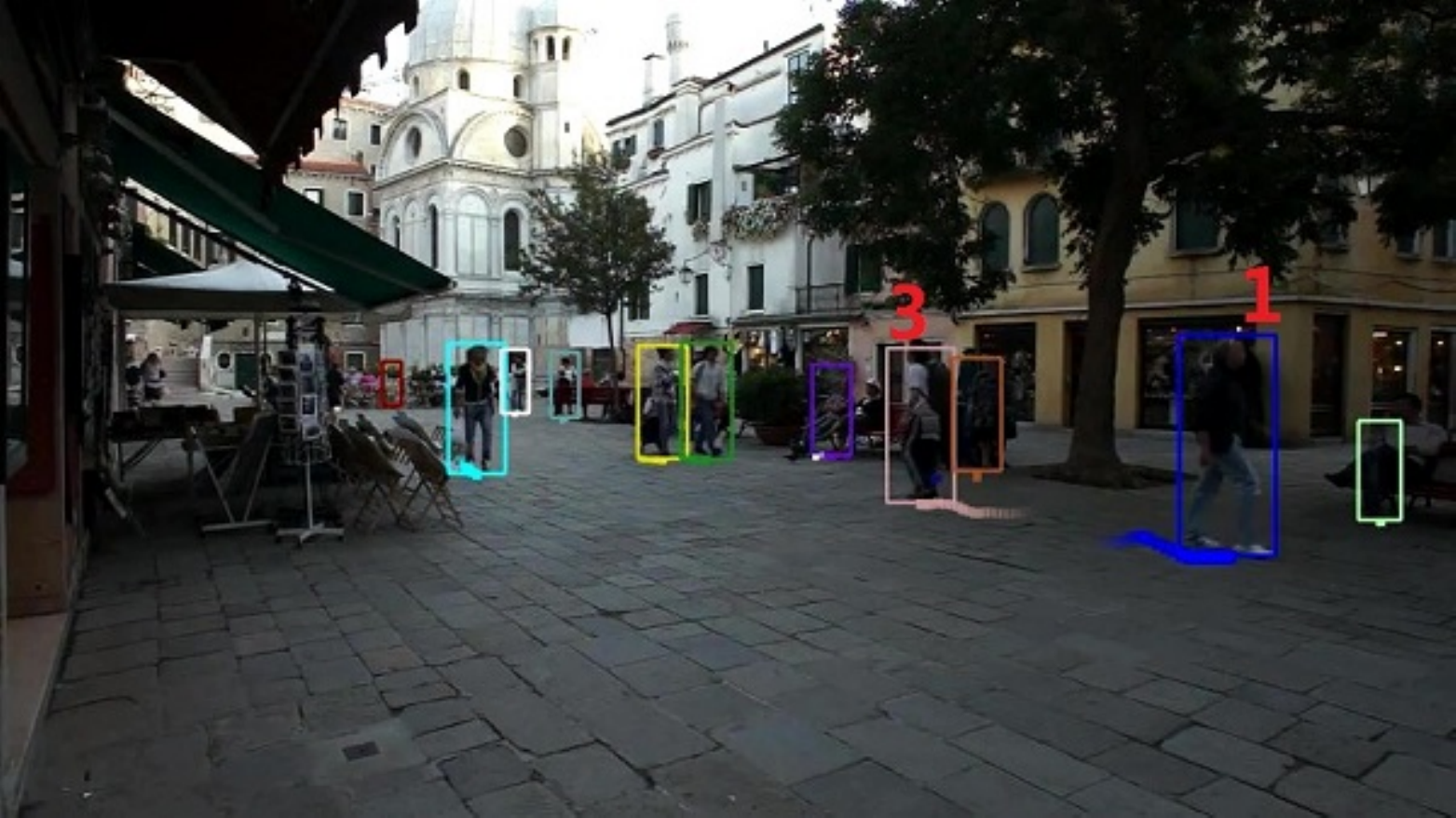}} \hspacefigure \\

Tracktor17~\cite{Phi2019} \\
\vspace{-0.10in}
\subfloat{\includegraphics[width=\widththird]{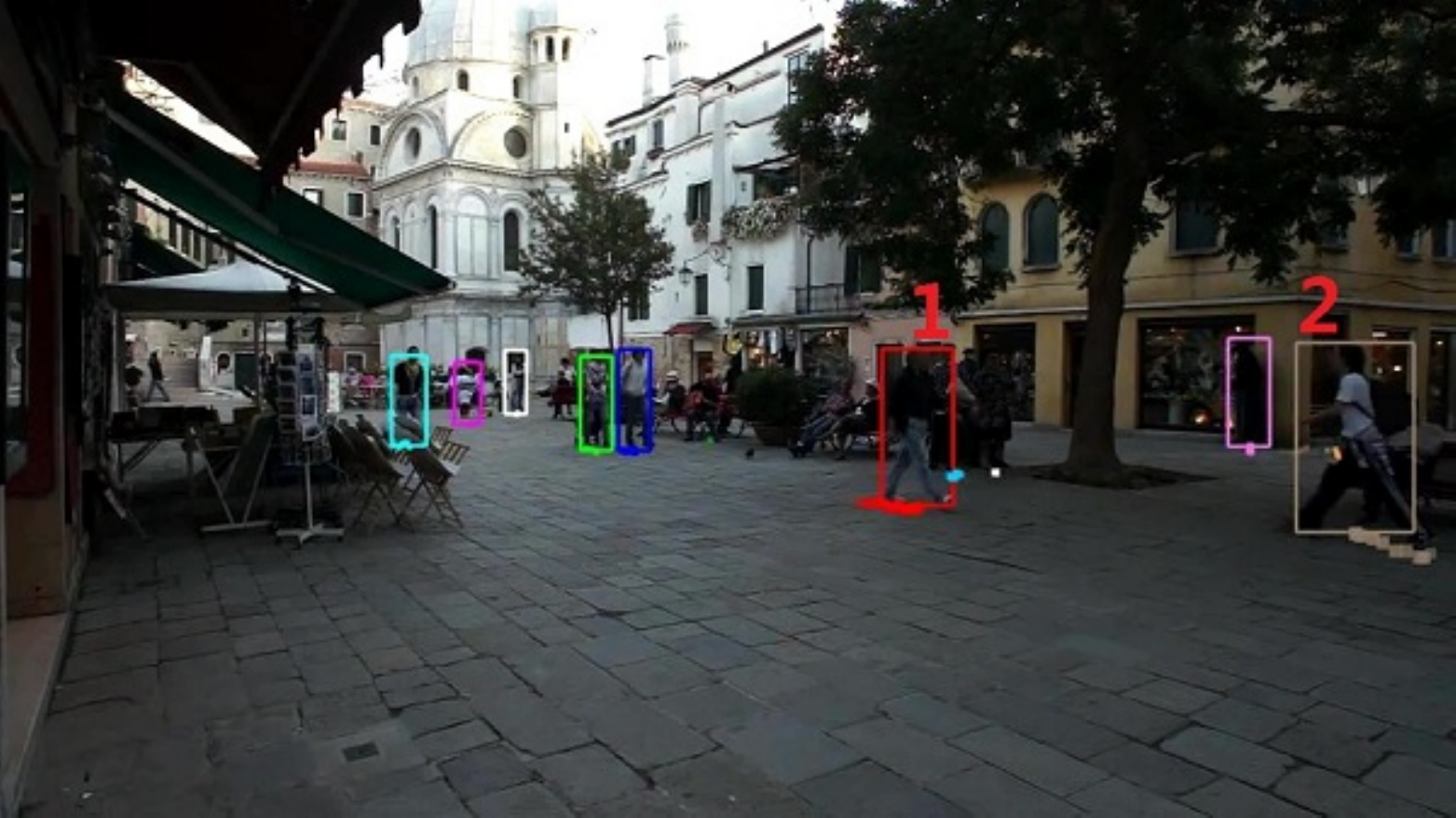}} \hspacefigure
\subfloat{\includegraphics[width=\widththird]{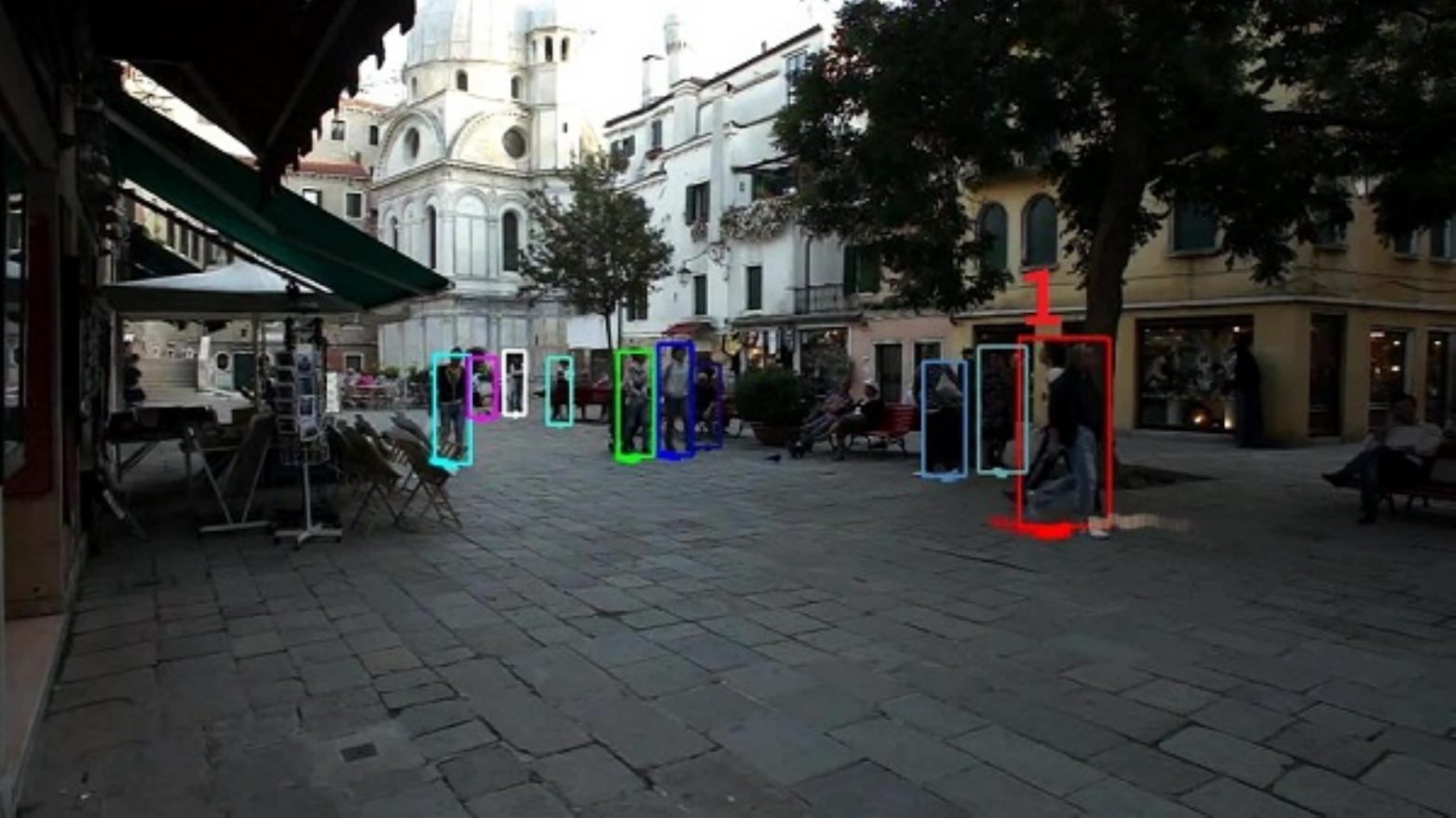}} \hspacefigure
\subfloat{\includegraphics[width=\widththird]{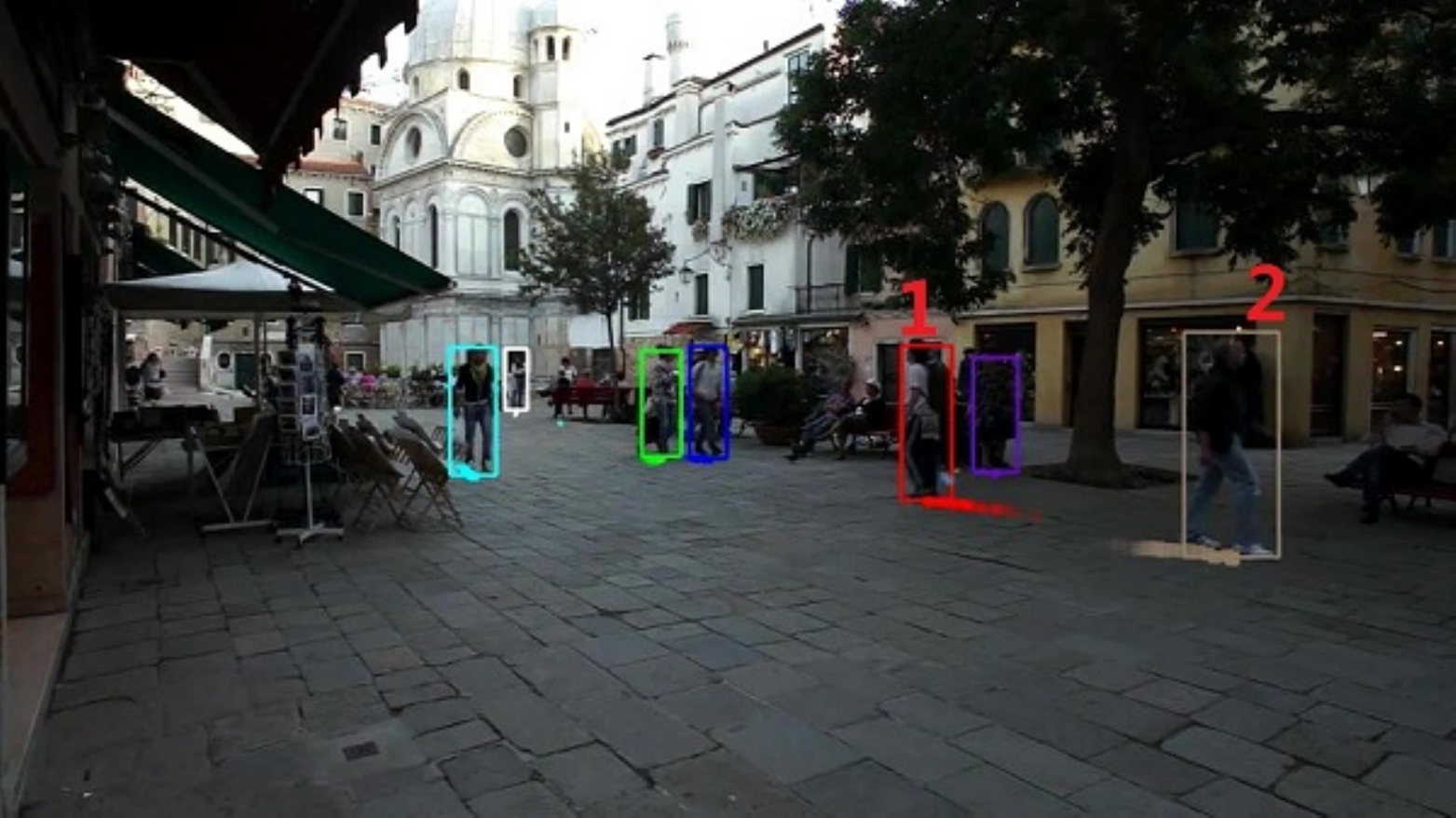}} \hspacefigure \\

TBC3 (ours)\\
\vspace{-0.10in}
\subfloat{\includegraphics[width=\widththird]{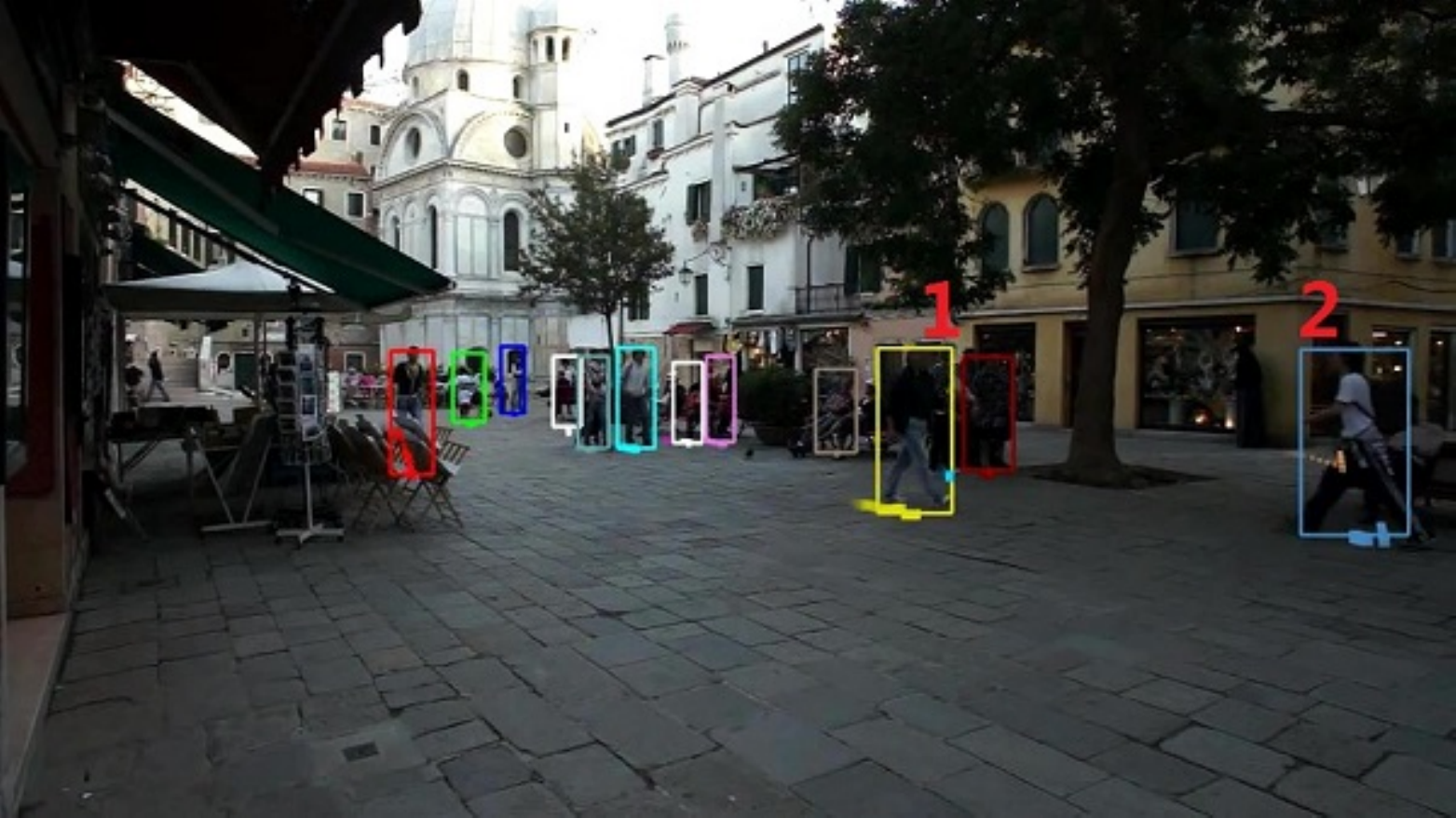}} \hspacefigure 
\subfloat{\includegraphics[width=\widththird]{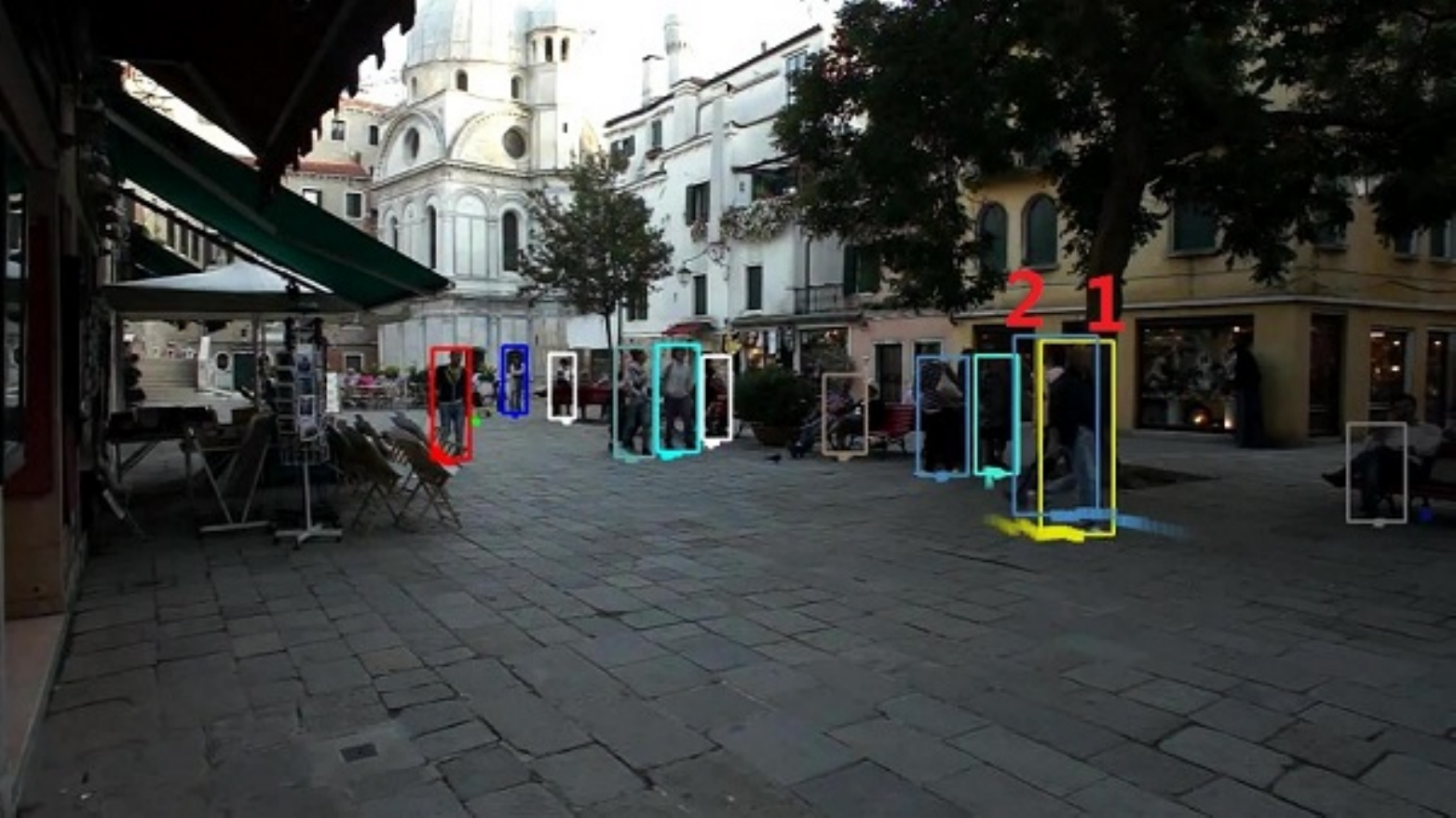}} \hspacefigure 
\subfloat{\includegraphics[width=\widththird]{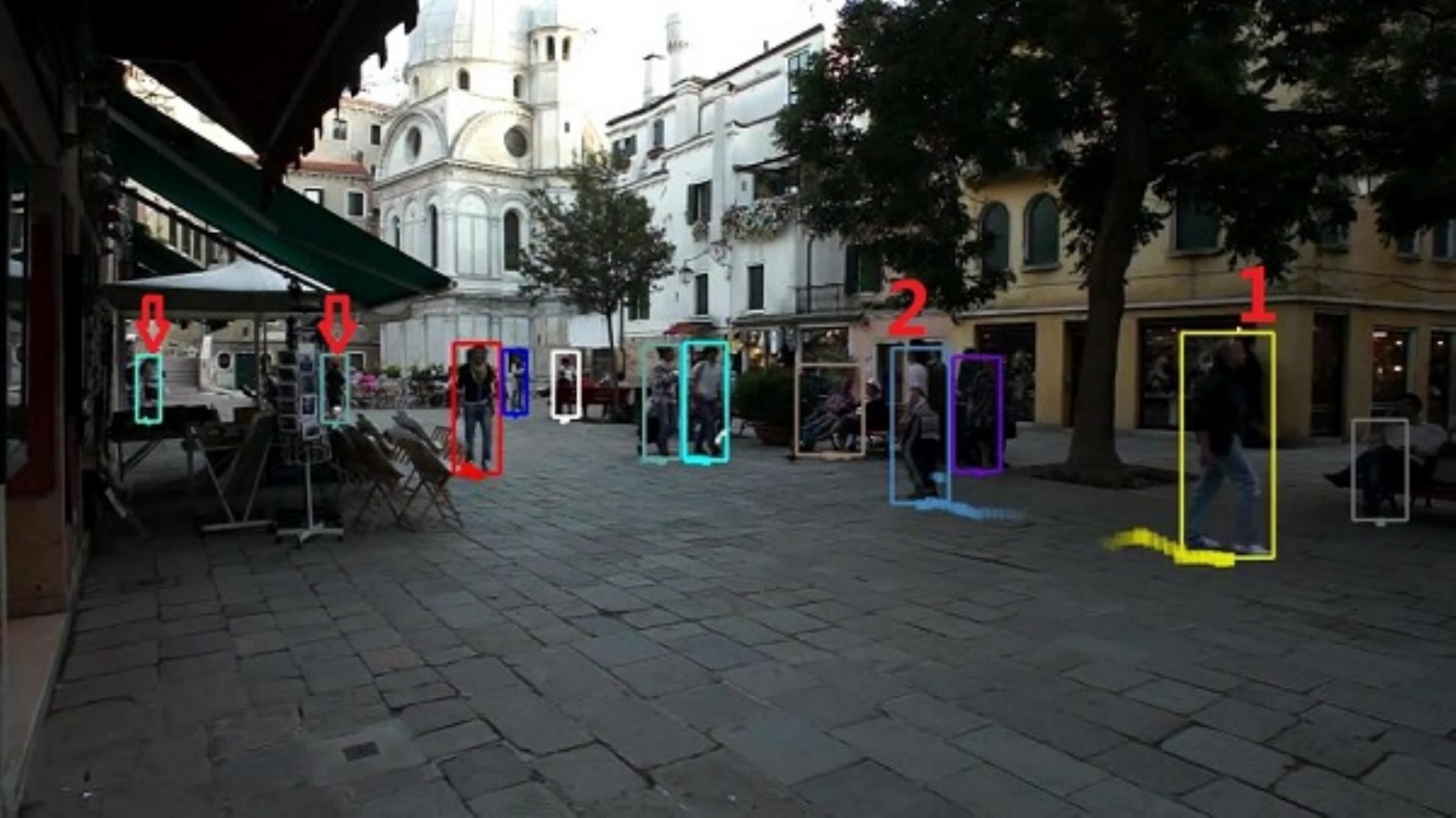}} \hspacefigure \\

\caption{Qualitative results on \emph{MOT17}. From left to right, frames 45, 85 and 110. Using both object detections and density maps, our TBC model can effectively reduce IDS. Besides, it can recover missed detections (marked red arrows)}\label{fig:mot}
\vspace{-0.08in}
\end{figure*}

{In Fig.~\ref{fig:mot}, we show tracking results of one example from \emph{MOT17}. 
JCC~\cite{keuper2018motion}~changes the object identity from ``2'' to ``3'' when two objects meet, while~Tracktor17~\cite{Phi2019}~causes identity switches. Using both object detections and density maps, our TBC model can effectively reduce IDS. Besides, it can recover missed detections (marked as red arrows).}

\begin{table}[htbp!]
\setlength{\hv}{0.10cm}
\renewcommand\arraystretch{1.25}
\caption{Multi-object tracking results on \emph{MOT20}.} \label{MOT20}
\centering
\begin{tabular}{|c|ccccc|}
  \hline
  Tracker&MOTA$\uparrow$&IDF1$\uparrow$ &MT$\uparrow$&ML$\downarrow$&IDS$\downarrow$\\
  \hline
UnsupTrack~\cite{karthik2020simple}
 &{53.6}&\textbf{50.6}&376&311&\textbf{2,178}\\

SORT~\cite{bewley2016simple}&
 42.7&45.1&208&326&4,470\\

TBC3&
\textbf{54.5}&{50.1}&\textbf{415}&\textbf{245}&2,445 \\
 \hline
\end{tabular}
\end{table}

\wh{Finally, we evaluate our model on \emph{MOT20} dataset where the average number of targets per frame can reach 100. Our model now ranks second in the leaderboard, which indicates its effectiveness crowd scenes. In Tab.~\ref{MOT20}, we compare our TBC3 model wth two released works UnsupTrack~\cite{karthik2020simple} and SORT~\cite{bewley2016simple} from MOT20. \cite{karthik2020simple} trained a ReID network to cluster tracklets, and thus achieves relatively higher IDF1. However, our TBC3 model can find more missed detecions using density maps, and thus has higher MOTA and MT. Please see {MOT20 challenge} for more details~(our tracker is denoted as ``MOT20-TBC'').}

\subsection{{Adding detections for \emph{UCSD}, \emph{LHI} and \emph{PETS2009}}}
{In the experiments on crowd scenes in Section \ref{text:ucsdlhi}, we only used the model in (\ref{eq8}) for tracking, and did not use the detections for \emph{UCSD}, \emph{LHI} and \emph{PETS2009}. Here, we introduce detections for them (denoted as ``TBC+Det''), and analyze if model (\ref{eq10}) can further improve the tracking performance. The tracking results when adding detections to TBC are summarized in Tab.~\ref{table7}. 
For \emph{LHI} and \emph{PETS2009}, TBC+Det model can further improve
TBC, e.g., MOTA is improved
from 57.5 to 59.6 on \emph{PETS2009}. For \emph{UCSD}, TBC+Det yields
slightly lower MOTA compared with TBC, but has higher
MT and IDF1. In summary, tracking results can be further improved when combining TBC model with object detections as in (\ref{eq10}).}

\section{Conclusions}
\label{text:conclusion}
In this paper, we propose a novel MOT approach explicitly designed for crowded scenes.
Unlike existing tracking-by-detection MOT methods that focus on
associating per-frame detections and existing density-based tracking approaches
that rely on heuristic short point-tracks,
our model explicitly accounts for the object counts inferred from density maps
and simultaneously solves multi-object detection and tracking over the whole video sequence.
This is achieved by modeling the joint problem as a network flow program 
for which the optimal solutions can be found using standard commercial solvers.
Our method achieves promising results on public datasets featuring people-, cell-, and fish-tracking scenarios.
{We also extend our model to use object detections, which can further improve performance on large-scale scenes.}

\section*{Acknowledgements}
The work described in this paper was supported by
a grant from the Research Grants Council of the Hong
Kong Special Administrative Region, China (Project No.
[T32-101/15-R] and CityU 11212518), by a Strategic Research Grant from City
University of Hong Kong (Project No. 7004887), and
by the Natural Science Foundation of China under Grant 91648118. Also, the work was  supported by the startup funding of Stevens Insitute of Technology.
We gratefully acknowledge the support of
NVIDIA Corporation with the donation of the Tesla K40
GPU used for this research.


\ifCLASSOPTIONcaptionsoff
  \newpage
\fi

\footnotesize
\bibliographystyle{IEEEtran}
\bibliography{IEEEexample}
\vspace{5cm}
\begin{IEEEbiography}[{\includegraphics[width=1in,height=1.25in,clip,keepaspectratio]{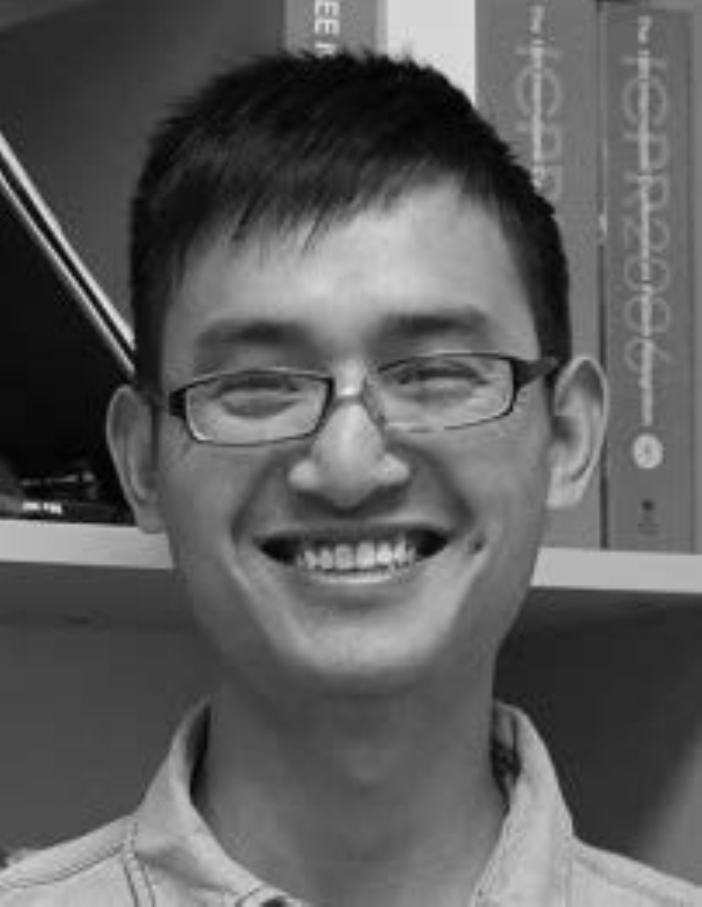}}]
{Weihong Ren} received the B.E. degree in Automation and Electronic Engineering from Qingdao University of Science and Technology, Qingdao, China, in 2013. He is currently in a joint PhD scheme offered by University of Chinese Academy of Sciences and City University of Hong Kong, China. His current research interests include people tracking, image restoration and deep learning.
\end{IEEEbiography}

\vspace{-2.0cm}
\begin{IEEEbiography}[{\includegraphics[width=1in,height=1.25in,clip,keepaspectratio]{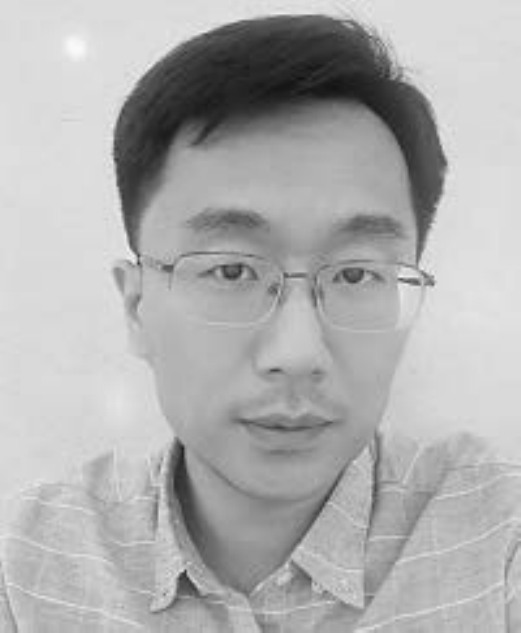}}]
{Xinchao Wang} is currently a tenure-track Assistant Professor at Stevens Institute of Technology, New Jersey, United States. Before joining Stevens, he was an SNSF postdoctoral fellow at University of Illinois Urbana-Champaign (UIUC). He received a PhD from Ecole Polytechnique Federale de Lausanne (EPFL) in 2015, and a first-class honorable degree from Hong Kong Polytechnic University (HKPU) in 2010. His research interests include artificial intelligence, computer vision, machine learning, medical image analysis, and multimedia. His articles have been published in major venues including CVPR, ICCV, ECCV, NeurIPS, AAAI, IJCAI, MICCAI, TPAMI, TIP, TMI, and TNNLS. He serves as an associate editor of Journal of Visual Communication and Image Representation (JVCI), as a senior program committee member of AAAI’19 and IJCAI’19, and as an area chair of ICME’19 and ICIP’19.
\end{IEEEbiography}

\vspace{-2.0cm}
\begin{IEEEbiography}[{\includegraphics[width=1in,height=1.25in,clip,keepaspectratio]{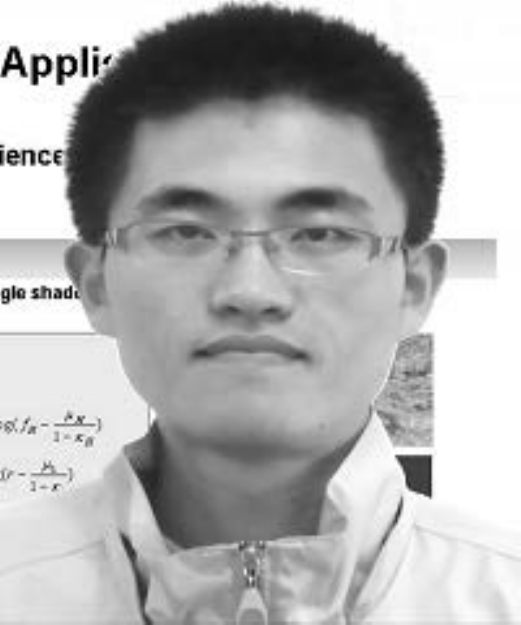}}]{Jiandong Tian}
received his B.S. degree in the department of automation, Heilongjiang University, P.R. China, in 2005. In 2011 he received a doctor's degree in Shenyang Institute of Automation, Chinese Academy of Sciences. Currently he is an associate professor in Shenyang Institute of Automation, Chinese Academy of Sciences. His research interests include Illumination and Reflectance Modeling,image processing, and pattern recognition.
\end{IEEEbiography}

\vspace{-2.0cm}
\begin{IEEEbiography}[{\includegraphics[width=1in,height=1.25in,clip,keepaspectratio]{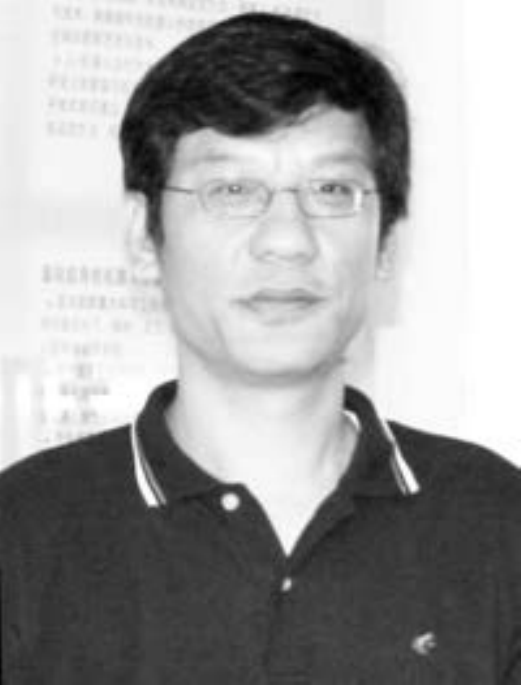}}]
{Yandong Tang} received the B.S. and M.S. degrees in mathematics from Shandong University, China, in 1984 and 1987, respectively, and the Ph.D. degree  in  applied  mathematics  from  the  University of Bremen, Germany, in 2002. He is currently a Professor  with  the  Shenyang  Institute  of  Automation, Chinese Academy of Sciences. His research interests
include  numerical computation, image  processing, and computer vision.
\end{IEEEbiography}

\vspace{-2.0cm}
\begin{IEEEbiography}[{\includegraphics[width=1in,height=1.25in,clip,keepaspectratio]{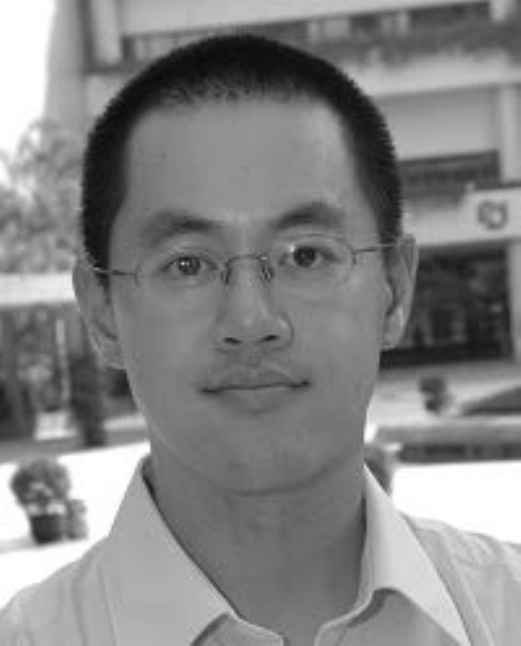}}]
{Antoni B. Chan} received the B.S. and M.Eng. degrees in electrical engineering from Cornell University, Ithaca, NY, in 2000 and 2001, and the Ph.D. degree in electrical and computer engineering from the University of California, San Diego (UCSD), San Diego, in 2008. 
He is currently an Associate Professor in the Department of Computer Science, City University of Hong Kong. His research interests include computer vision, machine learning, pattern recognition, and music analysis.
\end{IEEEbiography}

\end{document}